\documentclass[10pt,a4paper]{article}

\usepackage{hyperref}       
\usepackage{url}            
\usepackage{booktabs}       
\usepackage{amsfonts}       
\usepackage{nicefrac}       
\usepackage{microtype}      

\usepackage[nomain,toc,acronym,nogroupskip]{glossaries}

\usepackage{fullpage}
\usepackage[square,numbers]{natbib}

\usepackage{setspace}

\usepackage{paralist}

\usepackage{multirow}
\usepackage{rotating}

\usepackage{makecell}

\usepackage{subfig}

\usepackage{endfloat}

\usepackage{colortbl}

\usepackage{tikz}
\usetikzlibrary{decorations.pathreplacing,matrix}

\tikzset{ 
table/.style={
  matrix of math nodes,
  row sep=-\pgflinewidth,
  column sep=-\pgflinewidth,
  nodes={rectangle,text width=3em,align=center},
  text depth=1.25ex,
  text height=2.5ex,
  nodes in empty cells,
  left delimiter=[,
  right delimiter={]},
  ampersand replacement=\&
}
}

\usepackage[sfdefault]{overlock}

\newacronym{AD}{AD}{Alzheimer's disease}
\newacronym{MCI}{MCI}{Mild Cognitive Impairment}
\newacronym{PET}{PET}{Positron Emission Tomography}
\newacronym{MRI}{MRI}{Magnetic Resonance Imaging}
\newacronym{rCBF}{rCBF}{regional cerebral blood flow}
\newacronym{3DSSP}{3D-SSP}{Three-Dimensional Stereotactic Surface Projection}
\newacronym{FTD}{FTD}{Frontotemporal Dementia}
\newacronym{ICBM}{ICBM}{International Consortium for Brain Mapping}
\newacronym{SPM}{SPM}{Statistical Parametric Mapping}
\newacronym{PD}{PD}{positron density}
\newacronym{CSF}{CSF}{cerebrospinal fluid}
\newacronym{APOE}{APOE}{apolipoprotein E}
\newacronym{BMI}{BMI}{body mass index}
\newacronym{AB}{A$\beta$}{amyloid-beta}
\newacronym{GM}{GM}{gray matter}
\newacronym{WM}{WM}{white matter}
\newacronym{ROI}{ROI}{regions of interest}
\newacronym{ROR}{ROR}{regions of relevance}
\newacronym{PCC}{PCC}{posterior cingulate cortex}
\newacronym{PAC}{PAC}{parietal association cortex}
\newacronym{TAC}{TAC}{temporal association cortex}

\newacronym[longplural=Gaussian Processes]{GP}{GP}{Gaussian Process}
\newacronym{ARD}{ARD}{Automatic Relevance Determination}
\newacronym[longplural=Support Vector Machines]{SVM}{SVM}{Support Vector Machine}
\newacronym[longplural=Relevance Vector Machines]{RVM}{RVM}{Relevance Vector Machine}
\newacronym{PCA}{PCA}{Principal Component Analysis}
\newacronym{GA}{GA}{Genetic Algorithm}
\newacronym{NB}{NB}{Na\"{\i}ve Bayes}
\newacronym{STL}{STL}{Self-Taught Learning}
\newacronym{GPML}{GPML}{Gaussian Processes for Machine Learning}
\newacronym{PSD}{PSD}{positive semidefinite}
\newacronym[longplural=Reproducing Kernel Hilbert Spaces]{RKHS}{RKHS}{Reproducing Kernel Hilbert Space}
\newacronym{SE}{SE}{squared-exponential}
\newacronym{NN}{NN}{neural network}
\newacronym{LA}{LA}{Laplace Approximation}
\newacronym{EP}{EP}{Expectation Propagation}
\newacronym{VB}{VB}{Variational Bounds}
\newacronym{MAP}{MAP}{Maximum-a-Posteriori}
\newacronym{RBF}{RBF}{Radial Basis Function}
\newacronym{CV}{CV}{cross-validation}
\newacronym{MKL}{MKL}{Multiple Kernel Learning}
\newacronym{SILP}{SILP}{Semi-Infinite Linear Program}
\newacronym{BFGS}{BFGS}{Broyden-Fletcher-Goldfarb-Shanno}
\newacronym{AUC}{AUC}{Area Under Curve}
\newacronym{ROC}{ROC}{Receiver Operating Characteristic}
\newacronym{AVA}{AVA}{All-vs-All}
\newacronym{OVA}{OVA}{One-vs-All}
\newacronym{OVO}{OVO}{One-vs-One}
\newacronym{ADF}{ADF}{Assumed Density Filtering}
\newacronym{LBP}{LBP}{Loopy Belief Propagation}
\newacronym{SVD}{SVD}{singular value decomposition}
\newacronym{MKLARD}{MKL-AsRD}{Multiple Kernel Learning for Automatic Subspace Relevance Determination}
\newacronym{SDP}{SDP}{Semidefinite Program}
\newacronym{QCQP}{QCQP}{Quadratically-Constrained Quadratic Program}
\newacronym{SOCP}{SOCP}{Second-Order Cone Programming}
\newacronym{SMO}{SMO}{Sequential Minimal Optimization}
\newacronym{ECOC}{ECOC}{Error Correcting Output Codes}
\newacronym{LR}{LR}{logistic regression}
\newacronym{ICA}{ICA}{Independent Component Analysis}
\newacronym{sICA}{sICA}{spatial Independent Component Analysis}
\newacronym{AE}{AE}{autoencoder}
\newacronym{SAE}{SAE}{sparse autoencoder}
\newacronym{SPGP}{SPGP}{Sparse Pseudo-Input Gaussian Process}
\newacronym{MVPA}{MVPA}{multivariate predictive analysis}
\newacronym[longplural=Multiple Gaussian Processes]{MGP}{MGP}{Multiple Gaussian Process}
\newacronym{MTL}{MTL}{Multitask Learning}
\newacronym{CNN}{CNN}{convolutional neural network}

\newacronym{NIA}{NIA}{National Institute on Aging}
\newacronym{NIH}{NIH}{National Institutes of Health}
\newacronym{FNIH}{FNIH}{Foundation for the National Institutes of Health}
\newacronym{AA}{AA}{Alzheimer's Association}
\newacronym{ADNI}{ADNI}{Alzheimer's Disease Neuroimaging Initiative}
\newacronym{NIBIB}{NIBIB}{National Institute of Biomedical Imaging and Bioengineering}
\newacronym{FDA}{FDA}{Food and Drug Administration}
\newacronym{DOD}{DOD}{Department of Defense}
\newacronym{NCIRE}{NCIRE}{Northern California Institute for Research and Education}
\newacronym{CIHR}{CIHR}{Canadian Institutes of Health Research}
\newacronym{WHO}{WHO}{World Health Organization}
\makeglossaries

\begin{document}
\title{\textbf{Multiple Kernel Learning and Automatic Subspace Relevance Determination for High-dimensional Neuroimaging Data}}
\author{\textbf{Murat Se\c{c}kin Ayhan}\thanks{M.S. Ayhan is with the Department of Computer Engineering, I\c{s}\i{k} University, Istanbul, Turkey, 34980. However, the majority of this work was completed at the Center for Advanced Computer Studies at the University of Louisiana at Lafayette where he completed his studies.} \\
  Department of Computer Engineering \\
  I\c{s}{\i}k University \\
  \c{S}ile, Istanbul, TURKEY 34980 \\
  \texttt{muratseckin.ayhan@isikun.edu.tr}
  \and
  \textbf{Vijay~Raghavan} \\
  Center for Advanced Computer Studies \\
  University of Louisiana at Lafayette \\
  Lafayette, LA, USA 70503 \\
  \texttt{vijay@cacs.louisiana.edu} \\
  \and
  \textbf{The Alzheimer's Disease Neuroimaging Initiative (ADNI)}\thanks{Data in this article are from ADNI (adni.loni.usc.edu). As such, ADNI provided data but did not participate in writing of this report. 
  Mark and Mary Stevens Neuroimaging and Informatics Institute \\
  University of Southern California \\
  \texttt{adni@loni.usc.edu}  
}

\providecommand{\keywords}[1]{\textbf{\textit{Keywords---}} #1}

\date{}
\maketitle

\begin{abstract}
\acrlong{AD} is a major cause of dementia. Its 
diagnosis requires accurate biomarkers that are sensitive to disease stages. 
In this respect, we regard probabilistic classification as a method of designing a probabilistic biomarker for disease staging. 
Probabilistic biomarkers 
naturally support the interpretation of decisions and evaluation of uncertainty associated with them. In this paper, we obtain probabilistic biomarkers via \acrlongpl{GP}. \acrlongpl{GP} enable probabilistic kernel machines that offer flexible means to accomplish \acrlong{MKL}. Exploiting this flexibility, we propose a new variation of \acrlong{ARD} and tackle the challenges of high dimensionality through multiple kernels. 
Our research results demonstrate that the \acrlong{GP} models are competitive with or better than the well-known \acrlong{SVM} in terms of classification performance even in the cases of single kernel learning. Extending the basic scheme towards the \acrlong{MKL}, we improve the efficacy of the \acrlong{GP} models and their interpretability in terms of the known anatomical correlates of the disease. For instance, the disease pathology starts in and around the hippocampus and entorhinal cortex. 
Through the use of \acrlongpl{GP} and \acrlong{MKL}, we have automatically and efficiently determined those portions of neuroimaging data. 
In addition to their interpretability, our \acrlong{GP} models are competitive with recent deep learning solutions under similar settings. 
\end{abstract}


\section{Introduction}
\Gls{AD} is 
the 6\textsuperscript{th} leading cause of death in the United States~\citep{ADfacts2017}. 
Its pathology induces complex patterns that evolve as the disease progresses~\citep{Fan2008}. The pathology mainly starts in and around the \emph{hippocampus} and \emph{entorhinal cortex}; however, by the time of a clinical diagnosis, the cerebral atrophy is widespread and it involves, to a large extent, temporal, parietal, and frontal lobes~\citep{Whitwell2007,Fan2008}. 
Moreover, the \Gls{AD} pathology does not necessarily conform to anatomical boundaries \citep{Fan2008}. Thus, it is highly recommended that the entire brain, instead of predetermined \emph{\gls{ROI}}, be examined for accurate diagnosis \citep{Whitwell2007,Fan2008}. 

The most recent criteria and guidelines for \Gls{AD} diagnosis, as proposed by the \Gls{NIA}\footnote{{http://www.nia.nih.gov/}} 
and the \Gls{AA}, describe a \emph{continuum} in which an individual experiences a smooth transition from functioning normally, despite the changes in the brain, to failing to compensate for the deterioration of cognitive abilities \citep{ADfacts2017}. 
There are three stages of \Gls{AD}:
\begin{inparaenum}[i)]
	\item preclinical \Gls{AD},
	\item \Gls{MCI} due to \Gls{AD}, and 
	\item dementia due to \Gls{AD}. 
\end{inparaenum}
In the preclinical stage, measurable physiological changes may be observed; however, the symptoms are not developed \citep{ADfacts2017}. \Gls{MCI} is a transitional state between normal aging and \Gls{AD} \citep{Petersen2001}. Individuals with \Gls{MCI} experience measurable changes in their cognitive abilities, which are also notable by family and friends, but the symptoms are not significant enough to affect the patient's daily life \citep{ADfacts2017}. 
On the downside, \Gls{MCI} shares features with \Gls{AD} and it is likely to progress to \Gls{AD} at an accelerated rate \citep{Petersen2001}. 

\subsection{Biomarkers}\label{sec:biomarkers}
A \emph{biomarker} quantifies a biological state, i.e., the presence or absence of a disease, or the risk of developing a particular condition. According to the \Gls{WHO},  
``any substance, structure or process that can be measured in the body or its products and influence or predict the incidence of outcome or disease'' 
is a biomarker \citep{WHO2001}. 
Examples include blood glucose level for diabetes, cholesterol level for heart disease risk, and \gls{BMI} for obesity. Ideally, a biomarker is safe, easy to measure, repeatable, and sensitive and specific to the state or condition targeted. 

The guidelines for \Gls{AD} diagnosis identify biomarkers that measure 
\begin{inparaenum}[i)]
	\item	the \gls{AB} accumulation in the brain and 
	\item	if the neurons in the brain 
	are injured or degenerating \citep{ADfacts2017}. 
\end{inparaenum}
These biomarkers will be essential to identify the disease during the preclinical or \Gls{MCI} stages, administer disease-modifying treatments, and monitor the progression towards \Gls{AD}. 
However, more research is needed to evaluate the effectiveness of these biomarkers and investigate potential combinations
of them for more accurate screening \citep{ADfacts2017}. 

Neuroimaging techniques such as \Gls{PET} and \Gls{MRI} are useful in identifying brain changes due to tumors and other damages that can explain a patient's symptoms. As a result, they have been used as imaging biomarkers~\citep{Matsuda2007}. However, it is virtually impossible to visually detect a slight decrease in \acrlong{rCBF} or glucose metabolism in cases of early \Gls{AD}~\citep{Imabayashi2004}. Also, such an inspection is susceptible to other factors like subjectivity and experience of the physician~\citep{Matsuda2007,Kloppel2008}. On the other hand, voxel-based representations of neuroimagery can be used to perform both standardization and data-driven analysis of brain imagery \citep{Minoshima1995,Matsuda2007}. 

\subsection{Computerized Diagnosis}\label{sec:compDiag}
\citet{Kloppel2008} compared the accuracy of dementia diagnosis provided by radiologists to that of a computer-based diagnostic method. Basically, they extracted the segments of \gls{GM} from complete \Gls{MRI} scans, normalized them into a standard template, and ended up with several thousands of voxels. Then, they used \Glspl{SVM} for binary classification problems: 
\begin{inparaenum}[i)]
	\item	healthy (normal) vs. \Gls{AD} and 
	\item	fronto-temporal lobar degeneration vs. \Gls{AD}. 
\end{inparaenum} 
They also recruited radiologists with different levels of experience for the same problems. 
The well-trained and experienced radiologists are competitive with the computerized method; however, the overall performance of radiologists is susceptible to 
human factors \citep{Kloppel2008}. 
As a result, the radiologists
induce a larger variance in the diagnostic accuracy. In this regard, \citet{Kloppel2008} argue that the accuracy of the computerized diagnosis of \Gls{AD} is equal to or better than that of radiologists. 

In summary, a general adoption of computerized methods for visual image interpretation for dementia diagnosis is strongly recommended \citep{Minoshima1995,Imabayashi2004,Matsuda2007,Kloppel2008}.  
Given the variety of computational methods and growing interest of researchers from a diverse set of disciplines, better diagnostic tools are also expected in the near future. In this respect, \citet{Kloppel2008} stress the need 
for further research on probabilistic methods because medical experts could have more comfort from knowing the confidence levels associated with decisions and having better ability to interpret the 
models behind them. 

\subsection{Motivation of Work}\label{sec:Motivation}
Machine learning algorithms can be viewed in two main categories: \emph{discriminative} and \emph{generative}. Discriminative algorithms exploit labeled training data in order to uncover input--output relations. These methods can demonstrate excellent generalization performances even though 
little is known about the generating mechanisms of data. On the other hand, generative methods aim to capture the generating mechanisms of data and to discover the structure of classes \citep{Tu2007}. However, unless the representation extracted by a generative method completely epitomizes reality, its generalization performance is usually poor. 

The cost of neuroimaging 
data is high since the gathering process involves expensive imaging procedures and domain experts. Thus, sample sizes are small. Also, neuroimaging procedures usually generate high-dimensional data. 
An attempt to examine the complete brain imagery complicates statistical analysis and modeling, resulting in high computational complexity, and typically more sophisticated and perhaps less comprehensible models. 

Due to the complexity of atrophy patterns and high dimensionality of data, we have resorted to discriminative methods and focused on a particular family of algorithms known as kernel machines. \Gls{SVM} is probably the best known kernel machine and it has been the workhorse of the multivariate predictive analysis of brain data~\citep{Kloppel2008,Fan2008,Hinrichs2009,Ayhan2010,
Zhang2011,Yang2011,Hinrichs2012,Ayhan2013a}.
The popularity of \Gls{SVM} is mainly due to its theoretical soundness and guarantees on generalization performance in spite of high dimensionality. However, \Gls{SVM} characteristically lacks the support for probabilistic interpretation of the decision function and evaluation of uncertainty associated with decisions. Even though solutions (Section~\ref{sec:ProbOutput}) have been proposed to calibrate the \Gls{SVM} outputs for probabilistic information, these procedures are \emph{ad hoc} instead of being an integral part of the \Gls{SVM} framework. 
On the other hand, \Glspl{GP} enable a holistic view of Bayesian probability theory that meets with the need for computational tractability \citep{GPMLbook}. The result is a probabilistic kernel machine with 
comprehensive support for interpretability. 
For instance, a \Gls{GP} model can execute a process known as 
\Gls{ARD}~\citep{MacKay1996,Neal1996} by means of specialized kernel functions. 
The process yields an explanatory and sparse subset of features \citep{Wipf2007}. 
But, the high dimensionality of neuroimaging data has rendered \Gls{ARD} too expensive for this context.

In this study, we regard the induction of a classifier as the creation of a \emph{computational biomarker} for \emph{disease staging}.\footnote{
``\emph{Staging} is a method for measuring severity of specific, well-defined diseases. 
Staging defines discrete points in the course of individual diseases that are clinically detectable, 
reflect severity in terms of risk of death or residual impairment, and possess clinical significance 
for prognosis and choice of therapeutic modality'' \citep[p.1]{Gonnella1984}.
} 
Analogously, the creation of a probabilistic classifier results in a \emph{probabilistic (computational) biomarker}.
We propose to design probabilistic biomarkers 
via \Glspl{GP}. Also, \Glspl{GP} offer flexible means to carry out the \Gls{MKL} and tackle the challenges of high dimensionality. In this regard, we first investigate the conventional single kernel learning and postulate the advantages of using \Gls{GP} models rather than \Glspl{SVM} for biomarker induction. Then, 
we extend the basic scheme towards \Gls{MKL} and propose a new variation of \Gls{ARD}, namely, the \Gls{MKLARD}. It inherits the supervised feature selection characteristics of \Gls{ARD} and unites them with the automatic model selection capabilities of \Gls{MKL}. In contrast to sparse variants of \Glspl{GP}, such as the \Gls{RVM} and \Gls{SPGP}, \Gls{MKLARD} takes advantage of the full (non-sparse) \Glspl{GP}. 
Despite the use of full \Glspl{GP}, \Gls{MKLARD} is computationally feasible for the high-dimensional neuroimaging data. 
It has improved the performance of probabilistic biomarkers. Moreover, the proposed models are competitive with deep learning solutions under similar settings. 

The rest of the document is organized as follows. Section~\ref{sec:kernelMachines} reviews kernel machines and the basics of \Gls{MKL}. Section~\ref{sec:MKLARD} introduces \Gls{MKLARD}. Section~\ref{sec:Experiments} presents the experiments and results. Section~\ref{sec:discussion} compares and contrasts our work with previous methods including the deep learning solutions. Section~\ref{sec:Conclusion} concludes the study.

\section{Kernel Machines}\label{sec:kernelMachines}

Kernel machines translate the data points into an Euclidean feature space and aim to solve the machine learning problems in the new space. 
In this review section, we focus on 
supervised learning problems, for which 
the goal is to learn an appropriate function $y=f(\mathbf{x})$, where $\mathbf{x}=(x_1, x_2, ... , x_D)$ and $D$ is the number of dimensions, that maps inputs to outputs, given a data set $\mathcal{D} = \{(\mathbf{x}_i,y_i)\}$ where $i=1...N$. 

In the context of kernel machines, it is assumed that there is 
a function $\phi(\mathbf{x})$ that translates inputs\footnote{Input patterns are not necessarily in an inner product space. They can be any object of interest. To be able to use inner product as distance measure, we need to map them into an appropriate space \citep{Scholkopf2001}.} 
into a feature space and 
a kernel function $k(\mathbf{x}_i,\mathbf{x}_j) = \phi(\mathbf{x}_i) \phi(\mathbf{x}_j)^T$ implicitly achieves an embedding of observations into the feature space and yields the inner products. As a result, there is no need to explicitly compute the feature vectors and corresponding inner products. This is known as the \emph{kernel trick} \citep{Scholkopf2001,GPMLbook}. 
A kernel that induces \gls{PSD} Gram matrices is said to be \Gls{PSD}. A \Gls{PSD} kernel guarantees the uniqueness of a \emph{\gls{RKHS}} \citep{Scholkopf2001,GPMLbook}. The existence of \Gls{RKHS} allows us to use various distance measures that are accepted by Mercer's theorem \citep{Mercer1909}. 

In the next subsection, we review the well-known \Gls{SVM} and point out its disadvantages regarding model selection and probabilistic outputs. Then, we discuss the fundamentals of \Glspl{GP} and emphasize their capabilities for factor analysis and dimensionality reduction. We also review the special cases of \Gls{GP} models such as \Gls{RVM} and \Gls{SPGP} as they address the relevance determination and sparsity in different ways. 

\subsection{\acrlong{SVM}}\label{sec:SVM}
A typical \Gls{SVM} formulation for classification is 

\begin{equation}\label{eq_SVMprimal}
\begin{aligned}
\min_{w,b,\varepsilon}	\quad	&& \frac{1}{2}\mathbf{w}^T\mathbf{w} + C\sum_{i=1}^{N}\varepsilon_{i} \\
s.t.					\quad	&& y_{i}(\mathbf{w}^T\mathbf{x}_{i}+b) \geq 1 - \varepsilon_{i} \\
						\quad	&& \varepsilon_{i} \geq 0, \quad i \in \{1,\dots,N\},
\end{aligned}
\end{equation}
where $C > 0$ is the regularization parameter 
\citep{Vapnik1995,Chang2011}. This formulation is also known as $C$-Support Vector Classification ($C$-SVC). The dual of (\ref{eq_SVMprimal}) is written as 
\begin{equation}\label{eq_SVMdual}
\begin{aligned}
\min_{w,b,\varepsilon}	\quad	&& \frac{1}{2} {\boldsymbol\alpha}^T Q {\boldsymbol\alpha} - \mathbf{e}^T{\boldsymbol\alpha} \\
s.t.					\quad	&& \mathbf{y}^T {\boldsymbol\alpha} = 0, \\
						\quad	&& 0 \leq \alpha_{i} \leq C, \quad i \in \{1,\dots,N\},
\end{aligned}
\end{equation}
where $\mathbf{e} = [1,\dots,1]^T$, $Q$ is an $N \times N$ \gls{PSD} matrix, $Q_{ij}=y_{i}y_{j}k(\mathbf{x}_{i},\mathbf{x}_{j})$ and $k(\mathbf{x}_{i},\mathbf{x}_{j})$ is a kernel function 
\citep{Chang2011}. 
A typical choice is a \Gls{RBF} (\ref{eq_RBF}):
\begin{equation}\label{eq_RBF}
k_{RBF}(\mathbf{x}_{i},\mathbf{x}_{j}) = \text{exp}\left(-\gamma{\parallel}\mathbf{x}_{i}-\mathbf{x}_{j}{\parallel}^2\right).
\end{equation}
Once (\ref{eq_SVMdual}) is solved for $\boldsymbol{\alpha}$, 
the \Gls{SVM} decision function is 
\begin{equation}\label{eq_SVMdecfunc}
	f(\mathbf{x}_{*}) = \text{sgn}\left( \sum_{i=1}^{N}y_{i}\alpha_{i}k(\mathbf{x}_{i},\mathbf{x}_{*})+b \right),
\end{equation}
where $b$ is the intercept term. For $\alpha_i\neq0$, $\mathbf{x}_i$ is a \emph{support vector}. An SVM solution is sparse in the sense that the decision function depends only on 
support vectors. Also note that the output is a class label ($-1 \text{ or }+1$) \citep{Chang2011}. 

Training a typical \Gls{SVM} with an \Gls{RBF} kernel 
involves an exhaustive grid search for the values of hyperparameters 
$C$ and $\gamma$ \citep{Chang2011} . 
For a large number of parameter configurations, the grid search becomes prohibitively expensive. 
In practice, the search is conducted with respect to a finite set of \emph{discrete} parameter configurations predefined by user. Namely, the user roughly predicts the search space in which likely solutions exist. 
Then, for each configuration, (\ref{eq_SVMdual}) must be solved for $\boldsymbol{\alpha}$ in order to obtain (\ref{eq_SVMdecfunc}). Moreover, each decision function must be evaluated on a validation set in the search of the most plausible model. 
The validation set is essentially a portion of the valuable training data. 
Consequently, the solution to (\ref{eq_SVMdual}) is derived from a smaller training set. 
In this regard, \citet{Chang2011} suggest that the validation set be added to the training set, once the kernel parameters are optimized, and (\ref{eq_SVMdual}) be solved again in order to obtain the final model. However, such retraining does not necessarily mean that it will improve the model because \Gls{SVM} is a stable algorithm robust to perturbation of examples under mild conditions \citep{Scholkopf2001,Xu2009}. 

\subsubsection{Probabilistic Outputs}\label{sec:ProbOutput}

The \Gls{SVM} outputs can be calibrated so as to generate probabilistic information \citep{Platt1999}. 
A sigmoid function, $\lambda(af(\mathbf{x})+b)$, is fitted to the training set. Based on the solution of~\citet{Platt1999}, \citet{Lin2007} introduced a more robust algorithm with decent convergence properties.
Using class labels and decision values, 
$a$ and $b$ are estimated by 
maximizing the likelihood of training data~\citep{Wu2004,Chang2011}. Since such a training may cause the model to overfit to the data, a \emph{second-level} \gls{CV}
is required in order to obtain acceptable\footnote{If the decision values are clustered around $\pm1$, 
probability estimates can be inaccurate. The secondary \Gls{CV} reduces overfitting and 
induces a smoother distribution of decision values. 
} 
decision values \citep{Wu2004,Chang2011}. Unfortunately, it causes a major overhead regarding the model selection due to nested validation procedures. 
%
%

\subsection{Gaussian Processes for Regression}\label{sec:GPR}
\Glspl{GP} enable us to do inference in the function space. Explicitly, given $\mathcal{D}$,
an $N$-dimensional random vector $\mathbf{f}$ represents the function values induced by inputs. We wish to infer the best possible $\mathbf{f}$ in order to match to the target vector $\mathbf{y}$. 


A \Gls{GP} is specified by a mean function $m(\mathbf{x}_i)$ and a covariance function $k(\mathbf{x}_i,\mathbf{x}_j)$. 
Given an observation $\mathbf{x}_i$ as input, the value of $f(\mathbf{x}_i)$ is a sample from the process~\citep{GPMLbook}: 
\begin{equation}\label{eq_gpDef}
\begin{aligned}
f(\mathbf{x}_i) &\sim \mathcal{GP}(m(\mathbf{x}_i),k(\mathbf{x}_i,\mathbf{x}_j)) \text{, where} \\
m(\mathbf{x}_i) &= \mathbb{E}[f(\mathbf{x}_i)] \\
k(\mathbf{x}_i,\mathbf{x}_j) &= \mathbb{E}[(f(\mathbf{x}_i)-m(\mathbf{x}_i))(f(\mathbf{x}_j)-m(\mathbf{x}_j))] \text{.}
\end{aligned}
\end{equation}
(\ref{eq_gpDef}) indicates that $f(\mathbf{x}_i)$ and $f(\mathbf{x}_j)$ are jointly Gaussian. For $j=1...N$, $f(\mathbf{x}_i)$ depends on other observations, as well. 
In this regard, a \Gls{GP} model that uses all observations is considered to be \emph{full} (non-sparse) and special cases of \Gls{GP} models that provide sparse approximations (Section~\ref{sec:RVM} and Section~\ref{sec:SPGP}) to full \Glspl{GP} exist \citep{Tipping2001,Candela2004,GPMLbook,Snelson2006a,
Snelson2006b}.


In \Gls{GP} learning, inner products are computed with respect to a covariance 
matrix $\Sigma_f$ implied by the features: 
$k(\mathbf{x}_i,\mathbf{x}_j) = \phi(\mathbf{x}_i) \Sigma_f \phi(\mathbf{x}_j)^T$ \citep{GPMLbook}. 
Hence, in \Gls{GP} terminology, 
a kernel is a covariance function that estimates the covariance of two latent variables 
$f(\mathbf{x}_i)$ and $f(\mathbf{x}_j)$ in terms of input vectors $\mathbf{x}_i$ and $\mathbf{x}_j$ \citep{GPMLbook}. 
A simple covariance function that depends on an inner product of the input vectors is $k_{LIN}(\mathbf{x}_{i},\mathbf{x}_{j})={\sigma^2_f}\left(\mathbf{x}_{i}{\cdot}\mathbf{x}_{j}^{T}\right)$, where $\sigma_f$ is the scale parameter. A more widely used one is the \emph{\gls{SE}} covariance function: 
\begin{equation}\label{eq_covSEiso}
k_{SE}(\mathbf{x}_i,\mathbf{x}_j) = {\sigma^2_f}\text{exp}\left(-\frac{{\parallel}\mathbf{x}_i-\mathbf{x}_j{\parallel}^2}{2{\ell}^2}\right),
\end{equation} 
where $\ell$ is the bandwidth parameter. Another example is 
the \emph{\gls{NN}} covariance function:
\begin{equation}\label{eq_covNNone}
k_{NN}(\mathbf{x}_i,\mathbf{x}_j) = {\sigma^2_f}\text{sin}^{-1}\left( \frac{2\mathbf{\tilde{x}}_i^T\Sigma\mathbf{\tilde{x}}_j}{\sqrt{(1+2\mathbf{\tilde{x}}_i^T\Sigma\mathbf{\tilde{x}}_i)(1+2\mathbf{\tilde{x}}_j^T\Sigma\mathbf{\tilde{x}}_j)}} \right), 
\end{equation}
where $\tilde{\mathbf{x}}=(1,\mathbf{x})^T$ is an augmented input vector and $\Sigma$ is a covariance matrix 
for \emph{input-to-hidden} weights: $\mathbf{w}\sim\mathcal{N}(\mathbf{0},\Sigma)$ \citep{Neal1996,Williams1998,GPMLbook}. 
A~\Gls{GP} with the \Gls{NN} covariance function 
emulates a \Gls{NN} with a single hidden layer~\citep{GPMLbook}.

\subsubsection{Learning of hyperparameters}\label{sec:learningOfHypers}
A \emph{\Gls{GP}-prior}, e.g., $\mathbf{f} \sim \mathcal{N}(\mathbf{0}, K)$, is a prior distribution over the latent variables. The covariance matrix $K$ is dictated by the covariance function $k(\cdot,\cdot)$. Once combined with the likelihood associated with observed data, the \Gls{GP}-prior leads to a \emph{\Gls{GP}-posterior} in a function space. This Bayesian treatment promotes the smoothness of predictive functions~\citep{Williams1998} and the prior has an effect analogous to the quadratic penalty term used in maximum-likelihood procedures~\citep{GPMLbook}. 

The impact of the covariance function on the information processing capabilities of \Glspl{GP} is larger for small to medium-sized datasets~\citep{Duvenaud2011}. As the covariance function communicates our beliefs about the functions to be estimated, it should suit the problem at hand well. Otherwise, the likelihood term may fail to compensate for the inappropriate prior specification unless the dataset is large enough. 

Many covariance functions have adjustable parameters, such as $\ell$ and $\sigma_f$ in (\ref{eq_covSEiso}). In this regard, learning in \Glspl{GP} is equivalent to finding suitable parameters for the covariance function. Given the target vector $\mathbf{y}$ and the matrix $X$ that consists of training instances, this is accomplished by maximizing the log marginal likelihood function:
\begin{equation}\label{eq_logmarglik}
\text{log }p(\mathbf{y}|X) = \underbrace{ -\frac{1}{2}\mathbf{y}^T(K+\sigma^2_nI)^{-1}\mathbf{y} }_{\text{data fit term}} 
\underbrace{ -\frac{1}{2}\text{log }|K+\sigma^2_nI| }_{\text{complexity term}} \underbrace{ -\frac{N}{2}\text{log }2{\pi} }_{\text{a constant}},
\end{equation}
where $\sigma_n$ is due to the Gaussian noise model, $y_i = f_i + \varepsilon$ 
and $\varepsilon \sim \mathcal{N}(0,\sigma^2_n)$. 
Note that a large number of hyperparameters can be automatically approximated by maximizing (\ref{eq_logmarglik}) via \emph{continuous} optimization. 
(\ref{eq_logmarglik}) is solved just once and no validation set is required, 
which is desirable when the sample size is small.  

\subsubsection{Predictions}\label{sec:predictions}
\Gls{GP} regression yields a predictive Gaussian distribution (\ref{eq_predDist}), from which a prediction $f_* = f(\mathbf{x}_*)$ 
is sampled, given the training instances $X$, target vector $\mathbf{y}$ and test input $\mathbf{x}_*$. 
\begin{equation}\label{eq_predDist}
f_*|X,\mathbf{y},\mathbf{x}_* \sim \mathcal{N}(\bar{f}_*,\mathbb{V}[f_*]) \text{, where}
\end{equation}
\begin{equation}\label{eq_predMean}
\bar{f}_* = \mathbf{k}_*^T(K+\sigma^2_nI)^{-1}\mathbf{y}
\end{equation}
\begin{equation}\label{eq_predVar}
\mathbb{V}[f_*] = k(\mathbf{x}_*,\mathbf{x}_*) - \mathbf{k}_*^T(K+\sigma^2_nI)^{-1}\mathbf{y}
\end{equation}
and $\mathbf{k}_*$ is a vector of covariances between the test input $\mathbf{x}_*$ and the training instances. (\ref{eq_predMean}) gives the mean prediction $\bar{f}_*$, which is the \emph{empirical risk minimizer} for any symmetric loss function~\citep{GPMLbook}. 
(\ref{eq_predVar}) yields the predictive variance. 
Notice that the information extracted from the observations reduces the prior variance. 

\subsection{Gaussian Processes for Classification}\label{sec:GPC}
\Gls{LR} is a well-known binary classifier, for which a sigmoid function~(\ref{eq_logit}) assigns the class probabilities of a given input $\mathbf{x}_*$ based on the associated function value $f_*$: 
\begin{equation}\label{eq_logit}
p(y_*=+1|\mathbf{x}_*) = \lambda(f_*) = \frac{1}{1+\text{exp}(-f_*)}.
\end{equation}

\Gls{GP} classification generalizes\footnote{A squashing function with a domain $(-\infty, +\infty$) and a range $[0,1]$ ensures a valid probabilistic interpretation. Another typical choice is the cumulative density function of a standard normal distribution: $\Phi(f_*)=\int_{-\infty}^{f_*}\mathcal{N}(x|0,1)dx$ \citep{GPMLbook}.}
the idea and turns $\lambda(f_*)$ into a stochastic function, which implies a distribution over predictive probabilities. 
Once the \emph{nuisance} parameters are integrated out, we obtain the \emph{averaged predictive probability} via (\ref{eq_avgPredProb}) and (\ref{eq_predDistClass})~\citep{GPMLbook}. 
\begin{equation}\label{eq_avgPredProb}
\bar{\lambda}(f_*) = \int \lambda(f_*)p(f_*|X,\mathbf{y},\mathbf{x}_*)df_*, \text{ where}
\end{equation}
\begin{equation}\label{eq_predDistClass}
p(f_*|X,\mathbf{y},\mathbf{x}_*) = \int p(f_*|X,\mathbf{x}_*,\mathbf{f})p(\mathbf{f}|X,\mathbf{y})d\mathbf{f}
\end{equation}
%

Due to discrete targets, 
the likelihood term is not Gaussian. Neither is the posterior distribution over the functions, $p(\mathbf{f}|X,\mathbf{y})$. 
Thus, the exact computation in (\ref{eq_predDistClass}) is intractable and we resort to approximation methods. A comprehensive overview of algorithms for approximate inference in \Glspl{GP} for probabilistic binary classification is provided~\citep{Nickisch2008}. 
We consider two of these solutions: \Gls{LA} and \Gls{EP}.

\subsubsection{\acrlong{LA}}\label{sec:LA}
\Gls{LA} proposes a second order Taylor expansion around the mode of the posterior distribution and 
replaces $p(\mathbf{f}|X,\mathbf{y})$ with a Gaussian approximation $q(\mathbf{f}|X,\mathbf{y})$ 
centered at the \emph{\Gls{MAP}} estimate of $p(\mathbf{f}|X,\mathbf{y})$. 
The covariance matrix of the approximation is given by the inverse of the Hessian 
of negative log-posterior \citep{Williams1998,GPMLbook,Nickisch2008}. 

\Gls{LA} is a local method that exploits the properties, such as derivatives, of the posterior at a 
particular location only, e.g., at the posterior distribution's mode \citep{Nickisch2008}. Owing to its simplicity, 
\Gls{LA} is very fast; however, this method may lead to substantial underestimations of the mean and covariance, 
especially, in high-dimensional spaces 
because the mode and mean can be far from each other \citep{Kuss2005,Nickisch2008}. 

\subsubsection{\acrlong{EP}}\label{sec:EP}
The \Gls{EP} algorithm can perform approximate inference 
more accurately than other methods like Monte Carlo, \Gls{LA} and \Gls{VB} \citep{Minka2001PhD}. 
%
%
As a result, it is heavily used for \Gls{GP} learning. \Gls{EP} is a global method in the sense that the likelihood $p(\mathbf{y}|\mathbf{f})$ in (\ref{eq_EPposterior}) is approximated by means of many local approximations \citep{GPMLbook,Nickisch2008}. That is, $p(y_i|f_i)$ is approximated by a Gaussian function $\tilde{Z}_i\mathcal{N}(f_i|\tilde{\mu}_i,\tilde{\sigma}_i^2)$, 
where $\tilde{Z}_i$, $\tilde{\mu}_i$ and $\tilde{\sigma}_i^2$ are \emph{site parameters} \citep{GPMLbook}.

\begin{equation}\label{eq_EPposterior}
\begin{aligned}
	p(\mathbf{f}|X,\mathbf{y})	&= \frac{1}{Z}p(\mathbf{f}|X)p(\mathbf{y}|\mathbf{f})= \frac{1}{Z}p(\mathbf{f}|X)\prod_{i=1}^{N}p(y_i|f_i), \text{ where} \\
	Z &= \int p(\mathbf{f}|X)p(\mathbf{y}|\mathbf{f})d\mathbf{f} = p(\mathbf{y}|X). 
\end{aligned}
\end{equation}

\Gls{EP} is an iterative algorithm. 
Iterative minimization of local divergences results in a small global divergence~\citep{Nickisch2008}. 
As a result, \Gls{EP} algorithm delivers accurate marginals, reliable class probabilities and 
faithful model selection~\citep{Nickisch2008}. 
However, 
the convergence of EP is not generally guaranteed~\citep{Nickisch2008}. 

\subsection{\acrlong{ARD} and Supervised Factor Analysis}\label{sec:factorAnalysis}
%

A multipurpose covariance function can be written with respect to a symmetric matrix $M$ as in (\ref{eq_covSEisoGeneral}) \citep{GPMLbook}. 
\begin{equation}\label{eq_covSEisoGeneral}
k_{multi}(\mathbf{x}_i,\mathbf{x}_j) = {\sigma^2_f}\text{exp}\left(-\frac{(\mathbf{x}_i-\mathbf{x}_j)M(\mathbf{x}_i-\mathbf{x}_j)^{T}}{2}\right). 
\end{equation}
%

Possible choices for $M$ are $M_1 = \ell^{-2}I, M_2 = \text{diag}\left( \boldsymbol{\ell} \right)^{-2}$ and $M_3 = \Lambda \Lambda^{T} + M_1,$ 
where $\boldsymbol{\ell} = [\ell_1, \ell_2,\cdots,\ell_D]$ is a vector of bandwidth parameters for each dimension, $\Lambda$ is a $D \times k$ projection 
matrix, and typically $k \ll D$. Depending on the choice of $M_1$ or $M_2$, (\ref{eq_covSEisoGeneral}) 
can perform as a standard Gaussian-shaped kernel or implement \Gls{ARD}~\citep{MacKay1996,Neal1996,GPMLbook}, respectively. 
As a result of \Gls{ARD}, irrelevant features are effectively turned off by selecting large bandwidths for them. 
The \Gls{ARD} process yields an explanatory and sparse subset of features \citep{Neal1996,GPMLbook,Wipf2007}. 
But, its cost is $O(N^2)$ per hyperparameter~\citep{GPMLbook}, which makes it computationally expensive for high-dimensional data.  

While \Gls{ARD} removes the superfluous dimensions from the inference, $M_3$ allows us to exploit the rich covariances between the dimensions \citep{Snelson2006b}, and enables a \emph{factor analysis} \citep{GPMLbook}. 
The goal of the factor analysis is to find a small number of highly relevant directions in the input space, which is analogous to \Gls{PCA}. However, \Gls{PCA} is an unsupervised filter and it does not have access to the function classes, from which we sample our predictor $f(X)$. 
All that is known to the \Gls{PCA} algorithm is the set of observations $X$. On the other hand, a \Gls{GP} model can carry out the factor analysis in a supervised manner since 
all hyperparameters are jointly optimized with respect to the likelihood of data \citep{GPMLbook,Snelson2006b}. 

\subsection{\acrlong{RVM}}\label{sec:RVM}
\Gls{RVM} is a probabilistic kernel machine with sparsity properties that are analogous to \Gls{SVM}`s~\citep{Tipping2001,Bishop2003}. Observations that are assigned \emph{non-zero} weights via a Bayesian analysis are called \emph{relevance vectors}.\footnote{Despite the approach is reminiscent of \Gls{ARD}~\citep{MacKay1996,Neal1996,GPMLbook}, one should note the difference being made between the observations and features.} 
\Gls{RVM} is competitive with \Gls{SVM} in terms of generalization performance; however, it achieves higher sparsity in its solution 
\citep{Tipping2001,Bishop2003}. 
Given that the number of support vectors typically grows linearly with the training set, \Gls{RVM} is advantageous in practice \citep{Tipping2001,Bishop2003}. 

\Gls{RVM} is essentially a special \Gls{GP} model that can be constructed by the covariance function (\ref{eq_covRVM}) \citep{Tipping2001,Candela2004,GPMLbook}: 
\begin{equation}\label{eq_covRVM}
k_{RVM}(\mathbf{x}_i,\mathbf{x}_j)=\sum_{n=1}^{N}\frac{1}{\alpha_n}\psi_{n}(\mathbf{x}_i) \psi_{n}(\mathbf{x}_j)^T, 
\end{equation}
where $\boldsymbol\alpha$ is a vector of hyperparameters and $\psi_{n}(\mathbf{x}_i)$ is a basis function (\ref{eq_RVMbasis}) centered on the observation $\mathbf{x}_n$ \citep{GPMLbook,Tipping2001}. 
\begin{equation}\label{eq_RVMbasis}
\psi_{n}(\mathbf{x}_i)= \text{exp}\left(-\frac{{\parallel}\mathbf{x}_i-\mathbf{x}_n{\parallel}^2}{2{\ell}^2}\right),
\end{equation}
where $\ell$ is a bandwidth parameter. 
For a large $\alpha_n$, the basis function centered on $\mathbf{x}_n$ is effectively removed from the kernel computation. Thus, $\mathbf{x}_n$ is not a relevance vector. 

Selecting a subset of observations as relevance vectors, \Gls{RVM} achieves a sparse approximation to a full \Gls{GP} \citep{Tipping2001,Candela2004,GPMLbook}. 
However, a \Gls{GP}-prior induced by (\ref{eq_covRVM}) 
depends on the training data. Given a pair of inputs $\mathbf{x}_i$ and $\mathbf{x}_j$, (\ref{eq_covRVM}) 
passes over all observations and relates the prior to evidence, which is at odds with the Bayesian formalism \citep{GPMLbook}. 
Also, the selection of the relevance vectors for the estimation of the full \Gls{GP} likelihood may undermine the optimality of hyperparameters, particularly those found as a result of an \Gls{ARD} process \citep{Snelson2006a}. 

\subsection{\acrlong{SPGP}}\label{sec:SPGP}
\Gls{SPGP} \citep{Snelson2006b} exploits the supervised factor analysis along with \emph{pseudo-inputs}.\footnote{
$k(\mathbf{x}_i,\mathbf{x}_p) = {\sigma^2_f}\text{exp}\left(-\frac{(\mathbf{x}_i\Lambda-\mathbf{x}_p)(\mathbf{x}_i\Lambda-\mathbf{x}_p)^{T}}{2}\right)$, 
where $\mathbf{x}_p$ is a pseudo-input in the low dimensional space implied by the projection. 
} 
The pseudo-inputs are engineered entities. They do not belong to the training data. Instead, they are learned from it and used to approximate the full \Gls{GP} covariance matrix $K$ \citep{Snelson2006a,Snelson2006b}. The number of pseudo-inputs is typically much smaller than the number of training inputs: $P \ll N$ \citep{Snelson2006a,Snelson2006b}. Therefore, the \Gls{SPGP} is a sparse approximation to the full \Gls{GP} \citep{Snelson2006a,Snelson2006b}. 

A \Gls{SPGP} is parameterized by 
\begin{equation}\label{eq_SPGPparam}
\underbrace{Pk}_{\text{pseudo-input parameters}} + \underbrace{Dk}_{\text{projection parameters}} + \underbrace{2}_{\sigma_f , \ell} 
\end{equation}
hyperparameters \citep{Snelson2006b}. (\ref{eq_SPGPparam}) indicates that the scale of optimization and quality of approximation are determined by $P$ and $k$, given the $D$-dimensional data. In practice, small values for $P$ and $k$ may suffice for many problems~\citep{Snelson2006b}. However, the dimensionality of neuroimaging data complicates the optimization, even if only one pseudo-input and one factor are deemed acceptable. Also note that the solution would probably provide a very poor approximation to the full \Gls{GP}. 
In this regard, we seek ways to retain the benefits of the full \Glspl{GP} while discovering the most relevant dimensions of data through kernels in an efficient manner and avoid solving a large scale optimization problem that results in a poor approximation. 

\subsection{\acrlong{MKL}}\label{sec:MKL}
A kernel function $k(\cdot,\cdot)$ encodes our beliefs about the predictive function that we aim to learn from data. 
But, a single kernel 
restricts us to a certain class of functions. To relax the restriction, we can specify many kernels as in (\ref{eq_SimpleMKL}) and 
simultaneously learn their properties 
via \Gls{MKL}. 

\begin{equation}\label{eq_SimpleMKL}
k(\mathbf{x}_i,\mathbf{x}_j) = \sum_{m=1}^{M} \beta_mk_{m}(\mathbf{x}_i,\mathbf{x}_j), 
\end{equation}
where $\mathbf{x}_i$ and $\mathbf{x}_j$ are the input vectors, $k_{m}(\cdot,\cdot)$ is the $m$-th 
\emph{basis} kernel and $\beta_m$ is the weight associated with 
it \citep{SimpleMKL}. 

\Gls{MKL} is a generalization of the single kernel learning. 
Given a single data representation 
and several basis kernels, each of which is defined on the same input space, an \Gls{MKL} solution delivers an optimal combination of the kernels. 
As a result, \Gls{MKL} 
allows for \emph{automatic model selection} and insights into the learning problem at hand \citep{Bach2004,Hinrichs2012}. 
In addition, \Gls{MKL} enables a principled way of integrating feature representations obtained from different data sources or modalities \citep{Sonnenburg2006,SimpleMKL,Gonen2011,Gonen2012}. 
Under such a multimodal scenario, each kernel computes the similarities with respect to a given modality and they are combined into a final similarity measure. Data integration improves the performances and interpretability of models~\citep{Lanckriet2004,Sonnenburg2006,SimpleMKL,Hinrichs2009,Zhang2011,Hinrichs2012,Young2013a,Young2013b}. 

\subsubsection{Regularization of Mixing Weights}
The optimal combination of kernels is achieved via the mixing weights $\beta_1 \dots \beta_m$, which must be tuned and regularized. The goal of regularization is to control the model complexity and prevent overfitting due to the sampling error. A common practice is to use a \emph{norm} of the $M$-dimensional parameter vector as a regularizer. In general, $L_p$ norm is $||\boldsymbol\beta||_p = (\sum_{m=1}^{M}|\beta_m|^{p})^{1/p}$. 
The most common forms are the $L_1$ and $L_2$ norms of the weights. 

\paragraph{Lasso, Group Lasso and Multiple Kernel Learning:}\label{sec:groupLasso}
Regularization by the $L_1$ norm is also known as \emph{Lasso} and it leads to sparse solutions through weight vectors which contain many zeros \citep{Tibshirani1996,Zhao2006}. Lasso is used for variable selection in \emph{least square regression}; however, the consistency of Lasso depends on whether it can recover the sparsity pattern, or not, when the true model is one with a sparse vector \citep{Zhao2006,Bach2008}. In the presence of lowly correlated features, Lasso is consistent, but, strong correlations between features hinder the consistency of Lasso \citep{Zhao2006,Bach2008}. 

\emph{Group Lasso} \citep{Bach2008} assumes that the correlated features are grouped together
and it defines a regularizer with respect to the $L_2$ norms of the weights corresponding to groups of features. (\ref{eq:groupLasso}) depicts the Group Lasso optimization problem based on the squared loss. 

\begin{equation}\label{eq:groupLasso}
\min_{w,b} \frac{1}{2N} \left [
\sum_{i=1}^{N} (y_i - (\mathbf{w}^T\mathbf{x}_{i}+b))^2 + \lambda \sum_{m=1}^{M} d_m ||\mathbf{w}_m||_2 
\right ]
\end{equation}
where $\mathbf{d}=(d_1, d_2, ... d_M)^T$ is a vector of \emph{strictly positive} weights, $\mathbf{w} \in \mathbb{R}^D$, and $\mathbf{w}_m$ represents the weights of the $m$-th group \citep{Bach2008}. As a result of the Group Lasso, all the weights in a group are driven towards zeros all together; however, the growth of group sizes from one to larger numbers leads to weaker consistency results \citep{Bach2008}. 

The Group Lasso also allows for the replacement of the sum of $L_2$ (Euclidean) norms with the sum of Hilbertian norms defined over \Glspl{RKHS}~\citep{Bach2004,Bach2008}. Such a transformation enables the learning of the best convex combination of a set of \Gls{PSD} kernels, which is equivalent to \Gls{MKL} \citep{Bach2004,Bach2008}. In this setting, Group Lasso becomes \emph{nonparametric} and it leads to sparse combinations of kernel functions of separate random variables \citep{Bach2008}. 

Despite the sparsity benefits, $L_1$ regularization may significantly degrade the generalization performance for large-scale kernel combinations~\citep{Cortes2009,Gonen2012}. On the other hand, $L_2$ regularization never degrades the performance and, in contrast, it achieves significant improvements~\citep{Cortes2009,Gonen2012}.

\subsubsection{Solutions for Multiple Kernel Learning}\label{sec:MKLsolutions}
\citet{Lanckriet2004} introduced the \Gls{MKL} problem as a \Gls{SDP}. The \Gls{SDP} is solved for a linear combination of fixed kernels. If the mixing weights are restricted to be positive, the \Gls{SDP} reduces to a \Gls{QCQP} \citep{Lanckriet2004}. However, it rapidly becomes intractable as the training set or scale of kernel combination grows large \citep{SimpleMKL}. Also, the problem is convex but non-smooth \citep{SimpleMKL}. \citet{Bach2004} introduced a smoothed version of the problem via \Gls{SOCP} and \Gls{SMO}, which 
addressed the scalability issues of \citet{SimpleMKL} for medium-scale problems. 

\citet{Sonnenburg2006} revised the MKL problem as a \Gls{SILP}. This solution has computational advantages since it iteratively uses a typical \Gls{SVM} with a single kernel~\citep{Sonnenburg2006,SimpleMKL,Chang2011}. 
Thus, it is applicable to large datasets. But, the linear problem includes constraints, the number of which may increase along with iterations \citep{Sonnenburg2006,SimpleMKL}. 
A solution may be delayed due to new constraints~\citep{Sonnenburg2006}. 

\emph{SimpleMKL}~\citep{SimpleMKL} is an efficient algorithm for \Gls{MKL}. The mixing weights are determined via a gradient descent that wraps around a classical SVM solver~\citep{SimpleMKL}. 
In terms of classification performance, SimpleMKL and \Gls{SILP} are competitive with each other \citep{SimpleMKL}. 
However, 
SimpleMKL has empirically demonstrated better convergence properties. Thus, it is significantly more efficient than \Gls{SILP} \citep{SimpleMKL}. 
Nevertheless, 
it still aims to find an optimal linear combination of the basis kernels, properties of which are fixed~\citep{SimpleMKL}. 
Given the objective of \Gls{MKL} for automatic model selection, the requirement of \emph{pre-specifying} the kernel properties, such as the bandwidth of an \Gls{RBF} or the degree of a polynomial kernel, is a step backwards, even though SimpleMKL extends the capabilities of \Glspl{SVM}. 

The Bayesian framework of \Glspl{GP} offers more flexibility for parameter optimization. 
The fact that many hyperparameters can be \emph{automatically} tuned via \Gls{GP} learning 
is a factor in the adoption of \Glspl{GP} under various \Gls{MKL} scenarios. 
A notable example is the \emph{Bayesian Localized \Gls{MKL}} \citep{Christoudias2009}, which states that the set of discriminative features depends on the locality of observations. Thus, multiple local views of data should be preferred to a single (global) view. The local views are estimated by clustering the data in the joint feature space; each view is processed by a covariance function and the final covariance metric is obtained by a linear combination of those \citep{Christoudias2009}. 
In this setting, each covariance function is a product of two functions: 
a parametric covariance function to compute the similarities over the feature space, and 
a non-parametric covariance function to represent the confidence of similarities in each view. With this Bayesian approach to \Gls{MKL}, local feature importance can be learned, and the locally weighted views of data improve the classification performance \citep{Christoudias2009}. Also, the method is robust to \emph{over-clustering}; a rough estimate of the number of clusters in data is sufficient \citep{Christoudias2009}. 

\Glspl{MGP} \citep{Archambeau2011} is an \Gls{MKL} framework that aims at obtaining convex combinations of covariance functions associated with different modalities in the hopes of better explaining data and improving the generalization performances of models. It also addresses all $L_p$ norms of the mixing weights at once and aims to determine in the light of data whether sparsity in kernel combination is appropriate or not \citep{Archambeau2011}. 
However, a \Gls{MGP} model essentially corresponds to multiple independent \emph{non-\Glspl{GP}}, from which a \Gls{GP} can be recovered by mathematical manipulation \citep{Archambeau2011}. 
After all, a sum of multiple latent function values yields a prediction \citep{Archambeau2011}. 

\Glspl{GP} are typically used for single task prediction. Namely, the goal is to predict a single output, given a set of features. On the other hand, \Gls{MTL} leverages a shared representation to extract specific information for multiple tasks \citep{Caruana1998}. 
\citet{Melkumyan2011} used \Glspl{GP} for \Gls{MTL}. In order to model the dependencies between multiple outputs, they adopted \Gls{MKL} and used various covariance functions for different tasks. The use of \Gls{MKL} enabled the \Gls{GP} models to estimate the target functions more accurately as well as to better capture the dependencies between tasks \citep{Melkumyan2011}. 

\section{\acrlong{MKL} for Automatic Subpsace Relevance Determination}\label{sec:MKLARD}


\Gls{GP} models execute \Gls{ARD} via covariance functions and their hyperparameters. As a result, the most relevant dimensions are determined in a supervised fashion. But, the \Gls{ARD} achieved by $M_2$ in  (\ref{eq_covSEisoGeneral}) is prohibitive for the high-dimensional neuroimaging data. Thus, we shy away from the conventional approach, but still want to exploit its elegance. 

Given the flexibility of the \Gls{GP} framework and the benefits of \Gls{MKL}, we investigate a new variation of \Gls{ARD} through multiple kernels. Our goal is to enable the expressiveness of \Gls{ARD} in high-dimensional settings. 
To this end, 
we propose an \Gls{ARD} procedure with respect to coarse groups, namely, bags of features rather than considering individual features. As a result, we will dramatically reduce the number of hyperparameters, obtain a tractable \Gls{ARD} solution for the analysis of neuroimaging data, and achieve sparsity at the level of feature bags. We call the new procedure \Gls{MKLARD}. It will be parameterized by the hyperparameters of the covariance functions and hence inherit the elegance of \Gls{ARD} as well as its characteristics for supervised feature selection. 


It takes two basic steps to set up \Gls{MKLARD}: 
\begin{inparaenum}[i)]
	\item define the feature bags and 
	\item assign the basis kernels accordingly. 
\end{inparaenum}
Then, each bag, which corresponds to a \emph{subspace}, will be processed by a basis kernel and 
the information extracted from each will be weighted and combined into a final similarity measure. 
An \Gls{MKLARD} problem can be specified with (\ref{eq_MKLoverSpaces}). 

\begin{equation}\label{eq_MKLoverSpaces}
k(\mathbf{x}_i,\mathbf{x}_j) = \sum_{s=1}^{S} \beta_sk_{s}(\mathbf{x}_{is},\mathbf{x}_{js}),
\end{equation}
where $\mathbf{x}_i$ and $\mathbf{x}_j$ are the input vectors, $s$ is the subspace indentifier, $k_{s}(\cdot,\cdot)$ is a \emph{basis} kernel and $\beta_s$ is the weight associated with the subspace. A subvector $\mathbf{x}_{is}$ denotes the part of the input $\mathbf{x}_{i}$ that lies in the subspace $s$, where $s \in \left\{1,2,...,S \right\}$. Despite their resemblance, (\ref{eq_SimpleMKL}) and (\ref{eq_MKLoverSpaces}) are semantically different. 
(\ref{eq_MKLoverSpaces}) is a combination of multiple functions, each of which processes certain parts ($\mathbf{x}_{is}$ and $\mathbf{x}_{js}$) of inputs, even in the case of single modality~\citep{Williams1998}. 
For $\beta_s\neq0$, 
$\mathbf{x}_{is}$ and $\mathbf{x}_{js}$ are the relevant portions of inputs. 
Otherwise, $\mathbf{x}_{is}$ and $\mathbf{x}_{js}$ are effectively pruned away from the inference. 

In summary, we consider a global view of data and a single \Gls{GP} which is specified with multiple covariance functions and aimed at single task learning. Despite the global view, the subtlety of \Gls{MKLARD} should be noted. We wish to exploit the locality information in order to better capture the spatial patterns of brain atrophy. In this regard, we consider the portions of high-dimensional inputs, instead of the clusters obtained by an analysis of the joint feature space (Figure~\ref{figureLocalViewVSSubspace}). 

\begin{figure}[!h]
	\centering
	\[
	X=
	\begin{tikzpicture}[baseline,decoration=brace]
		\matrix (m) [table] {
			x_{1,1} \& x_{1,2} \& x_{1,3} \& \cdots \& x_{1,d} \& \cdots \& x_{1,D} \\
			\vdots \& \vdots \& \ddots \& \cdots \& \vdots \& \vdots \& \vdots \\
			\,\,\,\,\,x_{k,1}\,\,\,\, \& x_{k,2} \& x_{k,3} \& \cdots \& x_{k,d} \& \vdots \& x_{k,D} \\
			\vdots \& \vdots \& \vdots \& \vdots \& \ddots \& \vdots \& \vdots \\
			x_{N,1} \& x_{N,2} \& x_{N,3} \& \cdots \& x_{N,d}\& \vdots \& x_{N,D} \\
};
    \draw[decorate,transform canvas={xshift=-1.2em},thick] (m-3-1.south west) -- node[left=2pt] {local view} (m-1-1.north west);
    \draw[decorate,transform canvas={xshift=23.0em},thick] (m-1-1.north east) -- node[right=2pt] {global view} (m-5-1.south east);
    \draw[decorate,transform canvas={yshift=0.5em},thick] (m-1-1.north west) -- node[above=2pt] {subspace} (m-1-5.north east);
\end{tikzpicture}
\]
	\caption[Comparison of the local and global views of data]{Comparison of the local and global views of data. Given an $N \times D$ data matrix $X$, a local view corresponds to a cluster of observations with size $k$, whereas the global view embraces the entire dataset \citep{Christoudias2009}. On the other hand, a \emph{portion} of inputs corresponds to a $d$-dimensional subspace of the input space.}
	\label{figureLocalViewVSSubspace}
\end{figure}

\subsection{Composite Kernels for \Gls{MKLARD}}\label{sec:compKernels}
Adopting $k_{LIN}(\cdot,\cdot)$, $k_{SE}(\cdot,\cdot)$ and $k_{NN}(\cdot,\cdot)$ 
as \emph{basis} kernels, we define three composite kernels as follows: 

\begin{equation}\label{eq_covLINComp}
k_{LINcomp}(\mathbf{x}_i,\mathbf{x}_j) = \sum_{s=1}^{S} k_{LIN}(\mathbf{x}_{is},\mathbf{x}_{js}),
\end{equation}

\begin{equation}\label{eq_covSEisoComp}
k_{SEcomp}(\mathbf{x}_i,\mathbf{x}_j) = \sum_{s=1}^{S} k_{SE}(\mathbf{x}_{is},\mathbf{x}_{js}),
\end{equation}

\begin{equation}\label{eq_covNNoneComp}
k_{NNcomp}(\mathbf{x}_i,\mathbf{x}_j) = \sum_{s=1}^{S} k_{NN}(\mathbf{x}_{is},\mathbf{x}_{js}),
\end{equation}
where each {subspace} is assigned a local kernel. These kernel combinations indicate a conic sum where the mixing weights are restricted to be \emph{non-negative}: $\beta_s = \sigma^2_{f_{s}}$. Also, we optimize with respect to $\sigma_{f_{s}}$. Figure~\ref{fig:GPComplexitySurface} shows a surface obtained by the complexity term in (\ref{eq_logmarglik}). Clearly, the parameters are driven towards zero, which leads to sparse solutions as in the case of the nonparametric Group Lasso. 

\begin{figure}[h!]
	\centering
	\includegraphics[width=0.65\linewidth]{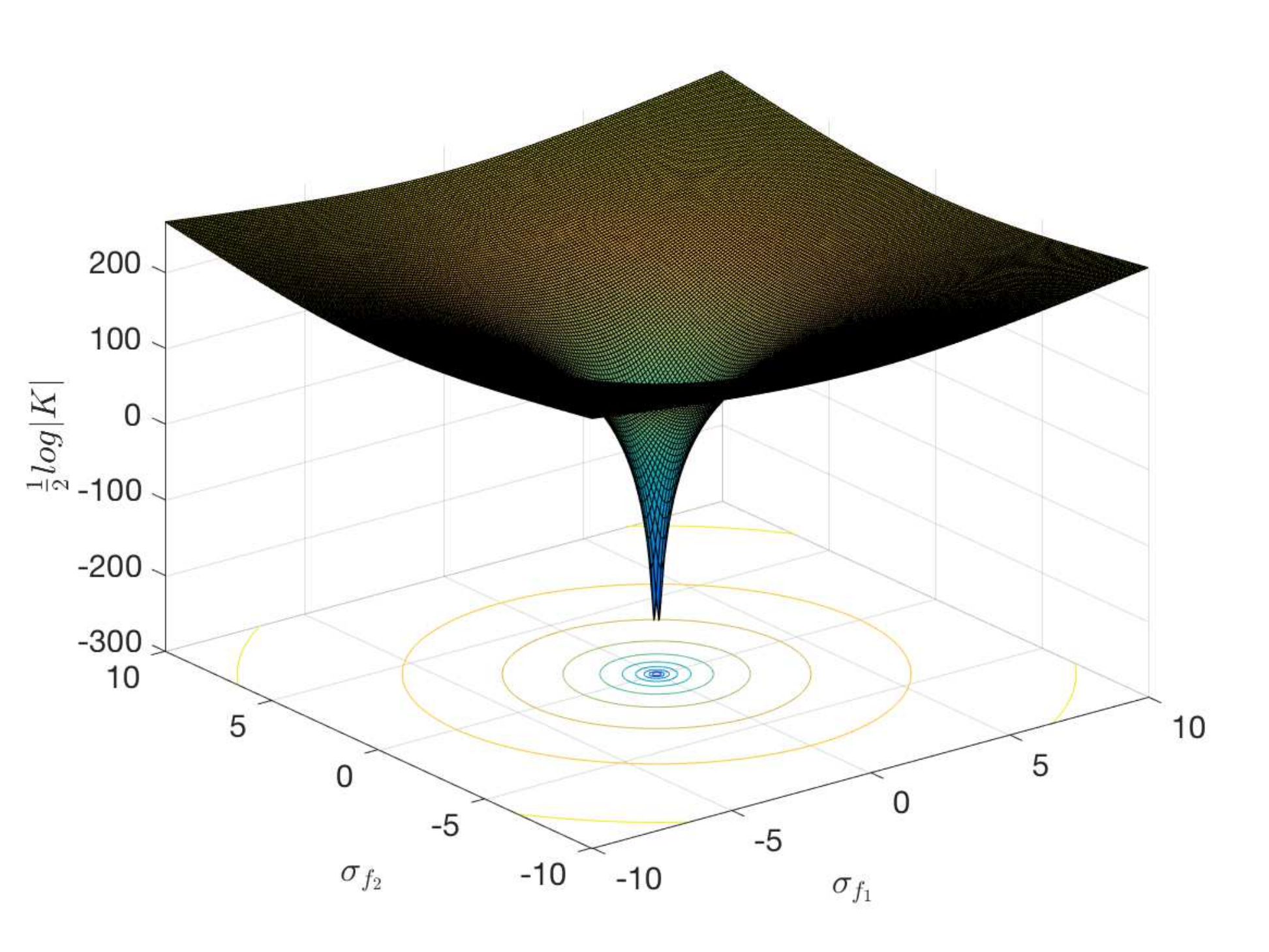}
    \caption[Gaussian process complexity surface ]{Gaussian process complexity surface obtained by the mixtures of two covariance functions (\ref{eq_covNNone}) applied to 100 training instances from a Normal/AD separation task. Also note that maximizing $-\frac{1}{2}log|K|$ is equal to minimizing $\frac{1}{2}log|K|$ and additive noise $\sigma_n^2$ is missing is due to classification. 
}
	\label{fig:GPComplexitySurface}
\end{figure}

In regards to the types of basis kernels, we prefer homogeneous combinations and do not interleave different types of basis kernels in an attempt to simplify the problem. Otherwise, a heterogenous combination would require the normalization of kernel values since each kernel would have different output scales\footnote{For $\sigma_{f}=1$, $k_{LIN}(\cdot,\cdot)$, $k_{SE}(\cdot,\cdot)$, and $k_{NN}(\cdot,\cdot)$ have different output scales.} and the larger ones could dominate others. A basis kernel could cope with such a dominance by upscaling its outputs by taking on larger values for $\sigma_{f_s}$. However, we would not know whether the magnitude of $\sigma_{f_s}$ is due to the relevance of $\mathbf{x}_{is}$ to the problem or the competition between different kernel types. 

\subsection{On Efficiency and Generality}\label{sec:effAndGen}
\Gls{MKLARD} relaxes the requirement of \Gls{ARD} from evaluating the relations between observations at the granularity of individual dimensions to assessing the relations in subspaces  and reduces the computational demands of the conventional approach. However, in the case of a large number of observations, the computational burden of \Gls{MKLARD} could be dominated by kernel computations, especially if the kernels cannot be stored in main memory. In this regard, we assume that the entire computation takes place in memory, which also suits the small sample characteristics of our problem. 
Under our assumption, the efficiency of \Gls{MKLARD} mainly depends on the granularity of feature bags. The smaller the feature bags, the larger will be the number of basis kernels and hyperparameters. 
As an extreme case, if all the feature bags are of cardinality of 1, each dimension gets assigned a basis kernel. In this extreme case, \Gls{ARD} and \Gls{MKLARD} are equivalent in the sense that both are aimed at individual dimensions. 
However, depending on the numbers of hyperparameters associated with the basis kernels, 
the overall computation required by such an \Gls{MKLARD} may be ridiculously higher 
than that required by \Gls{ARD}. 
Therefore, great care has to be taken in designing composite kernels and utilizing them for the discovery of relevant subspaces. At the other extreme, a big bag that can fit all features reduces the \Gls{MKL} problem to the traditional single kernel approach. 

As demonstrated by the two extremes above, \Gls{MKLARD} is a generalized solution for \Gls{ARD}. It can account for a wide range of possibilities at different levels of computational complexity. However, in practice, a trade-off has to be established between the full-fledged \Gls{ARD} and the single kernel learning. To this end, we heuristically define an upper bound for the total number of hyperparameters of a composite kernel as follows: $\sum_{s=1}^{S} h_{s} < D+1$, 
where $h_{s}$ is the number of hyperparameters associated with the $s$-th basis kernel, $D$ is the number of dimensions, and $D+1$ hyperparameters are required by \Gls{ARD}. For (\ref{eq_covSEisoComp}), $\forall s \in \left\{1,2,...,S \right\} h_{s}=2$, and an upper bound for the number of basis kernels is $S < (D+1)/2$. Due to the high dimensionality of neuroimaging data, we prefer $S \ll (D+1)/2$. 
Thus, \Gls{MKLARD} is proposed as an \emph{approximate}, but efficient, \Gls{ARD} solution. 


\subsection{Occam's Razor}\label{sec:OccamsRazor}
The principle of Occam's razor states that given two models with the same generalization error, we should prefer the simpler one because the simplicity is a design goal \citep{Domingos1998}. 
\citet{Domingos1998} revised the razor that we should prefer the more comprehensible one, given two models with the same generalization error. While the comprehensibility is domain-dependent, in practice, we should constrain the model selection\footnote{
``\dots model selection is essentially open ended. \dots 
However, this should not be a cause for despair, rather seen as a possibility to learn.'' \citep[p.108]{GPMLbook}. 
} 
using domain-knowledge in order to promote simplicity and comprehensibility \citep{Domingos1998}. 

\begin{quotation}
``Incorporating such constraints can simultaneously improve accuracy 
(by reducing the search needed to find an accurate model) 
and comprehensibility 
(by making the results of induction consistent with previous knowledge).'' \citep[p.5]{Domingos1998}. 
\end{quotation}

The razor motivates us to use the prior domain-knowledge regarding the inputs. 
Accordingly, we aim to 
	avoid expensive search procedures required to find plausible feature bags and  
	improve the classification performance and comprehensibility of the \Gls{GP} models. 

\subsubsection{Domain-Knowledge for Feature Bags}\label{sec:DomainKnow4Bags}
Recall that a kernel function represents our prior belief about the functions we wish to learn. Given a collection of 391 parametric images derived from \Gls{PET} scans and a taxonomy of 15,964 features that complies with the Talairach-Tournoux atlas, \citet{Ayhan2012a} handpicked certain regions of the brain containing characteristic patterns of \Gls{AD} based on domain-knowledge and showed that the use of domain-knowledge improves the prior specification even in case of the single kernel learning. Handpicking of cortical regions also resulted with significant computational gain, during both feature selection and training, and with no significant loss of classification performance \citep{Ayhan2012a}. Then, \citet{Ayhan2013b} investigated more flexible and sensible prior specifications via (\ref{eq_covSEisoComp}) and (\ref{eq_covNNoneComp}), which is absolutely desirable from the Bayesian perspective. Given the taxonomy of 15 anatomical regions, \citet{Ayhan2013b} specified anatomically motivated composite kernels and integrated information from many regions, each of which exhibits certain characteristics regarding the progression of \Gls{AD}. 

Neuroimaging procedures capture the snapshots of the metabolic demands or activities of neurons in the brain into 3D volumes, the structures of which are known beforehand. For instance, an \Gls{MRI} scan is acquired one \emph{axial} slice at a time (Figure~\ref{figureMRIAD}). These slices are combined into a big volume of brain imagery, following certain processing steps, i.e., slice timing corrections \citep{Thatcher1994,SPM8manual}. 
However, a taxonomy of voxels is not always available. In this regard, we consider the slices as feature bags, namely \emph{pseudo} regions. In addition, we can use the spatial properties of voxels to come up with richer structural formations, e.g., cubes. 
One obvious advantage of using cubes is due to their ability to capture the 3D information about patterns in their proximity. A pattern of structural deformation that cuts across multiple slices may be captured by a single cube. 

\begin{figure}[!h]
	\centering
	\begin{tabular}{ccc}
		\subfloat[Axial view (\emph{before})]{\includegraphics[width=0.3\linewidth]{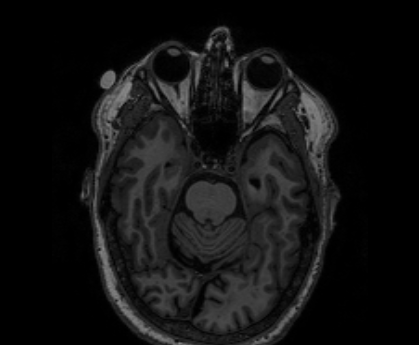}\label{figureMRIAD:a}}
		&
		\subfloat[Sagittal view (\emph{before})]{\includegraphics[width=0.3\linewidth]{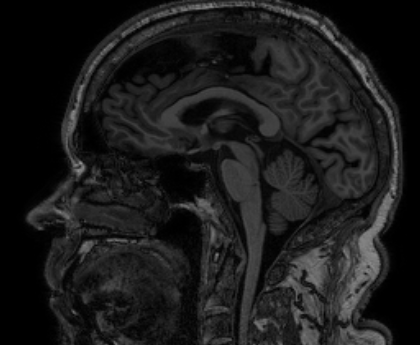}\label{figureMRIAD:b}}
		&
		\subfloat[Coronal view (\emph{before})]{\includegraphics[width=0.3\linewidth]{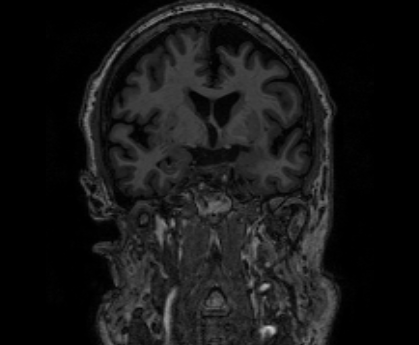}\label{figureMRIAD:c}}
		\\
		\subfloat[Axial view (\emph{after})]{\includegraphics[width=0.3\linewidth]{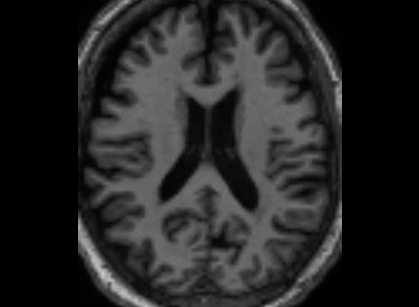}\label{figureMRIAD:d}}
		&
		\subfloat[Sagittal view (\emph{after})]{\includegraphics[width=0.3\linewidth]{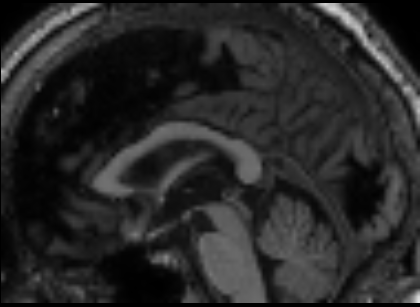}\label{figureMRIAD:e}}
		&
		\subfloat[Coronal view (\emph{after})]{\includegraphics[width=0.3\linewidth]{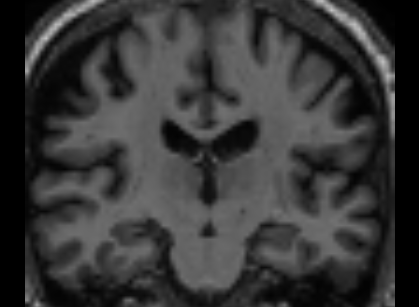}\label{figureMRIAD:f}}
	\end{tabular}
    \caption[Visualization of an MRI scan from the AD class before normalization]{Visualization of an MRI scan from the AD class \emph{before} and \emph{after} normalization. Following the normalization, the dimensionality of neuroimages reduces from $170 \times 256 \times 256$ to $79 \times 95 \times 68$.
    }
	\label{figureMRIAD}
\end{figure}

The structural information of \Gls{MRI} scans allows us to view the data from different perspectives. 
We wish to make an \emph{explicit} use of the prior information and constrain the search for an appropriate arrangement of basis kernels to a particular setting defined with respect to slices and cubes. We hope to establish a proper trade-off for the supervised feature selection via \Gls{MKLARD}. 

\section{Experiments}\label{sec:Experiments}
An \Gls{SVM}-based diagnostic method is competitive with or better than radiologists \citep{Kloppel2008}. Therefore, we assume a typical \Gls{SVM} configuration as the baseline method and investigate the benefits of utilizing \Glspl{GP} for predictive modeling of \Gls{AD}. In this respect, our experimental setup is reminiscent of \citep{Ayhan2013b} which demonstrated the advantages of using \Glspl{GP} for the analysis of parametric images derived from \Gls{PET} scans. 
However, we use a larger collection of actual \Gls{MRI} scans, no taxonomy of voxels is present, and the dimensionality is much higher. Thus, the \Gls{MRI} data enable us to demonstrate the impact of \Gls{MKLARD} under more complicated scenarios. Also, we provide a more detailed analysis of the diagnostic classification performance, including the predictive probabilities and multi-class classification results. 

Initially, we compare the performances of two kernel machines on basis of the single kernel learning. Then, we aim to improve the diagnostic classification performance of the \Gls{GP} models via \Gls{MKLARD}. We do not consider \Gls{SVM} under \Gls{MKL} settings, mainly due to the computational costs involved in \Gls{MKL}. 
The single kernel learning results also support us in so doing. 
In addition to benchmarking, 
we identify the prominent portions of the brain through data analysis and interpret the results in a manner consistent with the \Gls{AD} pathology. The impact of \Gls{MKLARD} on predictive probabilities is of interest to us, as well. 

In order to estimate the generalization performances of the specified algorithms, we apply 10-fold \Gls{CV}. Our performance metrics are classification accuracy, sensitivity, and specificity. 

\begin{equation}\label{eqAccuracy}
\text{Accuracy} = \frac{\text{True Pos. + True Neg.}}{\text{True Pos. + True Neg. + False Pos. + False Neg.}} \nonumber
\end{equation}
\begin{equation}\label{eqSensitivity}
\text{Sensitivity} = \frac{\text{True Pos.}}{\text{True Pos. + False Neg.}} \nonumber
\end{equation}
\begin{equation}\label{eqSpecificity}
\text{Specificity} = \frac{\text{True Neg.}}{\text{True Neg. + False Pos.}} \nonumber
\end{equation}

Sensitivity indicates our ability to recognize positive results, whereas specificity does it for negative results. For instance, given an \Gls{AD} patient, sensitivity is the probability that our classifier will assign the correct diagnosis. Specificity is the probability that our classifier will tell that the patient is healthy, given that he/she is, in fact, healthy. In medical diagnosis, the trade-off between these two measures are crucial. A highly sensitive test is valuable if there is a high penalty for missing positive cases \citep{Fletcher2012}. 
``Sensitive tests are also helpful during the early stages of a diagnostic workup, when several diagnoses are being considered, 
to reduce the number of possibilities'' \citep[p.50]{Fletcher2012}. 
On the other hand, highly specific tests are useful in reducing the risks as well as harm due to 
unnecessary procedures resulting from false positives \citep{Fletcher2012}. 
Overall, it is desirable for a diagnostic test to achieve high sensitivity and specificity simultaneously. However, in practice, this may not be possible and one measure can be traded for the other \citep{Fletcher2012}. In our experiments, we favor sensitivity because the risks involved in missing a positive case are usually greater than in misdiagnosing a healthy patient. 

For binary classification tasks, 
we provide another measure: \emph{\Gls{AUC}}. The \Gls{AUC} measure is obtained via \emph{\Gls{ROC}} analysis and it summarizes the overall classification performance into one number. Based on the \Gls{AUC} scores, we compare the classification performances and test for significant differences. 

For \Glspl{SVM} \citep{Chang2011}, we use a single linear kernel as well as the \Gls{RBF} kernel. 
A grid search is also performed using a fold of data. As a result, 80\%, 10\%, and 10\% of data are used for training, validation, and testing, respectively. Note that we do not retrain \Glspl{SVM} after adding the validation set to the training set. Also, we consider the generation of probabilistic outputs along with class labels for each \Gls{SVM} configuration tested in single kernel learning scenarios. 
For the \Gls{GP} models \citep{GPMLtoolbox}, we use all the kernel functions described in Section~\ref{sec:GPR} and Section~\ref{sec:compKernels}. 
Since the \Gls{GP} models do not require validation sets, the training and test splits correspond to 
90\% and 10\% of data. 
We fit a constant mean function 
to data 
and 
\Gls{EP} is used for inference, unless stated otherwise. 

\subsection{Data Acquisition}\label{sec:DataAcq}

Data used in this study are from the \Gls{ADNI} (\href{http://adni.loni.usc.edu/}{adni.loni.usc.edu}). 
The \Gls{ADNI} was launched in 2003 as a public-private partnership, led by Principal Investigator Michael W. Weiner, MD. The primary goal of \Gls{ADNI} has been to test whether serial \Gls{MRI}, \Gls{PET}, other biological markers, and clinical and neuropsychological assessment can be combined to measure the progression of \Gls{MCI} and early \Gls{AD}. 

\subsection{Data Characteristics}\label{sec:DataChar}
Table~\ref{tableDemograpghic} describes the demographics of the patients in the collection. There are three groups: Normal, \Gls{MCI} and \Gls{AD}. The \Gls{MCI} group is the largest and the dataset is skewed. In order to obtain a balanced dataset, 755 scans were randomly sampled without replacement from each group. 

\begin{table}[!h]
\caption[Demographics of MRI data]{Demographics of MRI data.}
\label{tableDemograpghic}
\begin{center}
\begin{tabular}{rcccccc}
\toprule
 							& Number of 	& \multicolumn{2}{c}{Sex}	& Average		& Number of \\
\cmidrule{3-4}
							& Subjects 	& M & F 					& Age	& MRI Scans \\
\midrule
Normal	& 232 & 113 & 119	& 76.2 &	1278 \\
MCI		& 411 & 267 & 114 	& 75.5 &	2282 \\
AD		& 200 & 103 & 97	& 76.0 &	755 \\
\bottomrule
\end{tabular}
\end{center}
\end{table}

\subsection{Stereotactic Normalization}\label{sec:StereoNorm}
A raw \Gls{MRI} scan (Figure~\ref{figureMRIAD:a}-\ref{figureMRIAD:c}) consists of massive amounts of voxels and it needs to go through a preprocessing pipeline before analysis. The goals of preprocessing include the minimization of differences in brain size, shape and position, and hence the standardization of brain regions and tissue boundaries across subjects. 
\Gls{SPM}~\citep{SPM8manual} is used 
to normalize the image data into an \Gls{ICBM} template~\citep{Gupta2013a,Thatcher1994}. 
Figure~\ref{figureMRIAD:d}-\ref{figureMRIAD:f} demonstrates the result of normalization. Note that 
no anatomical structure such as \Gls{GM}, \gls{WM} or \gls{CSF} is extracted. Instead, 
the whole brain data is deliberately preserved because brain atrophy can be measured both in \Gls{GM} and \Gls{WM} \citep{Fan2008}. We expect the machine learning software to figure out whatever is informative and determine the relevant portions of data.

\subsection{Arrangement of the Basis Kernels}

The number of basis kernels is essentially a hyperparameter for a given model. In the case of axial slices, the number is 68. However, if cubes are desired, it has to be determined according to user preferences. Given the dimensions of normalized \Gls{MRI} scans, 
the cube dimensions of 
$4 \times 4 \times 4$, $8 \times 8 \times 8$, $16 \times 16 \times 16$ and $32 \times 32 \times 32$ would result in 8160, 1080, 150, and 27 basis kernels, respectively, assuming that 
they were forbidden from overlapping.\footnote{3D convolution via cubes is possible; however, it would dramatically increase the number of basis kernels.}  
Due to the coarseness of the largest cube dimensions and excessive number of basis kernels induced by the smaller ones, we set the cube dimensions as $16 \times 16 \times 16$. 
This sweet spot not only leads to a significant saving on computational time but also helps us alleviate the potentially weaker consistency in the case of larger cube dimensions. Also note that, despite the fixed number and layout of kernels, we learn the kernel properties automatically from data, which is central to \Gls{MKL}. 

\subsection{Binary Classification Results}\label{sec:BinaryClassResults}
Table~\ref{table_BinaryClassificationResults} shows the binary classification results. However, in many cases, it is difficult to discern the differences between performances. Since the numbers given in the table correspond to single points on the average \Gls{ROC} curves of the classifiers, we judge their performances by the \Gls{AUC} scores in Figure~\ref{fig:AUCcomparison} while we keep the table for reference. 

From Table~\ref{table_BinaryClassificationResults} and Figure~\ref{fig:AUCcomparison}, we can tell that SVMs are fairly accurate. When we demand probabilistic outputs from \Glspl{SVM}, their performances are competitive with the basic configuration despite the use of smaller training sets. 
However, the generation of probabilities significantly complicates training. The differences between the performances of linear (LIN) and non-linear (RBF) \Glspl{SVM} indicate the non-linearity of the complex atrophy patterns. 
\Glspl{GP} tend to significantly outperform the baseline \Gls{SVM}, which is a linear one. Exceptions are those with a single \Gls{SE} covariance function or which used \Gls{LA} in slice-based \Gls{MKL} settings. Due to the quadratic form in the exponent of the \Gls{SE} covariance function, even the slightest change in a large number of input values easily causes the covariance between $f_i$ and $f_j$ to tend to zero. The same phenomenon was reported based on the experiments with \Gls{PET} data, as well \citep{Ayhan2013b}. 
However, \Glspl{SVM} with the \Gls{RBF} kernel are more robust to this situation, now. 
In our defense, the poor performance of such \Gls{GP} models can be rectified by the virtue of \Gls{MKLARD} when the most relevant inputs are emphasized for the calculation of similarities (Figure~\ref{fig:AUCcomparison}a and Figure~\ref{fig:AUCcomparison}c). 
On the contrary, Figure~\ref{fig:AUCcomparison}b 
shows a case in which slice-based \Gls{MKLARD} failed to recover from the catastrophe. In this case, 
the \Gls{EP} algorithm failed to complete during at least one iteration of \Gls{CV}, even though the remaining iterations resulted in performances that are comparable with other models'. Due to the instability concerns, we resorted to \Gls{LA} throughout \Gls{CV}. Evidently, \Gls{LA} failed to provide good estimates of mean and covariance. As a result, the performances are far from satisfying and the variation in the performance is larger compared with the others (Table~\ref{table_BinaryClassificationResults}, 22\textsuperscript{nd} row). Nevertheless, the use of cubes, instead of slices, remedies the situation. 

Another observation to be made from Figure~\ref{fig:AUCcomparison} is the competitiveness of the \Gls{GP} models equipped with linear covariance functions. Even a single function is quite satisfying in terms of classification performance. Considering the fewer number of hyperparameters and corresponding computational requirements, linear covariance function may be a natural choice for both single and multiple kernel learning scenarios. For those who want to make the most out of \Gls{GP} learning, the \Gls{NN} covariance function is also available. 

Lastly, we did not expect the competitive performance demonstrated by a single linear covariance function before the experiments. Given the performances of linear \Glspl{SVM}, this seems counter intuitive. 
We speculate that it may be due to the algorithmic differences under the hood of the respective methodologies. On the bright side, \Gls{GP} models can outperform \Glspl{SVM} even in linear settings. 

\begin{table}[!h]
\renewcommand{\arraystretch}{1.15}
\caption[The average generalization performances based on 10-fold \acrlong{CV}]{The average generalization performances based on 10-fold \Gls{CV}. Standard deviations are given in parentheses. The highest average in each category is given in boldface with a preference for smaller deviation in cases of equality. 
GP\textsuperscript{LA} means that \Gls{LA} was used. 
}
\label{table_BinaryClassificationResults}
\centering
\begin{tabular}{@{\hspace{2.5mm}}p{0.1cm} @{\hspace{2.5mm}}p{0.1cm} @{\hspace{2.5mm}}p{0.1cm} @{\hspace{2.5mm}}r @{\hspace{2.5mm}}l @{\hspace{2.5mm}}l @{\hspace{2.5mm}}l} 
\\
\toprule
		&	&		&	\textbf{Classifier}	& \textbf{Accuracy}	& \textbf{Sensitivity}	& \textbf{Specificity}	\\
\midrule
\multirow{13}{*}{\begin{sideways} \textbf{Normal vs. \Gls{AD}} \end{sideways}}
						&
\multirow{7}{*}{\begin{sideways} \textsl{Single Kernel} \end{sideways}}	
						&
\multirow{7}{*}{} 
						&	SVM LIN				&	89.20 (3.14)	&	89.60 (3.76)		&	88.80 (4.50)		\\
		&	&			&	SVM\textsuperscript{prob} LIN			&	89.20 (2.93)		&	89.87 (3.62)		&	88.53 (4.36)		\\
		&	&			&	SVM RBF				&	\textbf{90.60 (2.60)}	&	\textbf{90.80 (3.11)}		&	\textbf{90.40 (3.92)}		\\
		&	&			&	SVM\textsuperscript{prob} RBF			&	89.87 (2.39)		&	90.67 (2.35)		&	89.07 (3.86)		\\
\cmidrule{4-7}
		&	&			&	GP LIN				&	92.67 (1.57)	&   \textbf{92.40 (2.67)}		&	92.93 (1.89)		\\
		&	&			&	GP SE				&	83.27 (2.02)	&   82.27 (4.31)		&	84.27 (3.31)		\\
		&	&			&	GP NN				&	\textbf{92.80 (1.47)}	&   92.13 (2.70)		&	\textbf{93.47 (1.60)}		\\
\cmidrule{2-7}
						&
\multirow{6}{*}{\begin{sideways} \textsl{\Gls{MKLARD}} \end{sideways}} 
						&
\multirow{3}{*}{\begin{sideways} \textsl{Slices} \end{sideways}} 
						&	GP LIN 				&	93.00 (2.07)	&	92.40 (3.50)		&	\textbf{93.60 (1.86)}	\\
		&	&			&	GP SE 				&	91.20 (2.79)	&	91.87 (3.23)		&	90.53 (3.95)		\\
		&	&			&	GP NN				&	\textbf{93.20 (2.43)}	&	\textbf{93.20 (3.74)}		&	93.20 (2.39)		\\
\cmidrule{3-7}
			&			&
\multirow{3}{*}{\begin{sideways} \textsl{Cubes} \end{sideways}}
						&	GP LIN 				&	93.53 (1.34)	&	92.13 (2.91)		&	\textbf{94.93 (2.25)}		\\
		&	&			&	GP\textsuperscript{LA} SE				&	91.67 (2.04)		&	89.87 (3.28)		&	93.47 (2.31)		\\
		&	&			&	GP NN				&	\textbf{93.87 (1.72)}	&	\textbf{93.60 (2.73)}		&	94.13 (2.89)		\\

\midrule\midrule
\multirow{13}{*}{\begin{sideways} \textbf{Normal vs. \Gls{MCI}} \end{sideways}}
						&
\multirow{7}{*}{\begin{sideways} \textsl{Single Kernel} \end{sideways}}	
						&
\multirow{7}{*}{} 
						&	SVM LIN				&	81.40 (3.93)	&	80.67 (3.78)		&	82.13 (5.15)		\\
		&	&			&	SVM\textsuperscript{prob} LIN			&	81.20 (3.72)		&	81.60 (4.25)		&	80.80 (4.45)		\\
		&	&			&	SVM RBF				&	\textbf{84.80 (2.77)}	&	84.53 (3.78)		&	\textbf{85.07 (4.39)}		\\
		&	&			&	SVM\textsuperscript{prob} RBF			&	84.27 (2.83)		&	\textbf{84.53 (3.57)}		&	84.00 (3.98)		\\
\cmidrule{4-7}
		&	&			&	GP LIN 				&	86.13 (2.83)	&	84.00 (2.95)		&	\textbf{88.27 (3.81)}		\\
		&	&			&	GP SE				&	72.13 (8.38)	&	76.00 (9.28)		&	68.27 (23.14)		\\
		&	&			&	GP NN				&	\textbf{86.20 (2.78)}	&	\textbf{84.13 (2.91)}		&	\textbf{88.27 (3.81)}		\\
\cmidrule{2-7}
						&
\multirow{6}{*}{\begin{sideways} \textsl{\Gls{MKLARD}} \end{sideways}} 
						&
\multirow{3}{*}{\begin{sideways} \textsl{Slices} \end{sideways}} 
						&	GP LIN 				&	87.40 (2.75)	&	86.53 (3.52)		&	88.27 (3.97)		\\
		&	&			&	GP\textsuperscript{LA} SE 				&	72.27 (14.23)	&	73.60 (12.10)		&	70.93 (17.45)		\\
		&	&			&	GP NN				&	\textbf{88.07 (2.42)}	&	\textbf{86.93 (3.65)}		&	\textbf{89.20 (3.47)}		\\
\cmidrule{3-7}
			&			&
\multirow{3}{*}{\begin{sideways} \textsl{Cubes} \end{sideways}}
						&	GP LIN 				&	89.60 (1.89)	&	88.93 (2.60)		&	\textbf{90.27 (2.74)}		\\
		&	&			&	GP\textsuperscript{LA} SE				&	87.27 (3.09)	&	85.20 (4.64)		&	89.33 (3.56)		\\
		&	&			&	GP NN				&	\textbf{89.93 (1.65)}	&	\textbf{89.73 (3.02)}		&	90.13 (3.78)		\\

\midrule\midrule
\multirow{13}{*}{\begin{sideways} \textbf{\Gls{MCI} vs. \Gls{AD}} \end{sideways}}
						&
\multirow{7}{*}{\begin{sideways} \textsl{Single Kernel} \end{sideways}}	
						&
\multirow{7}{*}{} 
						&	SVM LIN				&	82.13 (3.73)	&	83.20 (4.92)		&	81.07 (4.52)		\\
		&	&			&	SVM\textsuperscript{prob} LIN			&	82.67 (3.40)		&	82.93 (4.98)		&	82.40 (4.65)		\\
		&	&			&	SVM RBF				&	87.27 (2.25)	&	88.27 (3.86)		&	\textbf{86.27 (4.03)}		\\
		&	&			&	SVM\textsuperscript{prob} RBF			&	\textbf{87.40 (2.28)}		&	\textbf{88.67 (4.13)}		&	86.13 (3.68)		\\
\cmidrule{4-7}
		&	&			&	GP LIN 				&	88.80 (3.37)	&	\textbf{91.07 (4.08)}		&	86.53 (4.28)		\\
		&	&			&	GP SE				&	80.27 (2.60)	&	79.33 (3.10)		&	81.20 (4.81)		\\
		&	&			&	GP NN				&	\textbf{88.87 (3.44)}	&	90.80 (4.33)		&	\textbf{86.93 (3.81)}		\\
\cmidrule{2-7}
						&
\multirow{6}{*}{\begin{sideways} \textsl{\Gls{MKLARD}} \end{sideways}} 
						&
\multirow{3}{*}{\begin{sideways} \textsl{Slices} \end{sideways}} 
						&	GP LIN 				&	\textbf{89.40 (2.44)}	&	\textbf{92.00 (4.36)}		&	86.80 (1.72)		\\
		&	&			&	GP\textsuperscript{LA} SE 				&	88.00 (2.24)	&	88.67 (3.46)		&	\textbf{87.33 (2.37)}		\\
		&	&			&	GP NN				&	88.73 (2.73)	&	91.20 (3.78)		&	86.27 (4.36)		\\
\cmidrule{3-7}
			&			&
\multirow{3}{*}{\begin{sideways} \textsl{Cubes} \end{sideways}}
					&	GP LIN 				&	88.87 (2.65)	&	\textbf{90.53 (3.41)}		&	87.20 (4.54)		\\
		&	&			&	GP\textsuperscript{LA} SE				&	87.53 (2.92)	&	88.40 (3.50)		&	86.67 (5.11)		\\
		&	&			&	GP NN				&	\textbf{89.13 (2.18)}	&	90.13 (2.45)		&	\textbf{88.13 (3.47)}		\\

\bottomrule
\end{tabular}
\end{table}

\begin{figure}[h!]
	\centering
	\begin{tabular}{c}
		\subfloat[Normal vs. AD]{\includegraphics[width=0.575\linewidth]{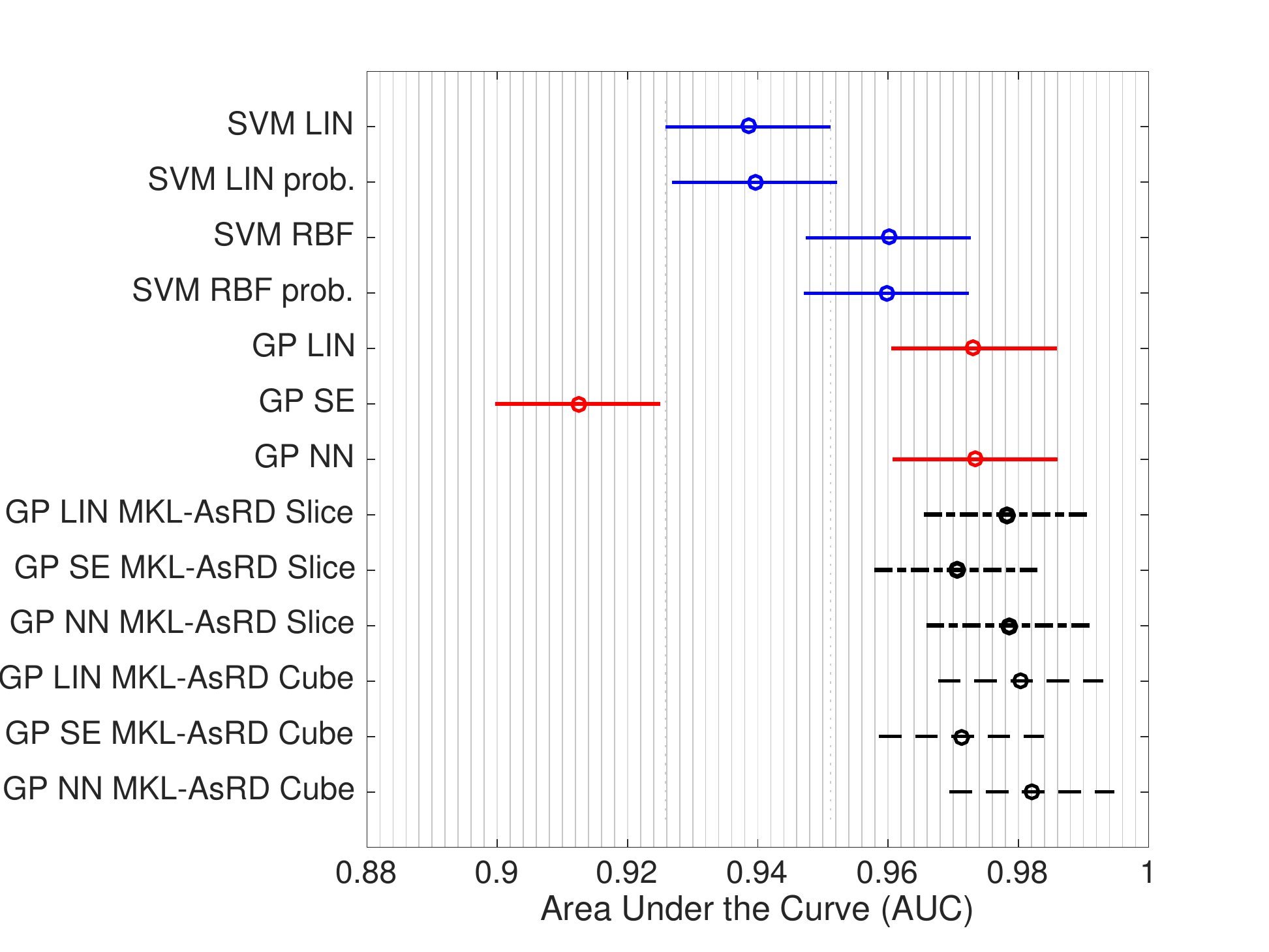}}
		\\
		\subfloat[Normal vs. MCI]{\includegraphics[width=0.575\linewidth]{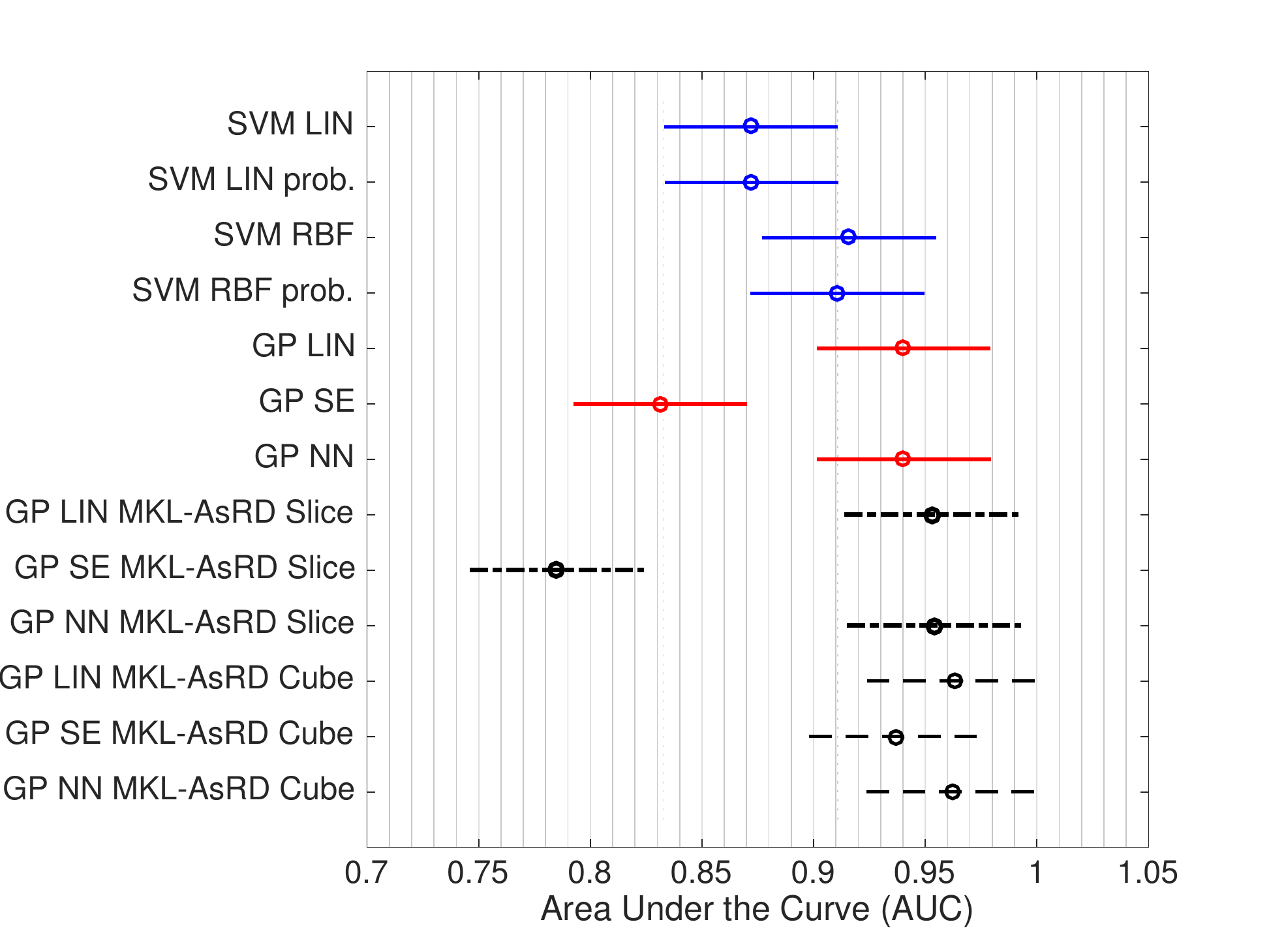}}
		\\
		\subfloat[MCI vs. AD]{\includegraphics[width=0.575\linewidth]{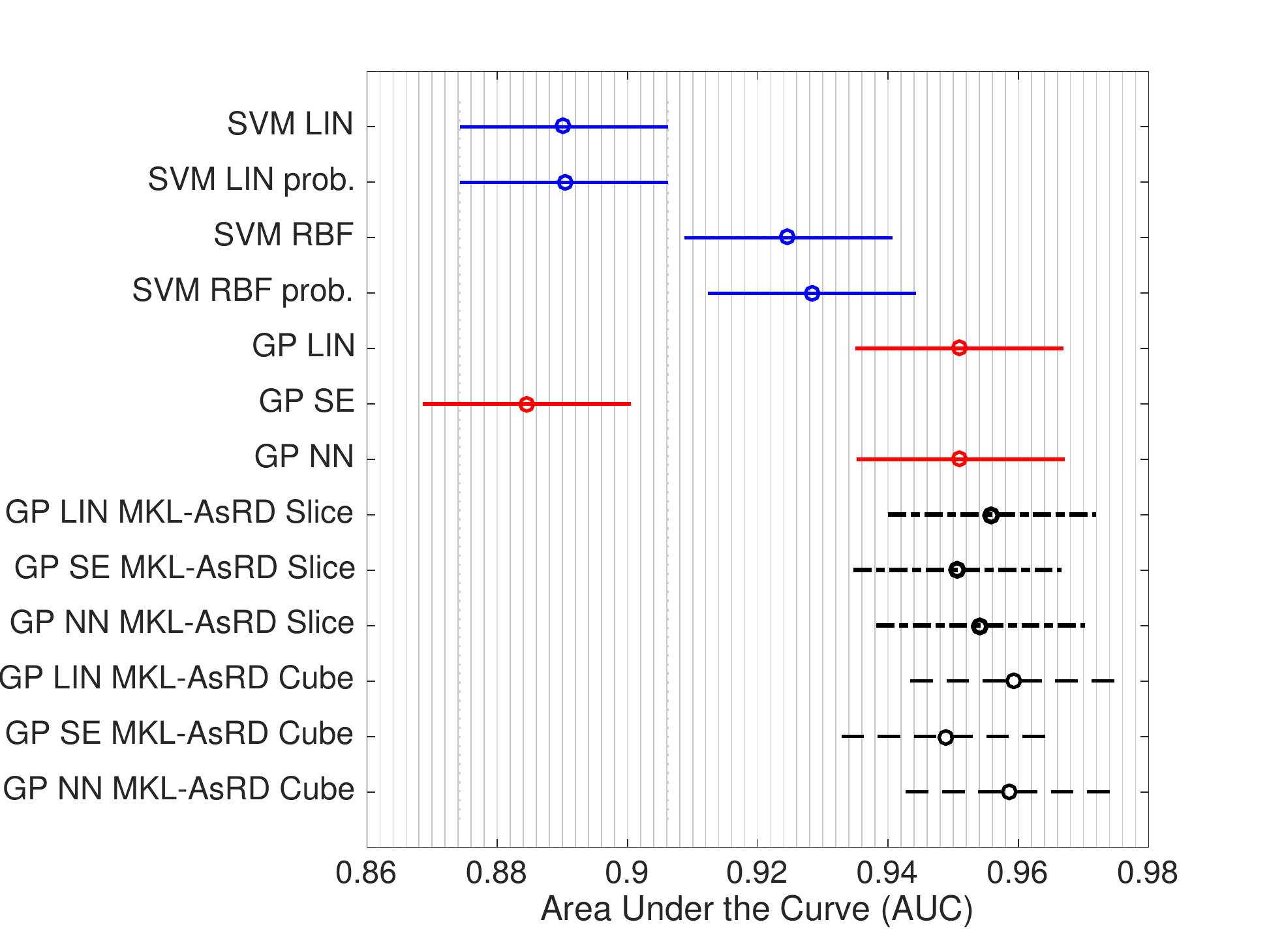}}
	\end{tabular}
    \caption[Average (mean) \Gls{AUC} results and comparison intervals for binary classification tasks]{Average (mean) \Gls{AUC} results and comparison intervals for binary classification tasks. 
    The confidence level is 95\%. Two means are significantly different if their comparison intervals do not overlap. 
}
	\label{fig:AUCcomparison}
\end{figure}

\subsubsection{Regions of Relevance and Comprehensibility}\label{sec:relComp}
Figure~\ref{fig:relScoresByMKLARD} shows the relevant portions of brain imagery that are determined via \Gls{MKLARD}. We consider the relevant slices and cubes as the \emph{\gls{ROR}}.
Since an anatomical taxonomy of them is not present, they are not exactly compatible with the known anatomical structures. However, Figure~\ref{fig:relScoresByMKLARD} shows that 
the models generally focus on the lower brain, where the \Gls{AD} pathology starts in and around the hippocampus and entorhinal cortex \citep{Whitwell2007,Fan2008}. 
Also note that the models are consistent in their preferences for the top regions in spite of the fundamental changes in their kernel configurations. 

The patterns of atrophy due to \Gls{AD} are bilateral in spite of the greater impact on the left hemisphere \citep{Whitwell2007}. 
The \Gls{GP} models with the cube-based \Gls{MKLARD} capability better respond to the bilateral patterns of the \Gls{AD} pathology as it results in more connected structures in comparison with the slice-based configuration.  (Figure~\ref{fig:relScoresByMKLARD}). 



\begin{figure}[h]
	\centering
	\begin{tabular}{ccc}
		\subfloat[Top 7 slices via GP LIN MKL-AsRD: 9,14,4,25,21,22,2]{\includegraphics[width=0.35\linewidth]{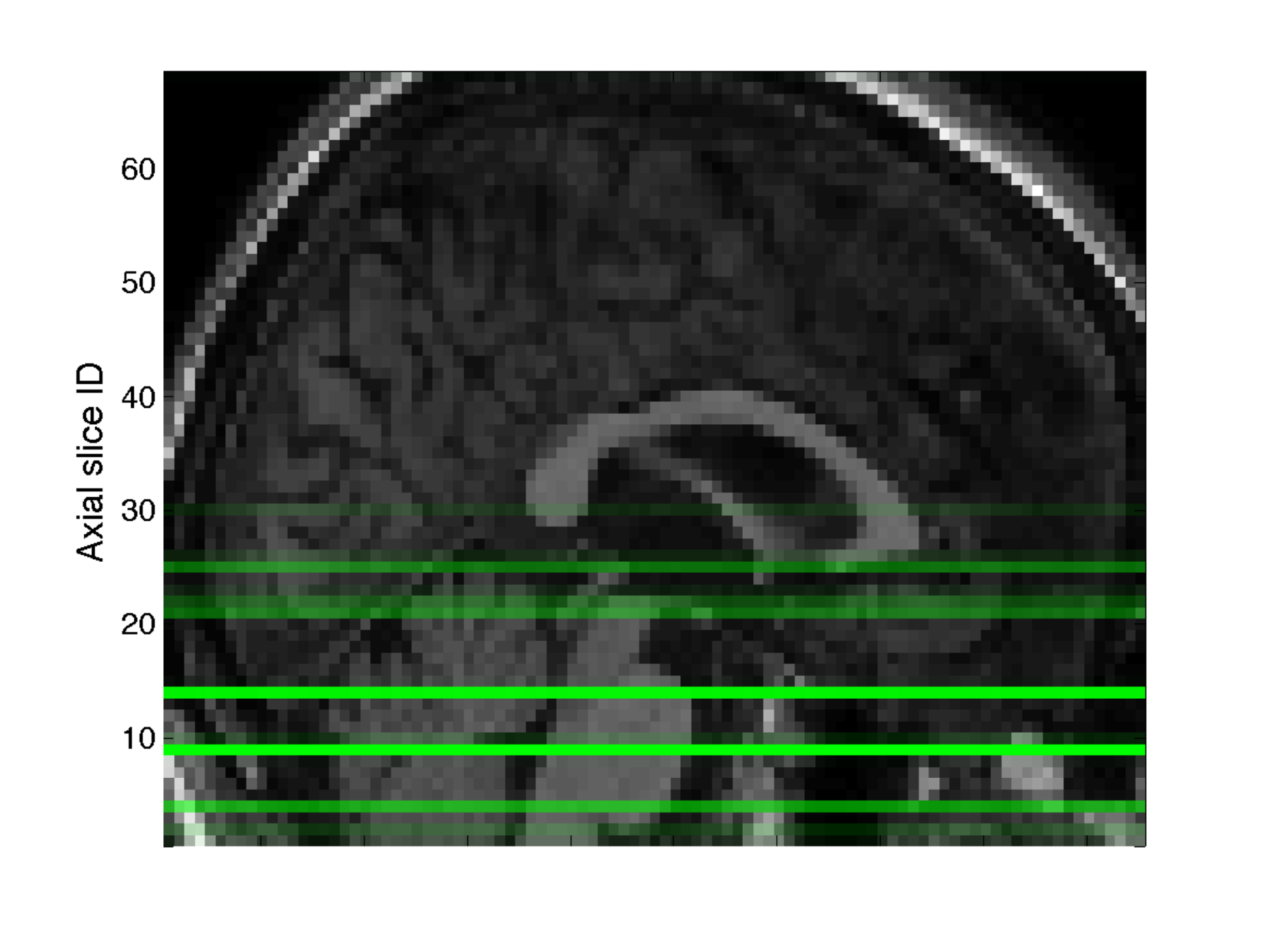}\label{fig:relScoresByMKLARD:a}}
		&
		\subfloat[Top 7 slices via GP SE MKL-AsRD: 14,9,24,8,10,23,11]{\includegraphics[width=0.35\linewidth]{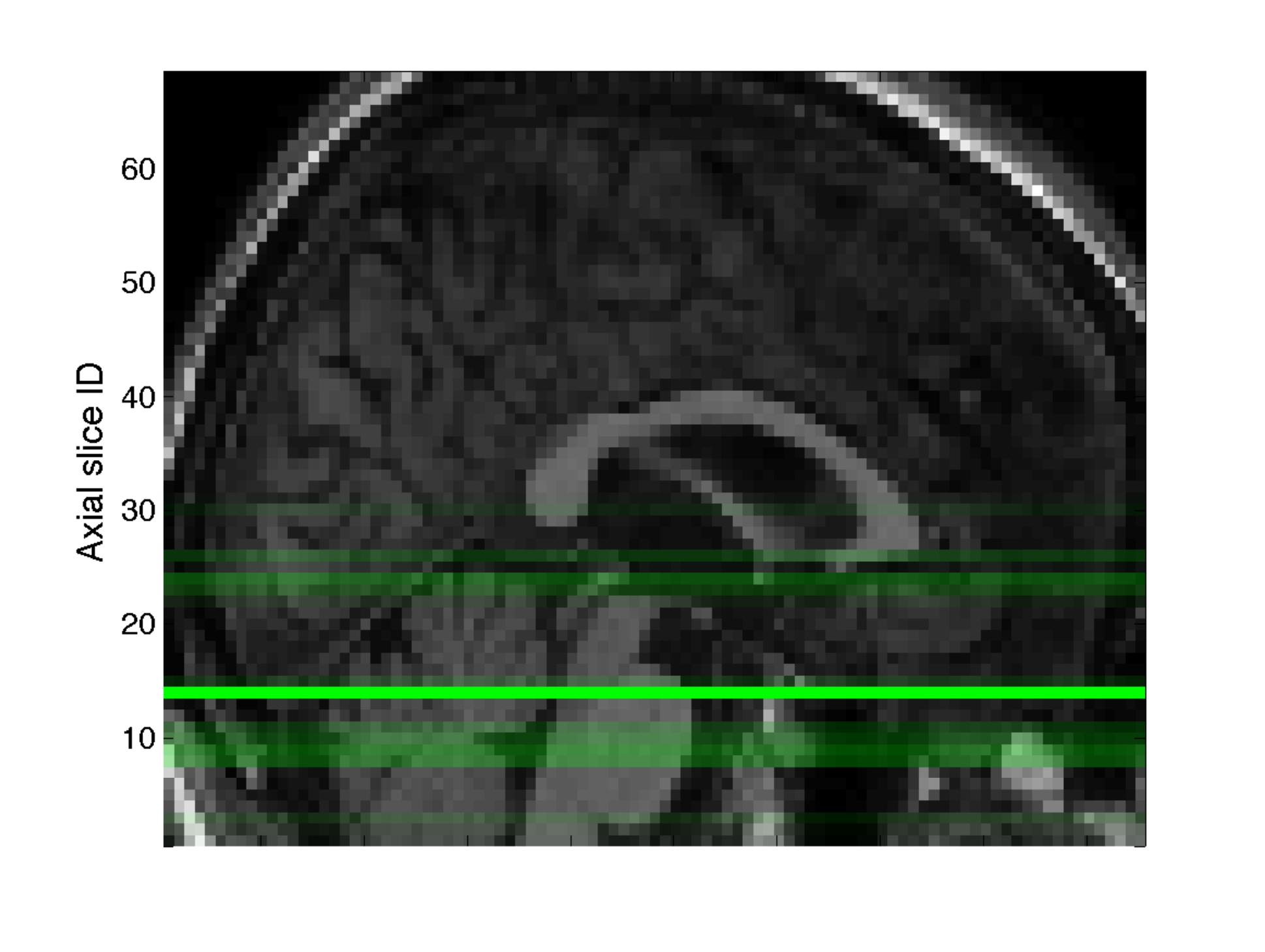}\label{fig:relScoresByMKLARD:b}}
		&
		\subfloat[Top 7 slices via GP NN MKL-AsRD: 14,30,4,9,26,22,15]{\includegraphics[width=0.35\linewidth]{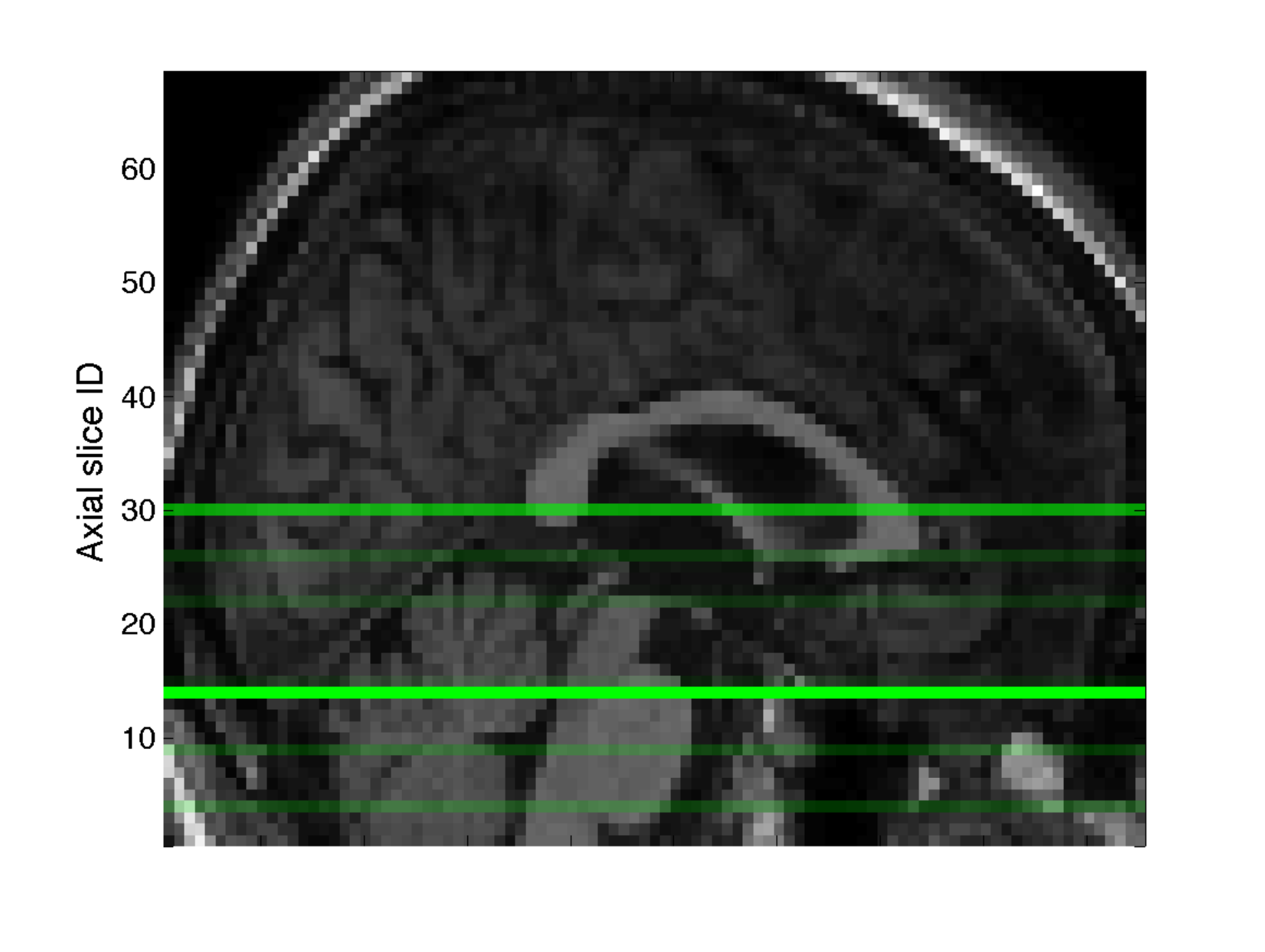}\label{fig:relScoresByMKLARD:c}}
		\\
		\subfloat[Top 15 cubes via GP LIN MKL-AsRD: 46,41,106,47,71,107,42,77, 112,97,96,81,13, 116,87]{\includegraphics[width=0.3\linewidth]{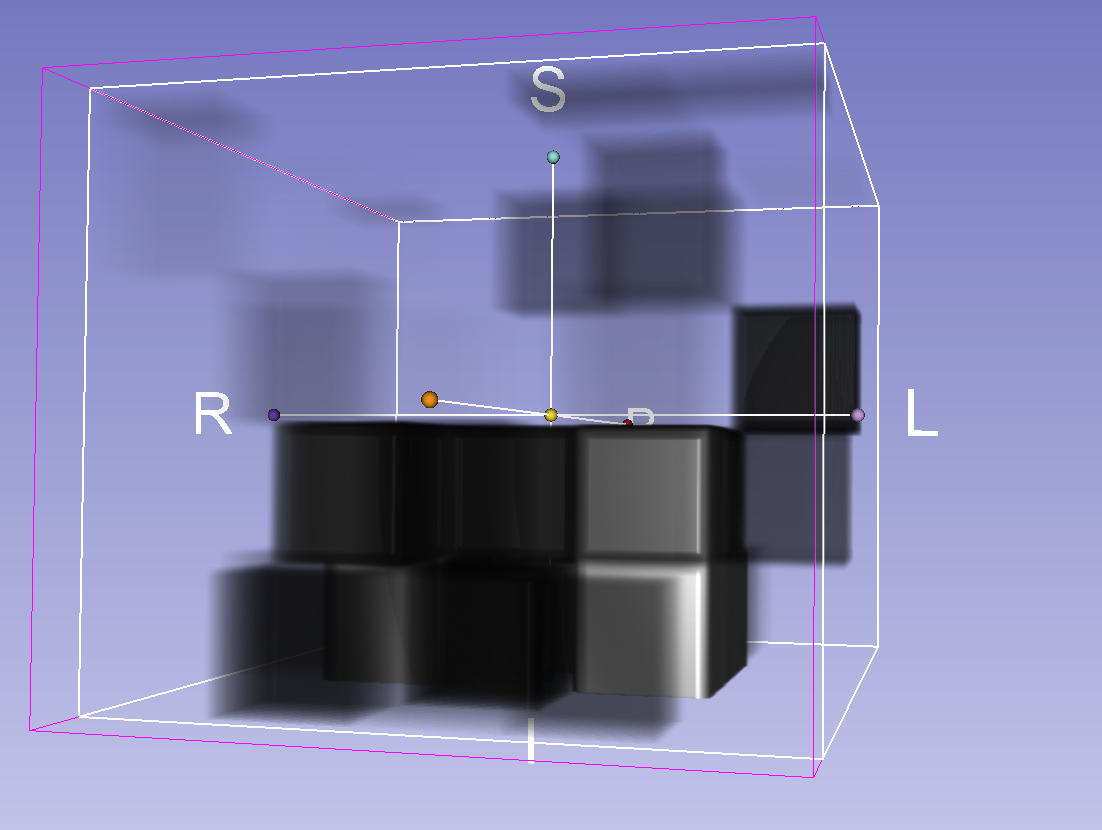}\label{fig:relScoresByMKLARD:d}}
		&
		\subfloat[Top 15 slices via GP SE MKL-AsRD: 47,41,77,107,106,46,71,97,52, 70,102,112,38, 43,103]{\includegraphics[width=0.3\linewidth]{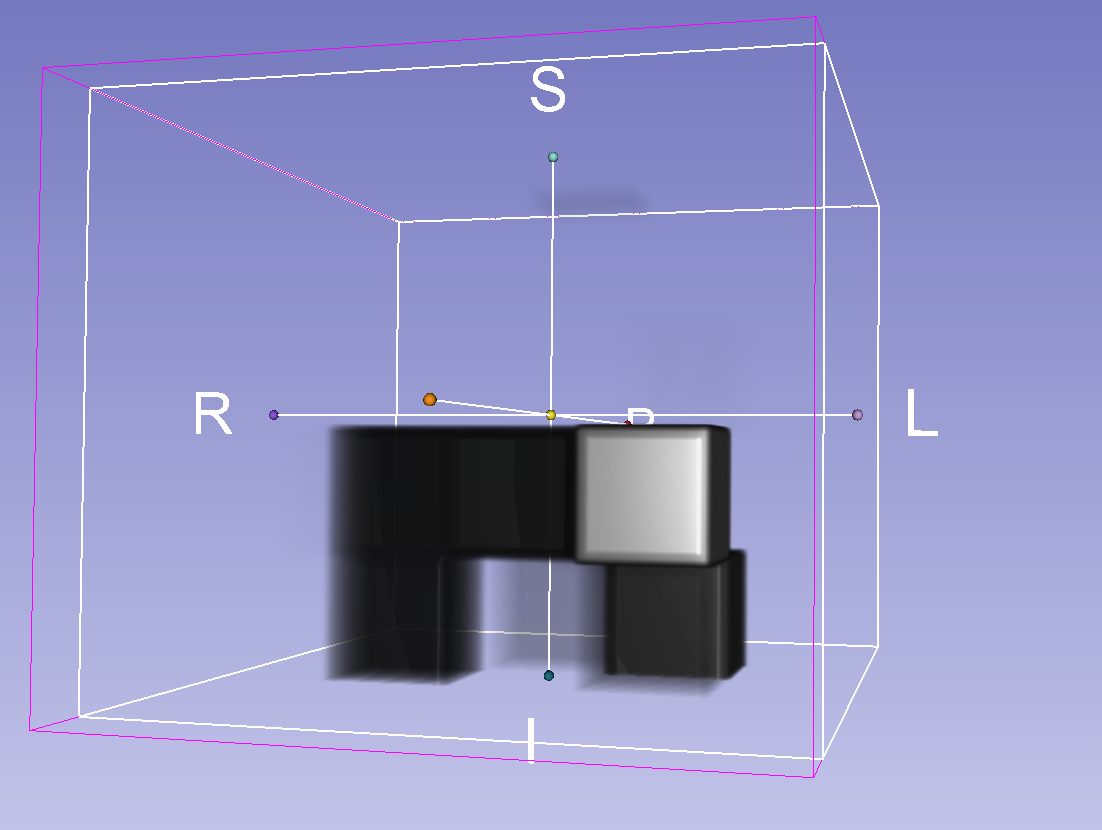}\label{fig:relScoresByMKLARD:e}}
		&
		\subfloat[Top 15 slices via GP NN MKL-AsRD: 107,77,41,42,81,56,47,97,101, 36,17,46,71, 106,112]{\includegraphics[width=0.3\linewidth]{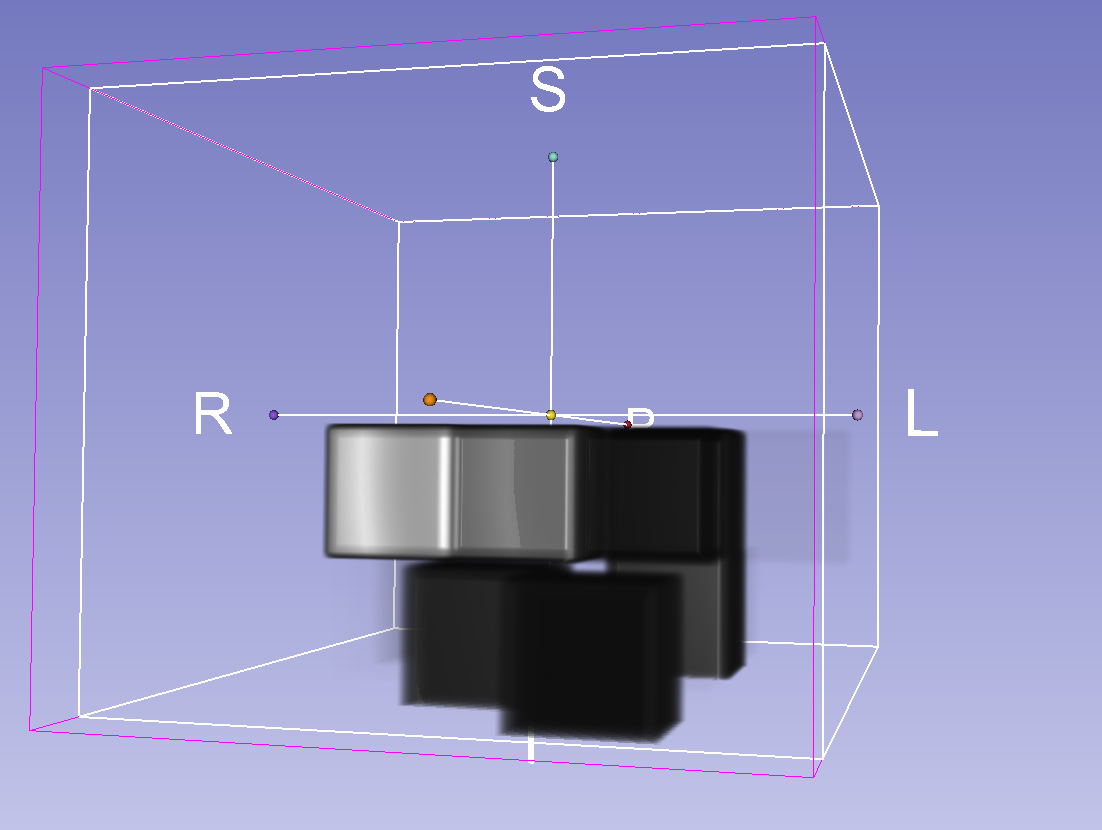}\label{fig:relScoresByMKLARD:f}}
	\end{tabular}
    \caption[Axial slices and cubes of relevance for Normal/AD classification.]{Axial slices (top row) and cubes (bottom row) of relevance for Normal/AD classification. Relevance scores are determined based on 10-fold \Gls{CV}. For each CV iteration, the mixing weights are normalized separately. Then, they are accumulated over iterations. Note that the normalized weights are always less than or equal to 1. Thus, the relevance scores are between 0 and 10. In this scheme, a relevance score towards the high end of the spectrum indicates that the associated slice or cube has been emphasized many times during our experiments. On the other hand, a score from the low end means that the slice/cube was pretty much ignored due to its low relevance to the classification task at hand. In the top row, the brighter (greener) a slice is, the more relevant it is. Similarly, in the bottom row, the brighter (grayer) a cube is, the more relevant it is. Invisible cubes are considered irrelevant due to their relevance scores and hence not shown in the figures.}
	\label{fig:relScoresByMKLARD}
\end{figure}

\subsubsection{Predictive Probabilities}\label{sec:predProb}
Figure~\ref{fig:histPredProb} indicates that the \Gls{GP} models using a single \Gls{SE} covariance function tend to be overly cautious about their predictions. Even though the signs of predictions ($f_*$) are mostly correct, the predictive probabilities ($p_*$) are centered around 0.5 (Figure~\ref{fig:histPredProb:d}).
Such a conservative behavior can be associated with the underestimations of mean and covariance via \Gls{LA} \citep{Kuss2005}. But, it can happen due to the inappropriateness of the \Gls{SE} covariance function even if the \Gls{EP} algorithm is used for inference (Table~\ref{table_BinaryClassificationResults}, row 6). \Gls{MKLARD} has a dramatic impact in this regard. The predictive probabilities are widespread and mostly closer to 0 or 1 (Figure~\ref{fig:histPredProb:e}), which is desirable for confident decisions. Also, the average classification accuracy goes up by 7.93\% as a result of \Gls{MKLARD}. Another benefit of the \Gls{MKLARD} is that it yields more confident models compared with the single kernel counterparts, even when \Gls{LA} is used for \Gls{GP} inference (Figure~\ref{fig:histPredProb:f}). 
The first and last rows of Figure~\ref{fig:histPredProb} show that \Gls{MKLARD} improves the predictive probability distributions in cases of the linear and \acrlong{NN} covariance functions, as well. Not to mention, the choice of cubes leads to more confident models, in comparison with the slice-based configurations.

\begin{figure}[!h]
	\centering
	\begin{tabular}{ccc}
		\subfloat[GP LIN]{\includegraphics[width=0.33\linewidth]{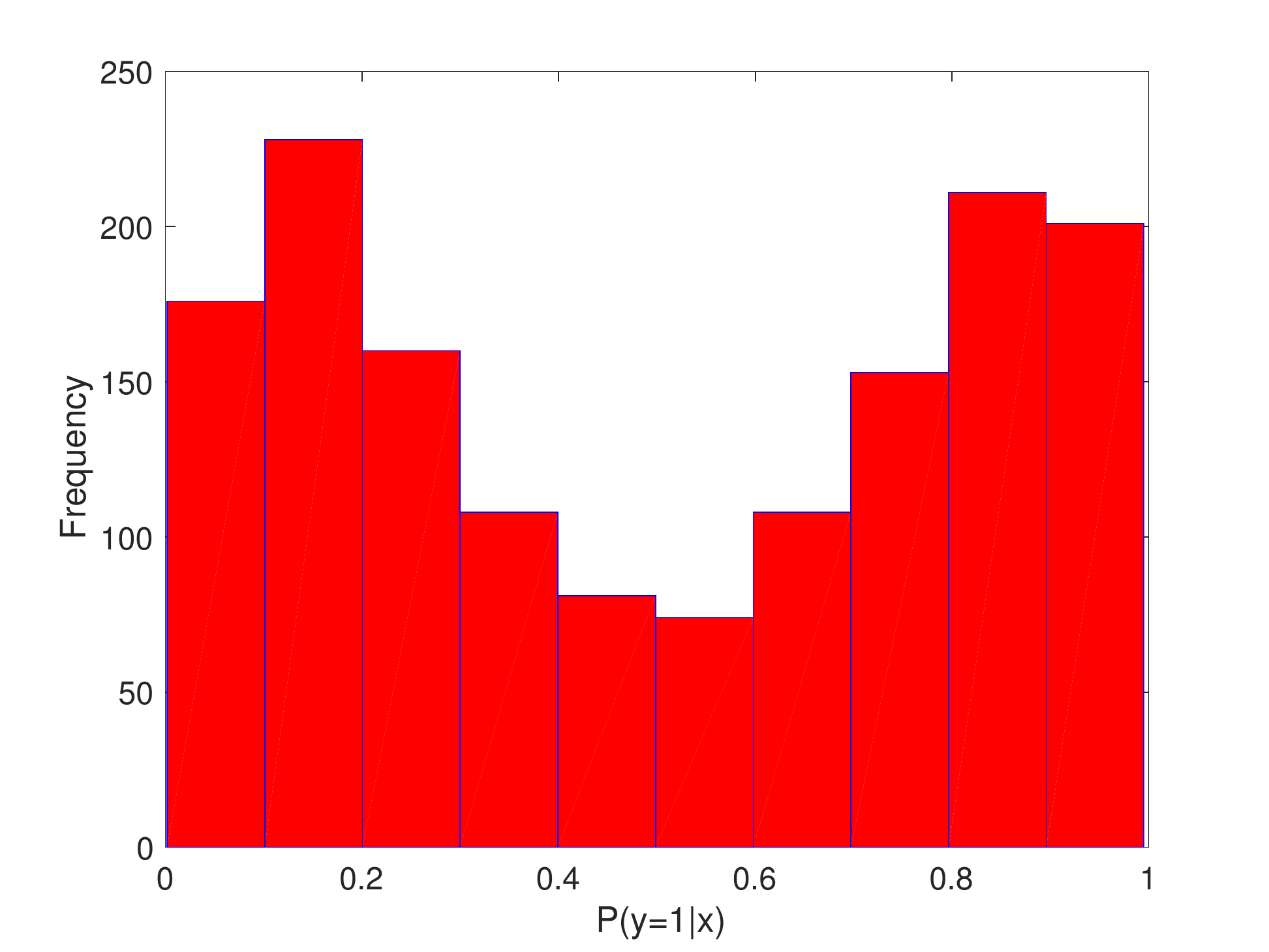}\label{fig:histPredProb:a}}
		&
		\subfloat[GP LIN MKL-AsRD Slice]{\includegraphics[width=0.33\linewidth]{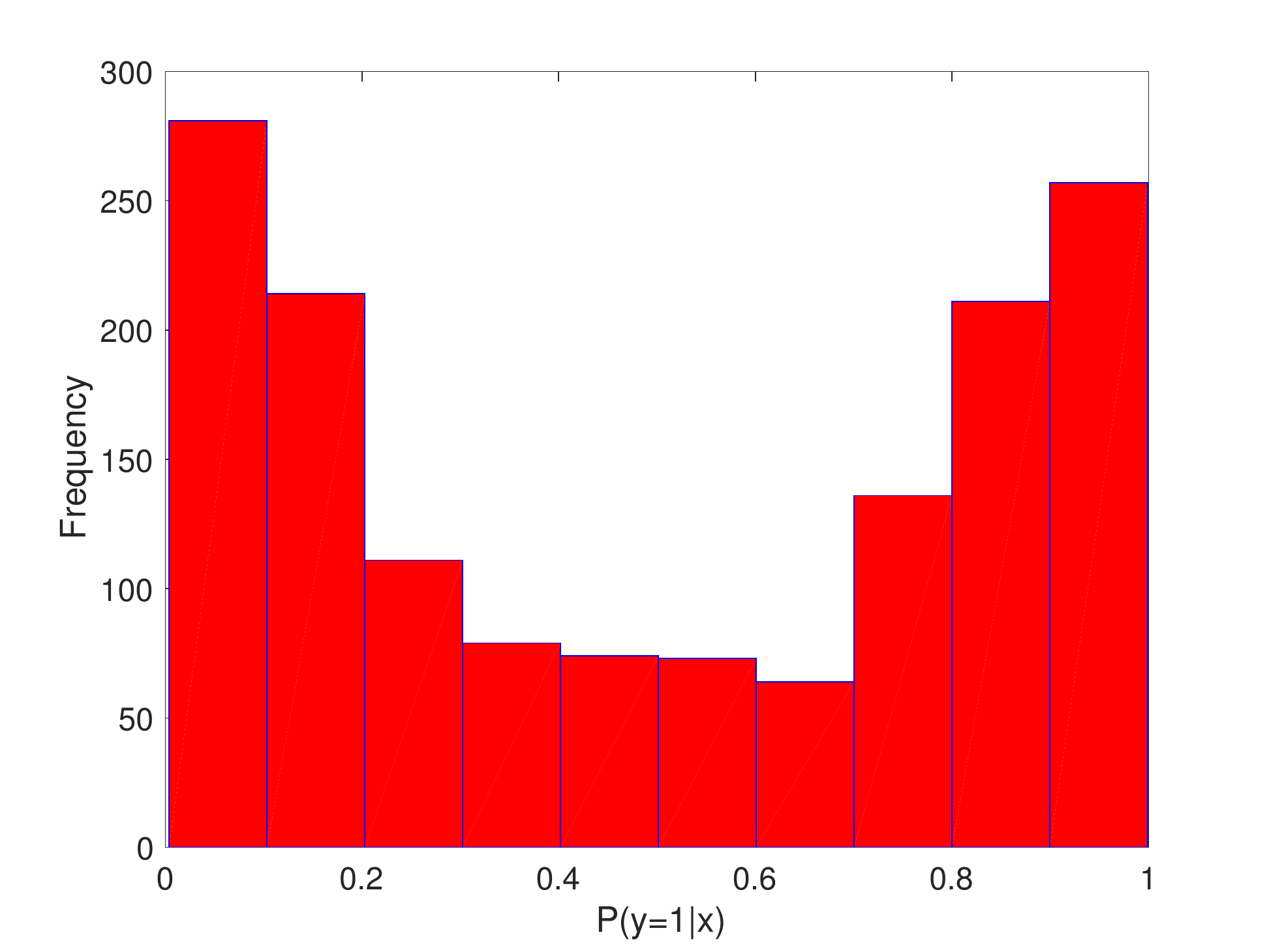}\label{fig:histPredProb:b}}
		&
		\subfloat[GP LIN MKL-AsRD Cube]{\includegraphics[width=0.33\linewidth]{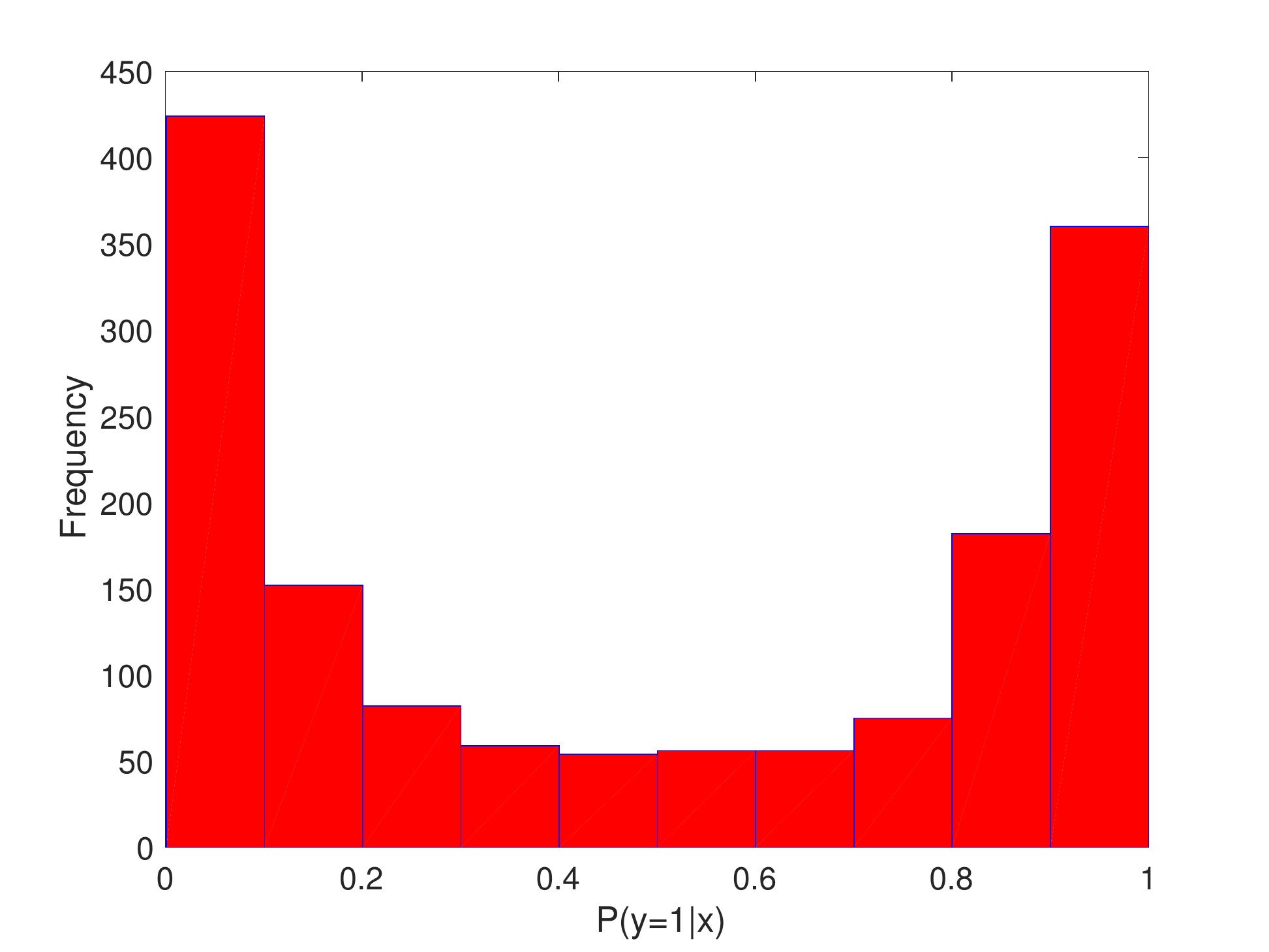}\label{fig:histPredProb:c}}
		\\
		\subfloat[GP SE]{\includegraphics[width=0.33\linewidth]{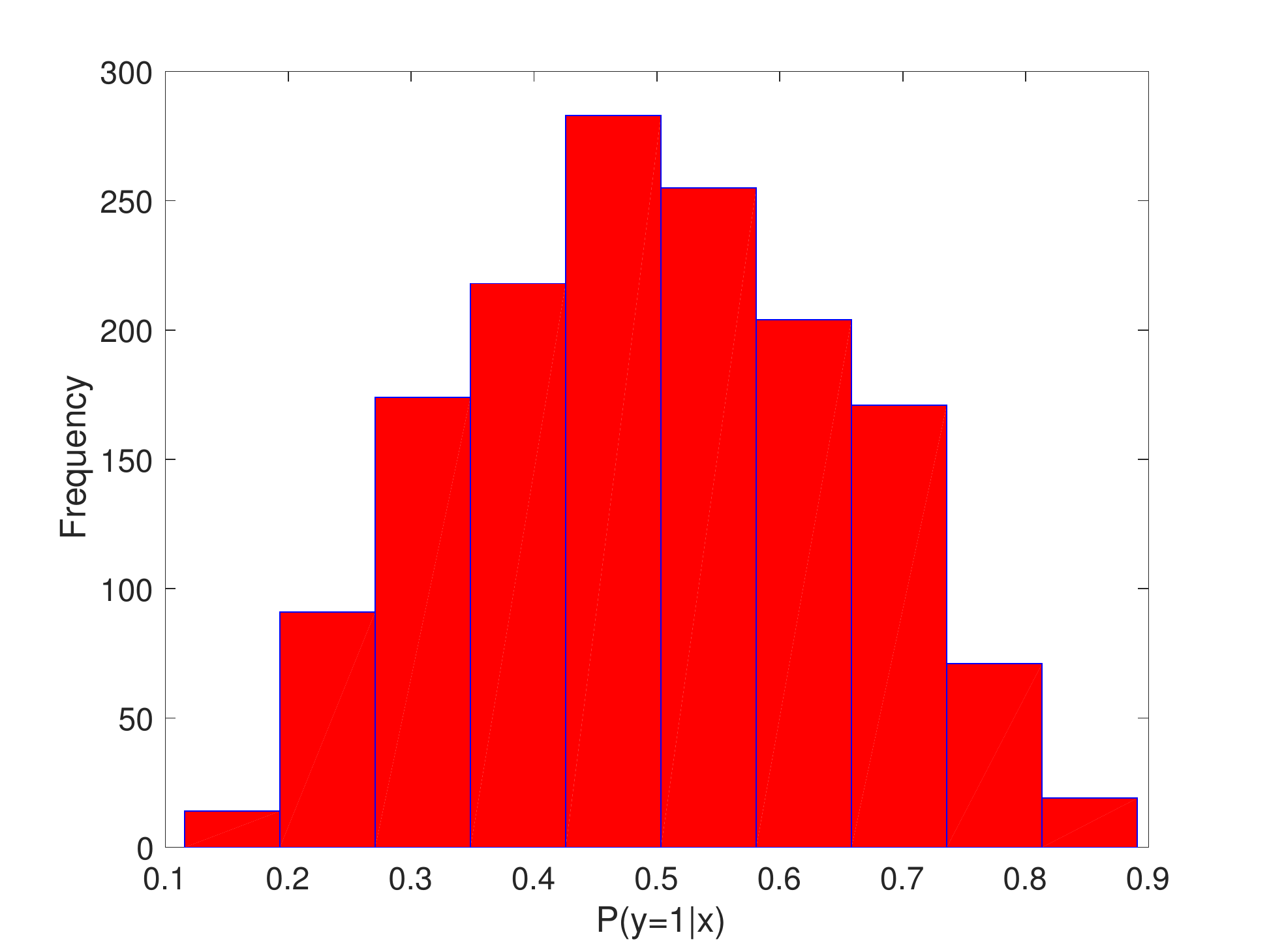}\label{fig:histPredProb:d}}
		&
		\subfloat[GP SE MKL-AsRD Slice]{\includegraphics[width=0.33\linewidth]{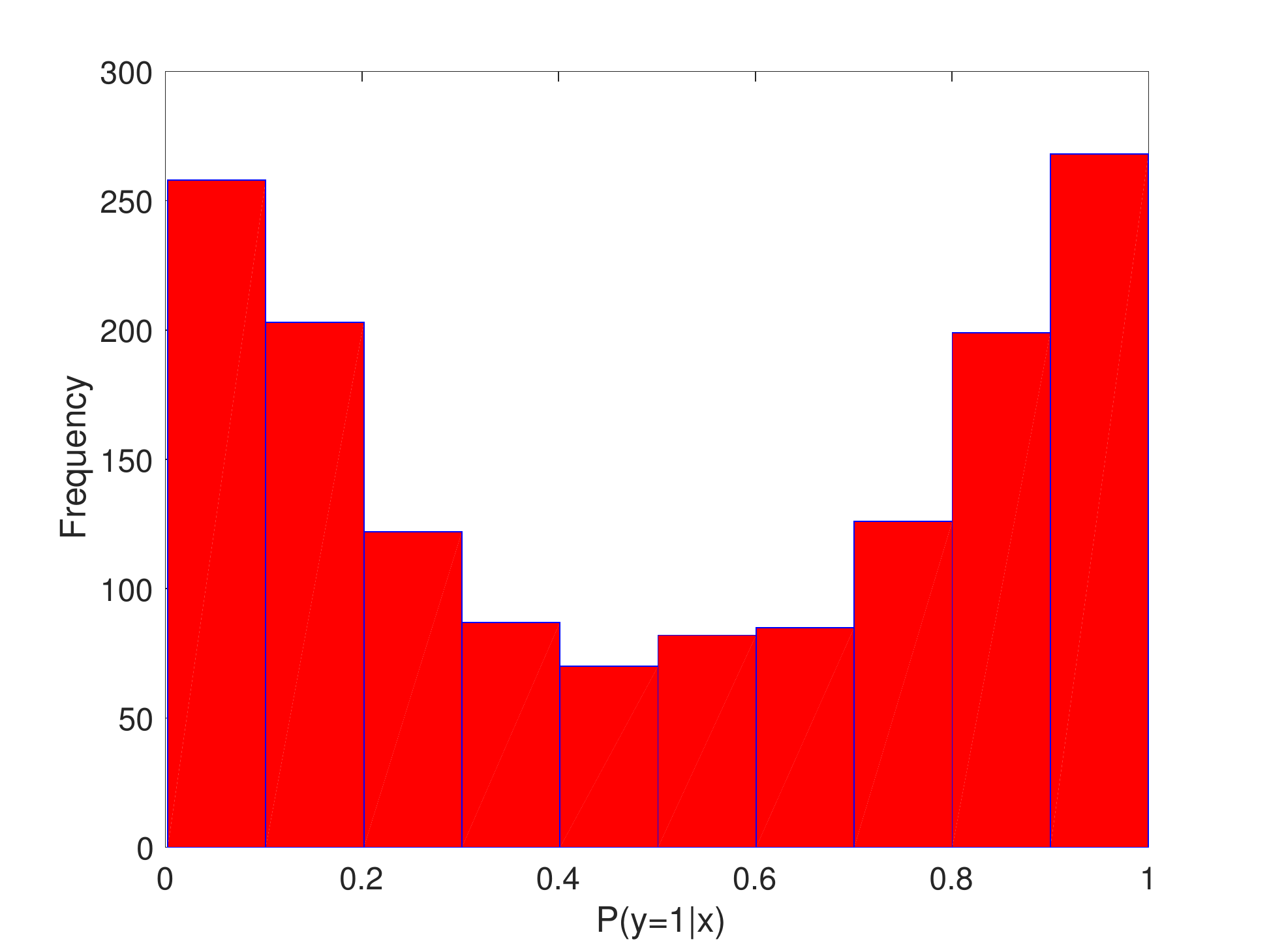}\label{fig:histPredProb:e}}
		&
		\subfloat[GP SE MKL-AsRD Cube]{\includegraphics[width=0.33\linewidth]{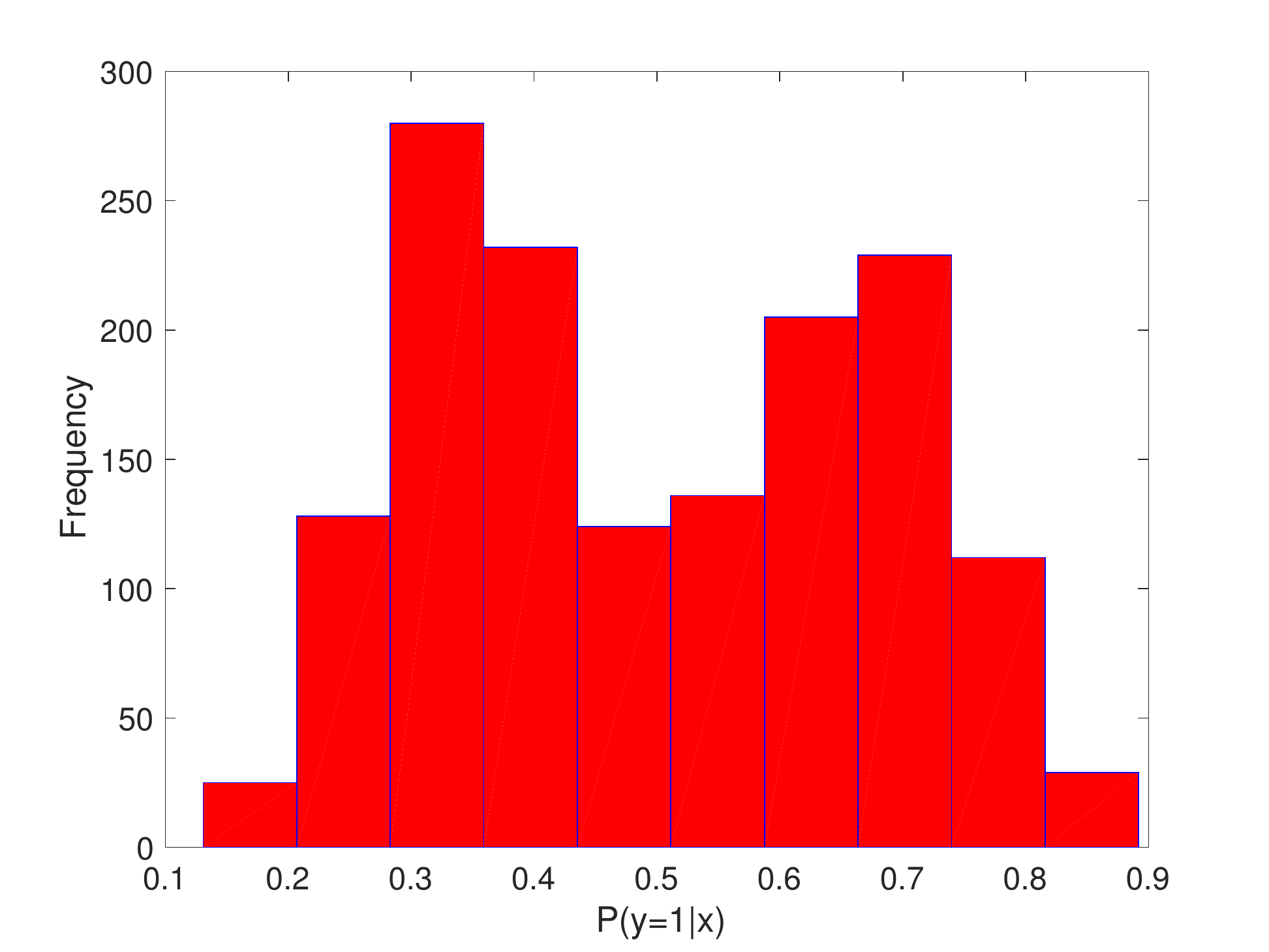}\label{fig:histPredProb:f}}
		\\
		\subfloat[GP NN]{\includegraphics[width=0.33\linewidth]{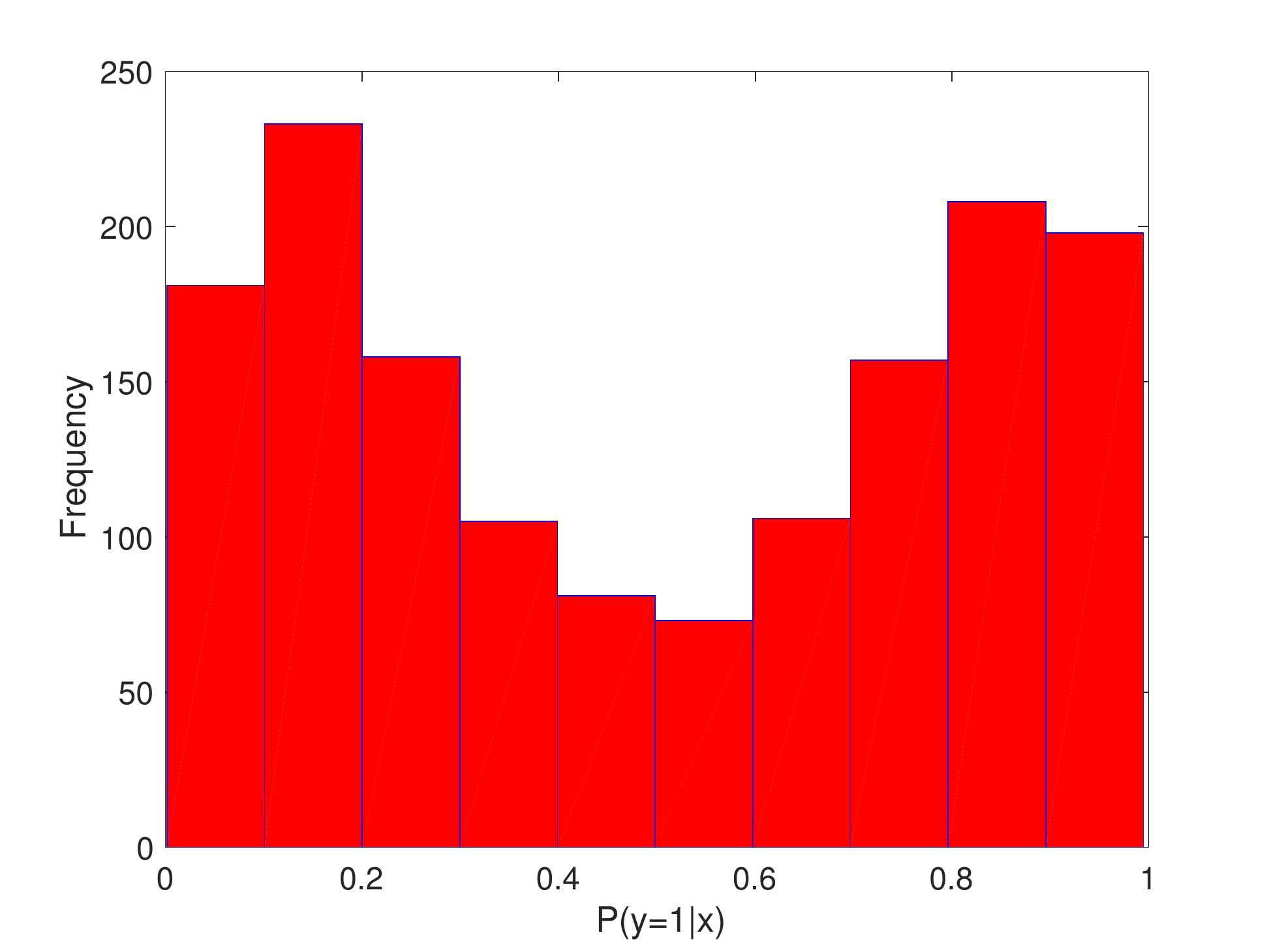}\label{fig:histPredProb:g}}
		&
		\subfloat[GP NN MKL-AsRD Slice]{\includegraphics[width=0.33\linewidth]{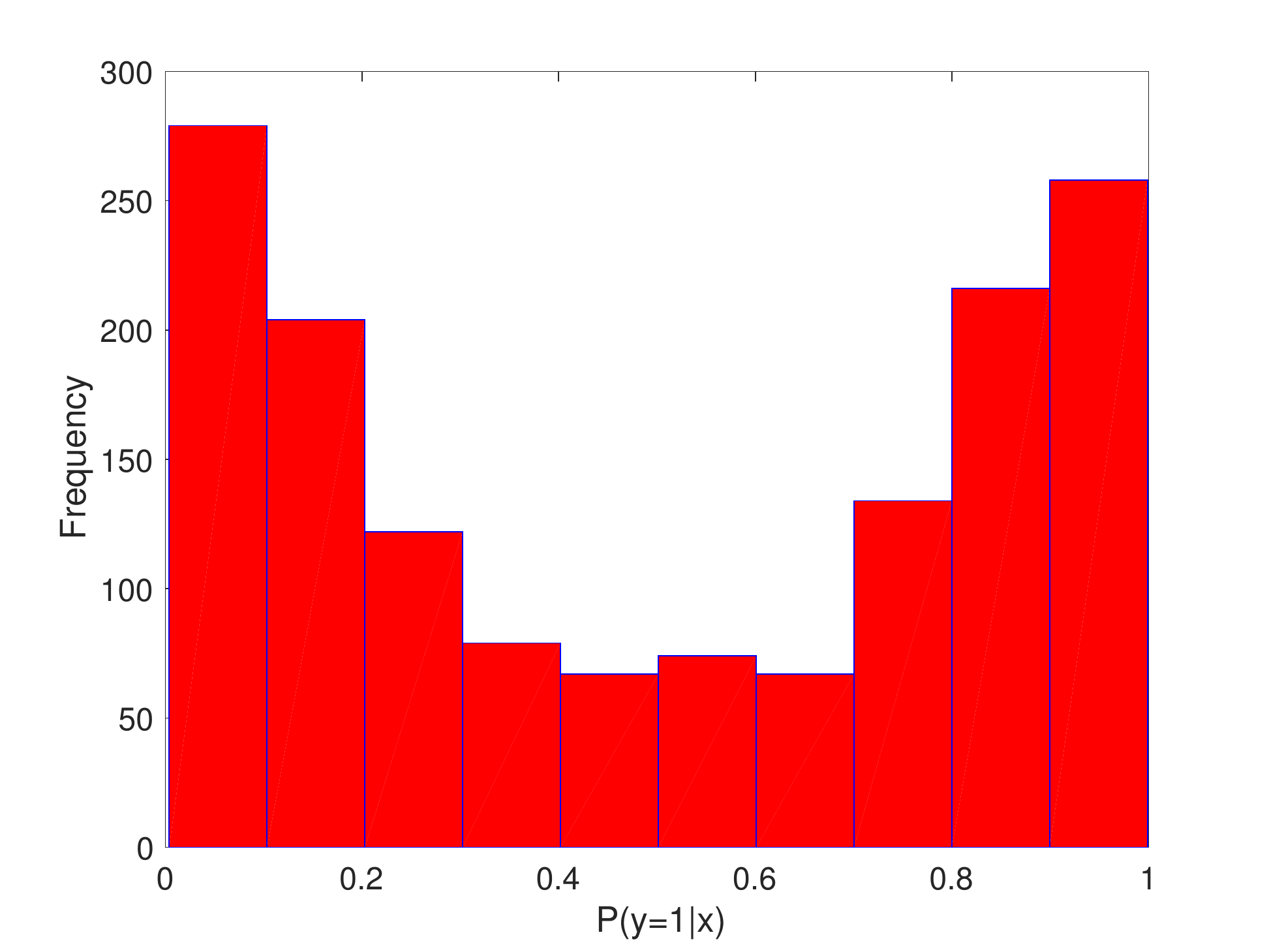}\label{fig:histPredProb:h}}
		&
		\subfloat[GP NN MKL-AsRD Cube]{\includegraphics[width=0.33\linewidth]{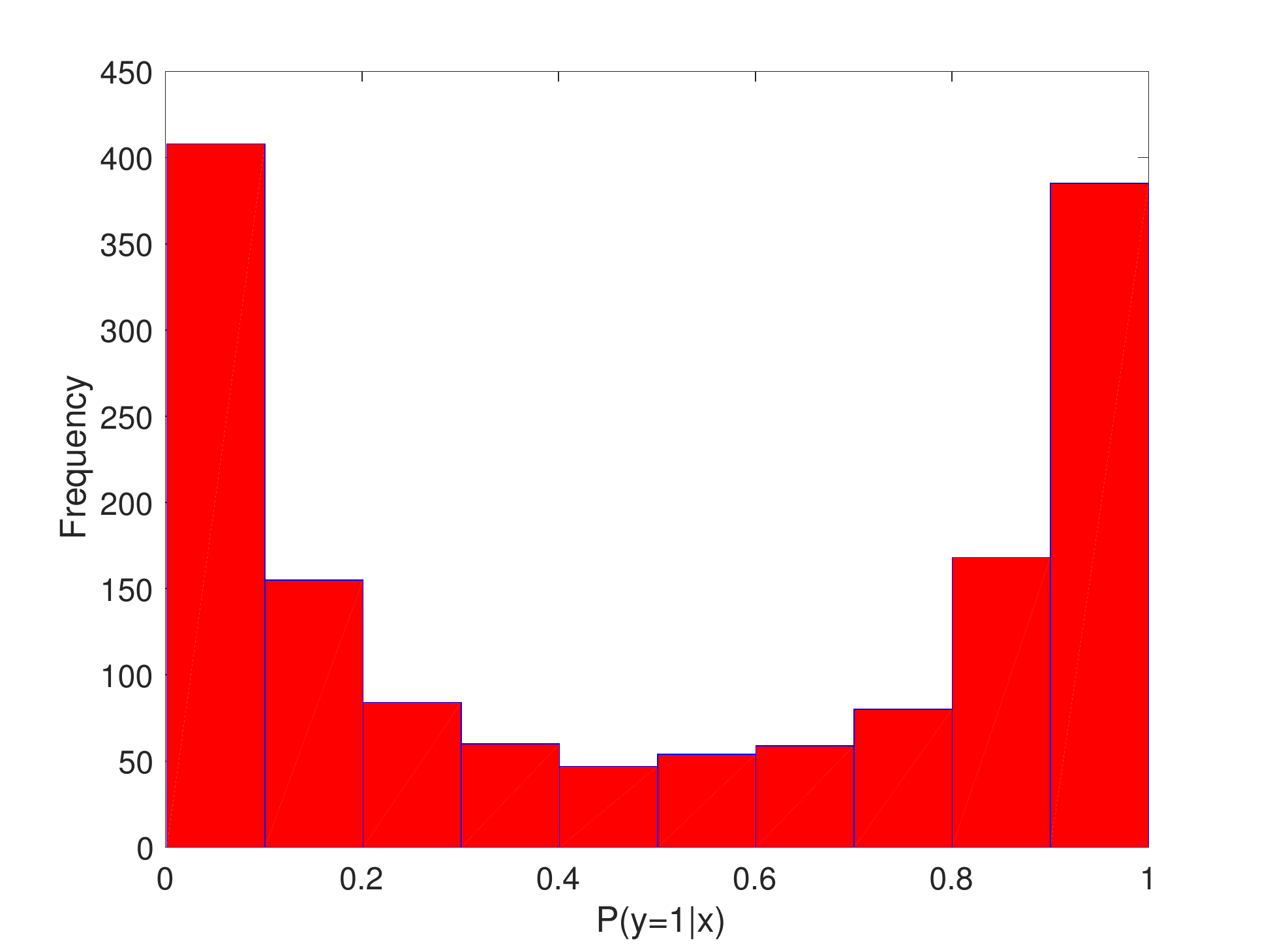}\label{fig:histPredProb:i}}
	\end{tabular}
    \caption[Histograms of predictive probabilities]{Histograms of predictive probabilities via GP  configurations for binary (Normal vs AD) classification task based on all test sets used during \Gls{CV}. 
} 
	\label{fig:histPredProb}
\end{figure}

\subsection{Multi-Class Classification Results}\label{sec:MultiClassResults}

Dementia diagnosis is essentially a \emph{multi-class} problem that can be decomposed into a series of binary classification problems. Probably the most popular schemes are 
\emph{\Gls{OVO}}
and \emph{\Gls{OVA}}. 
The \Gls{OVO} approach gives rise to $\binom{K}{2}$ classifiers, given $K$ classes. Each classifier separates a pair of classes. On the other hand, the \Gls{OVA} approach requires $K$ classifiers, each of which distinguishes a class from the others. 
\Gls{OVO} and \Gls{OVA} are competitive with other more complicated methods like 
single-machine schemes \citep{Rifkin2004,Huang2006}. 
Moreover, 
\citet{Rifkin2004} reported that 
``a simple \Gls{OVA} scheme is as accurate as any other approach, 
	assuming that the underlying binary classifiers are well-tuned regularized classifiers 
	such as \Glspl{SVM}'' \citep[p.1]{Rifkin2004}. 
%
%
Since we have established that the \Gls{GP} models are robust, high-performance classifiers, we use them for multi-class classification, 
adopting the \Gls{OVA} scheme.

Table~\ref{table_MultiClassResults} and Figure~\ref{fig:multiClassComparison} present the multi-class classification performances of the previously described configurations. The trend is similar to those observed with the binary classification experiments. \Glspl{SVM} set a fairly high baseline for the \Gls{GP} models to match. 
As one would expect, the \Gls{GP} models with a single \Gls{SE} covariance function perform poorly. However, \Gls{MKLARD} restores the modeling power of \Glspl{GP}, and two \Gls{GP} configurations, \Gls{GP} LIN \Gls{MKLARD} Slice and \Gls{GP} \Gls{NN} \Gls{MKLARD} Cube, significantly outperform the baseline \Gls{SVM}. 
The striking fact from Table~\ref{table_MultiClassResults} is that all classifiers struggle with the \Gls{MCI} class. The underlying classifiers trained for the discrimination of \Gls{MCI} from others 
always deliver low sensitivity but high specificity, which imposes a limitation on the multi-class accuracy. This phenomenon is also indicative of the large overlap between the \Gls{MCI} class and others, which has been reported by \citep{Fan2008}.

\begin{table}[h!]
\centering
\caption[Multi-class classification performances]
	{Multi-class classification performances via \Gls{OVA} approach. 
	For the GP models that utilize the SE covariance function, LA was used for inference. All other GP models were obtained via EP. 
	Optimal parameters for the underlying binary SVMs were determined by a grid-search per SVM.
	}
\label{table_MultiClassResults}
\begin{tabular}{@{\extracolsep{\fill}}p{3mm}p{18mm}p{13mm}p{23mm}p{23mm}p{23mm}p{20mm}}
\toprule
								&	\textsl{Classifier}			&	$+$ class				&	\textsl{Binary \newline Accuracy}	&	\textsl{Sensitivity}	&	\textsl{Specificity}	&	\textsl{Multi-class \newline Accuracy}	\\
\midrule
\multirow{21}{*}{\begin{sideways}Single Kernel\end{sideways}}	
								&	\multirow{3}{*}{SVM LIN}	&	AD						&	89.42 (2.04)			&	86.13 (5.59)			&   91.07 (2.25)			&	\multirow{3}{*}{80.22 (3.05)}			\\
								&								&	MCI						&	84.04 (3.03)			&	72.40 (4.49)			&   89.87 (3.62)			&											\\
								&								&	Normal					&	86.98 (2.56)			&	82.13 (4.58)			&   89.40 (2.50)			&											\\
\cmidrule{2-7}
								&	\multirow{3}{*}{SVM\textsuperscript{prob} LIN}	&	AD						&	89.33 (1.93)			&	85.87 (5.70)			&   91.07 (2.52)			&	\multirow{3}{*}{80.40 (3.48)}			\\
								&								&	MCI						&	83.96 (3.60)			&	72.93 (4.83)			&   89.47 (4.18)			&											\\
								&								&	Normal					&	87.51 (2.91)			&	82.40 (4.57)			&   90.07 (2.69)			&											\\
\cmidrule{2-7}
								&	\multirow{3}{*}{SVM RBF}	&	AD						&	90.89 (1.26)			&	88.53 (3.03)			&   \textbf{92.07 (2.00)}		&	\multirow{3}{*}{83.02 (1.69)}			\\
								&								&	MCI						&	86.27 (2.18)			&	75.60 (5.11)			&   91.60 (2.81)			&											\\
								&								&	Normal					&	88.89 (1.39)			&	84.93 (4.62)			&   90.87 (2.63)			&											\\
\cmidrule{2-7}
								&	\multirow{3}{*}{SVM\textsuperscript{prob} RBF}	&	AD						&	90.09 (1.72)			&	86.93 (4.44)			&   91.67 (1.76)			&	\multirow{3}{*}{82.18 (3.11)}			\\
								&								&	MCI						&	85.16 (3.23)			&	\textbf{75.60 (4.99)}			&   89.93 (4.03)			&											\\
								&								&	Normal					&	89.11 (2.08)			&	84.00 (4.91)			&   \textbf{91.67 (2.71)}			&											\\
\cmidrule{2-7}
								&	\multirow{3}{*}{GP LIN}	&	AD						&	\textbf{91.20 (1.58)}			&	\textbf{90.00 (4.37)}			&   91.80 (1.44)			&	\multirow{3}{*}{83.73 (1.38)}			\\
								&								&	MCI						&	86.84 (1.21)			&	73.33 (3.77)			&   \textbf{93.60 (2.40)}			&											\\
								&								&	Normal					&	89.42 (1.27)			&	\textbf{87.87 (4.46)}			&   90.20 (2.74)			&											\\
\cmidrule{2-7}
								&	\multirow{3}{*}{GP SE}		&	AD						&	81.42 (6.98)			&	76.93 (31.78)			&	83.67 (10.36)			&	\multirow{3}{*}{64.44 (8.94)}			\\
								&								&	MCI						&	69.91 (8.38)			&	43.73 (31.99)			&   83.00 (18.55)			&											\\
								&								&	Normal					&	77.56 (6.48)			&	72.67 (30.83)			&   80.00 (11.37)			&											\\
\cmidrule{2-7}
								&	\multirow{3}{*}{GP NN}		&	AD						&	\textbf{91.20 (1.58)}			&	\textbf{90.00 (4.37)}			&   91.80 (1.44)			&	\multirow{3}{*}{\textbf{83.78 (1.38)}}			\\
								&								&	MCI						&	\textbf{86.89 (1.19)}			&	73.47 (3.47)			&	\textbf{93.60 (2.40)}			&											\\
								&								&	Normal					&	\textbf{89.47 (1.31)}			&	\textbf{87.87 (4.46)}			&   90.27 (2.69)			&											\\
\midrule\midrule
\multirow{9}{*}{\begin{sideways}\Gls{MKLARD} (Slices)\end{sideways}}	
								&	\multirow{3}{*}{GP LIN}	&	AD						&	\textbf{91.91 (1.25)}			&	90.27 (4.40)			&   \textbf{92.73 (1.31)}			&	\multirow{3}{*}{\textbf{85.64 (1.75)}}			\\
								&								&	MCI						&	\textbf{88.44 (1.15)}			&	\textbf{77.87 (3.40)}			&   \textbf{93.73 (2.36)}		&											\\
								&								&	Normal					&	90.93 (2.21)			&	\textbf{88.80 (3.99)}			&   92.00 (2.29)			&											\\
\cmidrule{2-7}
								&	\multirow{3}{*}{GP SE}		&	AD						&	90.67 (2.67)			&   87.73 (5.99)			&	92.13 (1.96)			&	\multirow{3}{*}{83.51 (3.16)}			\\
								&								&	MCI						&	86.98 (2.11)			&   74.67 (5.11)			&   93.13 (3.31)			&											\\
								&								&	Normal					&	89.38 (2.05)			&   88.13 (5.05)			&   90.00 (2.39)			&											\\
\cmidrule{2-7}
								&	\multirow{3}{*}{GP NN}		&	AD						&	91.73 (1.64)			&   \textbf{90.40 (4.35)}			&   92.40 (1.48)			&	\multirow{3}{*}{85.24 (1.94)}			\\
								&								&	MCI						&	87.82 (1.28)			&	76.80 (4.45)			&   93.33 (2.37)			&											\\
								&								&	Normal					&	\textbf{90.93 (2.18)}			&	88.53 (3.99)			&	\textbf{92.13 (2.75)}		&											\\
\midrule\midrule
\multirow{9}{*}{\begin{sideways}\Gls{MKLARD} (Cubes)\end{sideways}}	
								&	\multirow{3}{*}{GP LIN}	&	AD						&	91.38 (1.19)			&	90.40 (4.30)			&   91.87 (1.88)			&	\multirow{3}{*}{85.38 (2.22)}			\\
								&								&	MCI						&	88.53 (2.67)			&	76.27 (7.24)			&   \textbf{94.67 (2.24)}			&											\\
								&								&	Normal					&	90.84 (2.00)			&	89.47 (4.24)			&   91.53 (3.24)			&											\\
\cmidrule{2-7}
								&	\multirow{3}{*}{GP SE}		&	AD						&	90.44 (1.74)			&	89.33 (4.91)			&	91.00 (2.58)			&	\multirow{3}{*}{84.49 (2.64)}			\\
								&								&	MCI						&	87.33 (2.30)			&	74.67 (5.86)			&	93.67 (2.40)			&											\\
								&								&	Normal					&	91.20 (2.48)			&	89.47 (4.28)			&	\textbf{92.07 (3.42)}		&											\\
\cmidrule{2-7}
								&	\multirow{3}{*}{GP NN}		&	AD						&	\textbf{92.67 (1.50)}		&   \textbf{90.80 (2.70)}			&   \textbf{93.60 (2.07)}			&	\multirow{3}{*}{\textbf{86.71 (2.20)}}			\\
								&								&	MCI						&	\textbf{89.24 (2.50)}			&	\textbf{78.53 (6.17)}			&	94.60 (1.87)			&											\\
								&								&	Normal					&	\textbf{91.51 (2.11)}			&   \textbf{90.80 (3.52)}			&   91.87 (2.59)			&											\\
\bottomrule
\end{tabular}
\end{table}

\begin{figure}[h!]
	\centering
	\includegraphics[width=0.65\linewidth]{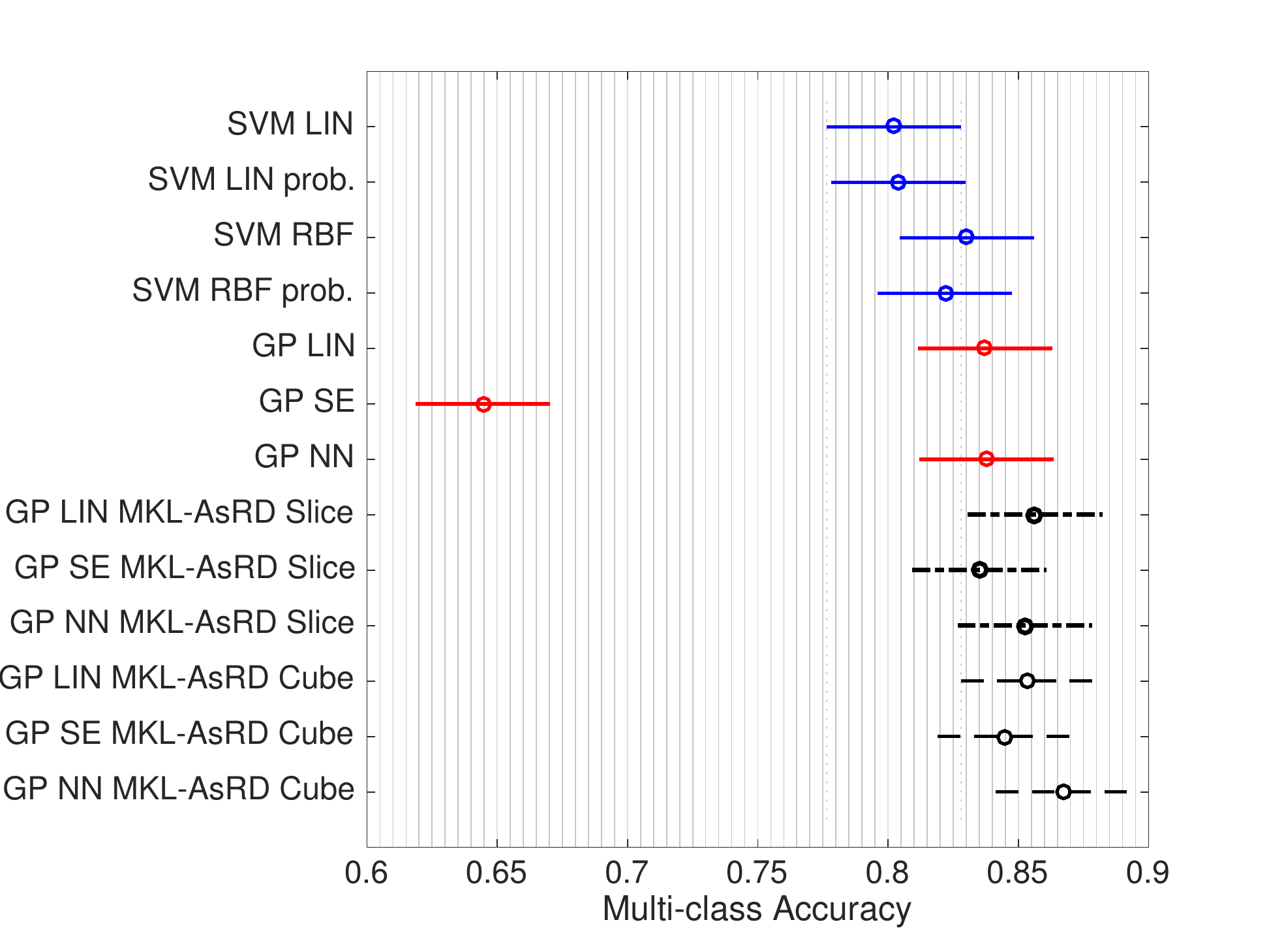}
    \caption[Average (mean) multi-class classification accuracies and comparison intervals.]{Average (mean) multi-class classification accuracies and comparison intervals. The confidence level is 95\%. Two means are significantly different if their comparison intervals do not overlap. 
}
	\label{fig:multiClassComparison}
\end{figure}

\section{Discussion}\label{sec:discussion}
\Gls{MKL} generalizes the use of kernels and enables sophisticated extensions to the capabilities of kernel machines. A particular example is the multimodal classification, for which each modality
is assigned its own kernel or kernels. 
\citet{Hinrichs2009} showed the benefits of the multimodal classification by simultaneously using \Gls{PET} and \Gls{MRI} data. 
Since each imaging procedure captures different aspects of the underlying disease pathology, a linear combination of them is more informative for \Gls{AD} diagnosis \citep{Hinrichs2009}. As a result, the multimodal \Gls{SVM}
outperforms the conventional one that uses only one modality at a time \citep{Hinrichs2009}. 
%
\citet{Zhang2011} considered three modalities obtained from \Gls{PET}, \Gls{MRI} and \gls{CSF}. Using an \Gls{SVM}, they, too, showed that \Gls{MKL} is superior to the single kernel learning. However, they used a grid search to determine the optimal mixing weights, which resulted in a secondary layer \gls{CV}. The grid search and second level CV significantly complicates the model selection. 

\citet{Young2013a} 
trained a \Gls{GP} classifier for the discrimination of \Gls{AD} from Normal based on the \Gls{PET}, \Gls{MRI}, \gls{CSF} and \gls{APOE} genotype data and used the model to predict the conversion from \Gls{MCI} to \Gls{AD}. They also compared the \Gls{GP} models to \Glspl{SVM} on the basis of their classification performances. Both kernel-machines were equipped with linear kernels 
and the \Gls{GP} models significantly outperformed the \Glspl{SVM} on the classification tasks involving multiple modalities \citep{Young2013a}. 
In addition, 
they regarded the probabilistic outputs as risk scores for conversion to \Gls{AD} and approximated the continuum with respect to discrete stages of the disease. In a follow-up study, \citet{Young2013b} demonstrated the use of real-valued predictions for inferring the rate of conversion from \Gls{MCI} to \Gls{AD}. The elimination of discrete class labels increases the robustness to mislabeling. But, both the classification and regression approaches require data from multiple modalities. As a result, the final training set size is determined by the modality with the smallest number of examples \citep{Young2013b}. The limitation on training data is a serious drawback that contaminates other multimodal approaches as well. 

In our work, we exploit the power of \Gls{MKL}, without using multiple data modalities. Thus, we can use larger datasets in the first place. Secondly, we adopt a different perspective for the deployment of multiple kernels. 
Given a modality, we seek to discover new and implicitly accessible modalities from the given one. To better understand the analogy we make here, one can leave the mathematical sophistication aside and think that each kernel translates a portion of data into a feature space. Then, the features are combined into a very long vector that represents the input image from the given modality. 
Notice that our proposal does not exclude the possibility of the conventional multimodality since it can be adopted for each one separately. 
It is also notable that the previous studies on multi-modality adressed only the binary classification problems, 
i.e., Normal vs. \Gls{AD}, or Normal vs. \Gls{MCI}, 
and small scale kernel combinations, e.g., 3-4 kernels. 
In our experiments, we consider the multi-class classification as well as 
make use of larger scale kernel combinations, e.g., 150 kernels. 

\subsection{Representation Learning via Deep Architectures}\label{sec:RepLearning}

\Gls{AE} is an \emph{unsupervised} learning algorithm that automatically learns features from data. It is typically used to learn a compact, distributed representation of data \citep{Bengio2009}. 
\citet{Gupta2013a} used a \gls{SAE} 
to obtain a set of bases from \emph{natural images}.\footnote{Natural images provide natural inputs to a biological visual system and are assumed to have statistical characteristics that are similar to what the visual system is adapted to \citep{Hyvarinen2009}.} 
Using these natural image bases, they learned new and low dimensional representations of high dimensional \Gls{MRI} imagery by considering individual slices of 3D scans. 
%
The results obtained via the natural image bases are given in Table~\ref{tableComparisonWithOthers}. 
In a follow-up study, \citet{Gupta2013b} investigated the advantages of the sparse, denoising and contractive \glspl{AE}, over the basic \Gls{AE}. 
In terms of classification performance, even the basic \Gls{AE} is competitive with the sophisticated ones, given a large enough basis set \citep{Gupta2013b}. 
However, 
the \Gls{SAE} enabled a new representation via only 100 bases, whereas 
the denoising, contractive and basic \Glspl{AE} required 200, 150 and 350 bases, respectively. 
The representation obtained by the \Gls{SAE} consisted of {61,200} features while the imagery contained {510,340} voxels \citep{Gupta2013a,Gupta2013b}. 


\begin{table}[h!]
	\caption[Summary of the results obtained by our proposed method and others]{Summary of the results obtained by \Gls{MKLARD}, natural images via \Gls{SAE} \citep{Gupta2013a}, stacked \Glspl{AE} \citep{Liu2015}, and the combination of \Gls{SAE} with \Gls{CNN} \citep{Payan2015}. The same \Gls{MRI} dataset was used by us, \citet{Gupta2013a}, and \citet{Payan2015}. The numbers are the estimates of the generalization performances in terms of classification accuracy. A notable exception is the division of \Gls{MCI} into subclasses by \citet{Liu2015}: nc-\Gls{MCI} and c-\Gls{MCI}. nc-\Gls{MCI} corresponds to the patients who have not converted to \Gls{AD}, whereas c-\Gls{MCI} patients did convert within 3 years from the initial screening.}
	\label{tableComparisonWithOthers}
	\begin{center}
		\begin{tabular}{r c c c >{\columncolor[gray]{0.85}}c >{\columncolor[gray]{0.85}}c c c}
		\toprule
				& \multicolumn{2}{c}{MKL-AsRD}	& SAE			& \multicolumn{2}{c}{Stacked AEs}		& \multicolumn{2}{c}{SAE + CNN}	 \\
		\cmidrule{2-8}
Classes			& 2D				&	3D			& Nat.Img.		& MRI only			& MRI+PET			& 2D				& 3D	\\
		\midrule
Normal vs. AD		& 93.20\%			&	93.87\%		& 94.74\%			& 82.59\%				& 91.40\%				& \textbf{95.39\%}	& \textbf{95.39\%}	\\
Normal vs. MCI		& 88.07\%			&	89.93\%		& 86.35\%			& 71.98\%				& 82.10\%				& 90.13\%			& \textbf{92.11\%}	\\
MCI vs. AD		& \textbf{89.40\%}	&	89.13\%		& 88.10\%			& \multicolumn{2}{c}{N/A} 						& 82.24\%			& 86.84\%	\\
		\midrule
Normal, MCI, AD		& 85.64\%		&	86.71\%		& 85.00\%			&					&					& 85.53\%			& \textbf{89.47\%}	\\
Normal, nc-MCI, c-MCI, AD	&		&				&				& 46.30\%				& 53.79\%				&				&	\\
		\midrule
		\midrule
Representation		&	\multicolumn{2}{c}{510,340} 		&	61,200 		& \multicolumn{2}{c}{N/A}						& 489,600			& 486,000	\\
				&	\multicolumn{2}{c}{voxels}			&	features		&					&					& features			& features	\\
		\bottomrule
		\end{tabular}
	\end{center}
\end{table}

\citet{Liu2015} used stacked \Glspl{AE} to learn new representations from both \Gls{MRI} and \Gls{PET} scans simultaneously. The multimodal dataset consisted of only 331 instances. 
Also, instead of learning features directly from the neuroimagery, they extracted 83 \Glspl{ROI} from each modality and applied a feature engineering strategy to compute the first set of features. Then, some of the features were selected via the \emph{elastic net}~\citep{Zou2005} on the basis of their discriminative power before being fed into the \Glspl{AE}. 
The extraction of predetermined \Glspl{ROI} resembles our selection of pseudo regions based on domain-knowledge. However, in our case, features are \emph{implicitly} learned via kernel functions, parameters of which are also automatically tuned with respect to data. 
The use of hand-crafted features by \citet{Liu2015} is at odds with the spirit of automated feature learning. 
The combined effect of the use of a small sample, predetermined \Glspl{ROI}, and hand-crafted features seems to have hindered their work (Table~\ref{tableComparisonWithOthers}, \emph{shaded region}).


Given the nature of neuroimages, 3D \glspl{CNN} offer a great deal of opportunities in terms of obtaining sparsity and discovering many interesting relationships between inputs; however, moving from 2D to 3D significantly complicates the convolutional learning process~\citep{Graham2015}. Simply, 3D convolution requires many more filters in comparison with 2D convolution. Despite one can reduce the number of convolution filters by forbidding them from overlapping,\footnote{\Gls{MKLARD} also avoids overlapping kernels. We will seek for smarter kernel utilizations in the near future.} 
extracting information from small and discrete patches is at odds with the need for the examination of complete brain imagery. Thankfully, \citet{Payan2015} proposed a deep architecture that consists of a \Gls{SAE} and a \Gls{CNN} followed by a feed-forward neural network. Basically, they used a \Gls{SAE} to learn 3D convolution filters. 
Then, these filters were used for feature extraction, which was followed by max-pooling and classification via a feed-forward neural network. \citet{Payan2015} also considered a similar feature extraction pipeline based on 2D filters. However, the 3D approach yields better performance as it better captures the spatial patterns of brain atrophy (Table~\ref{tableComparisonWithOthers}, \emph{the rightmost column}).

Table~\ref{tableComparisonWithOthers} also shows that \Gls{MKLARD} is competitive with the aforementioned deep learning approaches to the computerized diagnosis of \Gls{AD}. Its accuracy for the separation of \Gls{MCI} and \Gls{AD} instances is particularly encouraging for further investigation of \Gls{MKLARD} for early diagnosis and disease-modifying treatments. 
For instance, 
it can be pushed further into a feature learning pipeline in order to eliminate the use of predetermined \Glspl{ROI} (Section~\ref{sec:futureWork}). 

\section{Conclusion}\label{sec:Conclusion}
The proposed \Gls{GP} models are competitive with or better than the \Gls{SVM} which has been the workhorse of the \acrlong{MVPA} of brain data. The added value of the \Gls{GP} models is that they readily provide probabilistic information regarding their predictions. Moreover, the models with \Gls{MKLARD} capability respond to the patterns of \Gls{AD} pathology and emphasize the prominent anatomical regions and their proximities for accurate staging of \Gls{AD}. 
Last but not least, these models can compete with deep learning solutions under similar settings.
Supporting the view of \citet{Kloppel2008}, we consider them highly practical. Carefully trained and validated models can be deployed at neuroimaging centers in order to speed up the diagnostic processes with no compromise of accuracy and support accurate decision-making based on probabilistic reasoning in cases where there is a lack of access to an experienced physician. 

\subsection{Future Directions}\label{sec:futureWork}

\emph{Deep learning} sets a definitive direction, considering its impressive successes over a plethora of applications~ \citep{Hinton2006a,Hinton2006b,Salakhutdinov2007,Le2011,Krizhevsky2012,Szegedy2014,Dosovitskiy2014,Flynn2015}. 
Deep learning is dominated by \emph{\acrlongpl{NN}}\footnote{Neural networks are the first family of algorithms that enabled deep learning. However, the examples of Deep \Glspl{SVM} \citep{Wiering2013} and Deep \Glspl{GP} \citep{Damianou2013} have been proposed.} 
that consist of multiple layers of many hidden units for higher levels of representation. Given that neuroimages are big, naive approaches end in rapid memory blow-up as well as high computation time. In this regard, it is common for deep learning applications to extract information from small patches of larger images~\citep{Le2011,Krizhevsky2012,Gupta2013a,Gupta2013b}. Convolution via patches is also a key for learning representations~\citep{Lecun1998,Szegedy2014,Dosovitskiy2014,Flynn2015}. However, 
3D convolution is a complicated process due to the volumes of neuroimages. 
In this regard, as an immediate extension, 
we would like to combine deep learning methods 
with the GP models proposed as biomarkers. Our goal is to learn new representations of neuroimages from the most relevant portions of data. Considering that a structure observed in brain imagery is dictated by an anatomical as well as functional organization \citep{Cuingnet2013}, we expect to outperform the slice-based representations \citep{Gupta2013a,Gupta2013b,Payan2015} by better exploiting the spatial relationships in data. 
Also, we will eliminate the need for 3D convolution by learning representations from all \Glspl{ROR} at once. Last but not least, this approach can be easily extended to a multimodal scenario, under which multiple data types, e.g., PET and MRI, will be used simultaneously.

\subsubsection*{Acknowledgments}
Data collection and sharing for this project was funded by the \Gls{ADNI} (\Gls{NIH} Grant U01 AG024904) and DOD \Gls{ADNI} (Department of Defense award number W81XWH-12-2-0012). \Gls{ADNI} is funded by the \Gls{NIA}, the National Institute of Biomedical Imaging and Bioengineering, and through generous contributions from the following: AbbVie, \Gls{AA}; Alzheimer's Drug Discovery Foundation; Araclon Biotech; BioClinica, Inc.; Biogen; Bristol-Myers Squibb Company; CereSpir, Inc.; Eisai Inc.; Elan Pharmaceuticals, Inc.; Eli Lilly and Company; EuroImmun; F. Hoffmann-La Roche Ltd and its affiliated company Genentech, Inc.; Fujirebio; GE Healthcare; IXICO Ltd.; Janssen Alzheimer Immunotherapy Research \& Development, LLC.; Johnson \& Johnson Pharmaceutical Research \& Development LLC.; Lumosity; Lundbeck; Merck \& Co., Inc.; Meso Scale Diagnostics, LLC.; NeuroRx Research; Neurotrack Technologies; Novartis Pharmaceuticals Corporation; Pfizer Inc.; Piramal Imaging; Servier; Takeda Pharmaceutical Company; and Transition Therapeutics. The Canadian Institutes of Health Research is providing funds to support \Gls{ADNI} clinical sites in Canada. Private sector contributions are facilitated by the Foundation for the \Gls{NIH} (www.fnih.org). The grantee organization is the Northern California Institute for Research and Education, and the study is coordinated by the Alzheimer's Disease Cooperative Study at the University of California, San Diego. \Gls{ADNI} data are disseminated by the Laboratory for Neuro Imaging at the University of Southern California.

In preparation of this article, we used portions of text from our earlier work: \citep{Ayhan2012a},\citep{Ayhan2013b}, and \citep{Gupta2013a}. Specifically, Section~\ref{sec:GPR}, \ref{sec:GPC} and \ref{sec:OccamsRazor} are extended from the corresponding sections of our earlier work \citep{Ayhan2012a,Ayhan2013b}. The extended material offers a comprehensive review of the \Gls{GP} learning and a comparison of the \Gls{GP} classifier with \Gls{SVM}. Given that our work relies on the principle of Occam's razor, we reiterated the principle by using text from \citep{Ayhan2012a} and elaborated on the idea towards \Gls{MKLARD}. 
This study also shares data with \citep{Gupta2013a} but we test different hypotheses and our experimental setup is unique to us. We thank our collaborators for their contribution that led us to this study.

\begin{spacing}{0.8}
\small
\bibliographystyle{abbrvnat}
\bibliography{MKLARD}

\begin{thebibliography}{84}
\providecommand{\natexlab}[1]{#1}
\providecommand{\url}[1]{\texttt{#1}}
\expandafter\ifx\csname urlstyle\endcsname\relax
  \providecommand{\doi}[1]{doi: #1}\else
  \providecommand{\doi}{doi: \begingroup \urlstyle{rm}\Url}\fi

\bibitem[{A}lzheimer's {A}ssociation(2017)]{ADfacts2017}
{A}lzheimer's {A}ssociation.
\newblock 2017 {A}lzheimer's disease facts and figures.
\newblock \emph{{A}lzheimer's \& {D}ementia}, 13\penalty0 (4):\penalty0 325 --
  373, 2017.
\newblock ISSN 1552-5260.
\newblock \doi{https://doi.org/10.1016/j.jalz.2017.02.001}.
\newblock URL
  \url{http://www.sciencedirect.com/science/article/pii/S1552526017300511}.

\bibitem[Archambeau and Bach(2011)]{Archambeau2011}
C.~Archambeau and F.~Bach.
\newblock Multiple {G}aussian process models.
\newblock \emph{arXiv preprint arXiv:1110.5238}, 2011.

\bibitem[Ashburner et~al.(2010)Ashburner, Barnes, Chen, Daunizeau, Flandin,
  Friston, Gitelman, Kiebel, Kilner, Litvak, Moran, Penny, Stephan, Gitelman,
  Henson, Hutton, Glauche, Mattout, and Phillips]{SPM8manual}
J.~Ashburner, G.~Barnes, C.~Chen, J.~Daunizeau, G.~Flandin, K.~Friston,
  D.~Gitelman, S.~Kiebel, J.~Kilner, V.~Litvak, R.~Moran, W.~Penny, K.~Stephan,
  D.~Gitelman, R.~Henson, C.~Hutton, V.~Glauche, J.~Mattout, and C.~Phillips.
\newblock \emph{{SPM8 manual}}, July 2010.

\bibitem[Ayhan et~al.(2010)Ayhan, Benton, Raghavan, and Choubey]{Ayhan2010}
M.~S. Ayhan, R.~G. Benton, V.~V. Raghavan, and S.~Choubey.
\newblock Exploitation of 3{D} stereotactic surface projection for automated
  classification of {A}lzheimer's disease according to dementia levels.
\newblock In \emph{2010 IEEE International Conference on Bioinformatics and
  Biomedicine (BIBM)}, pages 516--519. IEEE, 2010.

\bibitem[Ayhan et~al.(2012)Ayhan, Benton, Choubey, and Raghavan]{Ayhan2012a}
M.~S. Ayhan, R.~G. Benton, S.~Choubey, and V.~V. Raghavan.
\newblock Utilization of domain-knowledge for simplicity and comprehensibility
  in predictive modeling of {A}lzheimer's disease.
\newblock In \emph{Proceedings of the 2012 IEEE International Conference on
  Bioinformatics and Biomedicine Workshops (BIBMW)}, BIBMW '12, pages 265--272,
  2012.
\newblock ISBN 978-1-4673-2746-6.

\bibitem[Ayhan et~al.(2013{\natexlab{a}})Ayhan, Benton, Raghavan, and
  Choubey]{Ayhan2013a}
M.~S. Ayhan, R.~G. Benton, V.~V. Raghavan, and S.~Choubey.
\newblock Exploitation of 3{D} stereotactic surface projection for predictive
  modelling of {A}lzheimer's disease.
\newblock \emph{Int. J. Data Min. Bioinformatics}, 7\penalty0 (2):\penalty0
  146--165, Apr. 2013{\natexlab{a}}.
\newblock ISSN 1748-5673.

\bibitem[Ayhan et~al.(2013{\natexlab{b}})Ayhan, Benton, Raghavan, and
  Choubey]{Ayhan2013b}
M.~S. Ayhan, R.~G. Benton, V.~V. Raghavan, and S.~Choubey.
\newblock Composite kernels for automatic relevance determination in
  computerized diagnosis of alzheimer’s disease.
\newblock In \emph{International Conference on Brain and Health Informatics},
  pages 126--137. Springer, 2013{\natexlab{b}}.

\bibitem[Bach(2008)]{Bach2008}
F.~R. Bach.
\newblock Consistency of the group lasso and multiple kernel learning.
\newblock \emph{J. Mach. Learn. Res.}, 9:\penalty0 1179--1225, June 2008.
\newblock ISSN 1532-4435.
\newblock URL \url{http://dl.acm.org/citation.cfm?id=1390681.1390721}.

\bibitem[Bach et~al.(2004)Bach, Lanckriet, and Jordan]{Bach2004}
F.~R. Bach, G.~R.~G. Lanckriet, and M.~I. Jordan.
\newblock Multiple kernel learning, conic duality, and the {SMO} algorithm.
\newblock In \emph{Proceedings of the Twenty-first International Conference on
  Machine Learning}, ICML '04, pages 6--13, 2004.
\newblock ISBN 1-58113-838-5.

\bibitem[Bengio(2009)]{Bengio2009}
Y.~Bengio.
\newblock Learning deep architectures for {AI}.
\newblock \emph{Found. Trends Mach. Learn.}, 2\penalty0 (1):\penalty0 1--127,
  Jan. 2009.
\newblock ISSN 1935-8237.

\bibitem[Bishop and Tipping(2003)]{Bishop2003}
C.~M. Bishop and M.~E. Tipping.
\newblock Bayesian regression and classification.
\newblock In J.~Suykens, G.~Horvath, S.~Basu, C.~Micchelli, and J.~Vandewalle,
  editors, \emph{Advances in Learning Theory: Methods, Models and
  Applications}, volume 190 of \emph{NATO Science Series III: Computer and
  Systems Sciences}, pages 267--285. IOS Press, 2003.

\bibitem[Candela(2004)]{Candela2004}
J.~Q. Candela.
\newblock \emph{Learning with uncertainty-{G}aussian processes and relevance
  vector machines}.
\newblock PhD thesis, Technical University of Denmark, 2004.

\bibitem[Caruana(1998)]{Caruana1998}
R.~Caruana.
\newblock Multitask {L}earning.
\newblock In \emph{Learning to {L}earn}, pages 95--133. Springer, 1998.

\bibitem[Chang and Lin(2011)]{Chang2011}
C.-C. Chang and C.-J. Lin.
\newblock {LIBSVM}: A library for support vector machines.
\newblock \emph{ACM Trans. Intell. Syst. Technol.}, 2\penalty0 (3):\penalty0
  27:1--27:27, May 2011.
\newblock ISSN 2157-6904.

\bibitem[Christoudias et~al.(2009)Christoudias, Urtasun, and
  Darrell]{Christoudias2009}
M.~Christoudias, R.~Urtasun, and T.~Darrell.
\newblock Bayesian localized multiple kernel learning.
\newblock Technical report, University of California at Berkeley, 2009.

\bibitem[Cortes et~al.(2009)Cortes, Mohri, and Rostamizadeh]{Cortes2009}
C.~Cortes, M.~Mohri, and A.~Rostamizadeh.
\newblock L2 regularization for learning kernels.
\newblock In \emph{Proceedings of the Twenty-Fifth Conference on Uncertainty in
  Artificial Intelligence}, UAI '09, pages 109--116, 2009.
\newblock ISBN 978-0-9749039-5-8.

\bibitem[Cuingnet et~al.(2013)Cuingnet, Glaunes, Chupin, Benali, and
  Colliot]{Cuingnet2013}
R.~Cuingnet, J.~A. Glaunes, M.~Chupin, H.~Benali, and O.~Colliot.
\newblock Spatial and anatomical regularization of svm: a general framework for
  neuroimaging data.
\newblock \emph{Pattern Analysis and Machine Intelligence, IEEE Transactions
  on}, 35\penalty0 (3):\penalty0 682--696, 2013.

\bibitem[Damianou and Lawrence(2013)]{Damianou2013}
A.~Damianou and N.~Lawrence.
\newblock Deep {G}aussian processes.
\newblock In \emph{Proceedings of the Sixteenth International Conference on
  Artificial Intelligence and Statistics (AISTATS 2013)}, pages 207--215, 2013.

\bibitem[Domingos(1998)]{Domingos1998}
P.~Domingos.
\newblock Occam's two razors: The sharp and the blunt.
\newblock In \emph{In Proceedings of the Fourth International Conference on
  Knowledge Discovery and Data Mining}, pages 37--43. AAAI Press, 1998.

\bibitem[Dosovitskiy et~al.(2014)Dosovitskiy, Springenberg, and
  Brox]{Dosovitskiy2014}
A.~Dosovitskiy, J.~T. Springenberg, and T.~Brox.
\newblock Learning to generate chairs with convolutional neural networks.
\newblock \emph{arXiv preprint arXiv:1411.5928}, 2014.

\bibitem[Duvenaud et~al.(2011)Duvenaud, Nickisch, and Rasmussen]{Duvenaud2011}
D.~K. Duvenaud, H.~Nickisch, and C.~E. Rasmussen.
\newblock Additive {G}aussian processes.
\newblock In J.~Shawe-Taylor, R.~Zemel, P.~Bartlett, F.~Pereira, and
  K.~Weinberger, editors, \emph{Advances in Neural Information Processing
  Systems 24}, pages 226--234. 2011.

\bibitem[Fan et~al.(2008)Fan, Batmanghelich, Clark, and Davatzikos]{Fan2008}
Y.~Fan, N.~Batmanghelich, C.~M. Clark, and C.~Davatzikos.
\newblock Spatial patterns of brain atrophy in {MCI} patients, identified via
  high-dimensional pattern classification, predict subsequent cognitive
  decline.
\newblock \emph{NeuroImage}, 39\penalty0 (4):\penalty0 1731--1743, 2008.
\newblock ISSN 1053-8119.
\newblock \doi{http://dx.doi.org/10.1016/j.neuroimage.2007.10.031}.

\bibitem[Fletcher et~al.()Fletcher, Fletcher, and Fletcher]{Fletcher2012}
R.~Fletcher, S.~Fletcher, and G.~Fletcher.
\newblock \emph{Clinical {E}pidemiology: {T}he {E}ssentials}.
\newblock M - Medicine Series.
\newblock ISBN 9781451144475.

\bibitem[Flynn et~al.(2015)Flynn, Neulander, Philbin, and Snavely]{Flynn2015}
J.~Flynn, I.~Neulander, J.~Philbin, and N.~Snavely.
\newblock Deepstereo: Learning to predict new views from the world's imagery.
\newblock \emph{CoRR}, abs/1506.06825, 2015.

\bibitem[G\"{o}nen(2012)]{Gonen2012}
M.~G\"{o}nen.
\newblock Bayesian efficient multiple kernel learning.
\newblock In J.~Langford and J.~Pineau, editors, \emph{Proceedings of the 29th
  International Conference on Machine Learning (ICML-12)}, ICML '12, pages
  1--8, New York, NY, USA, July 2012. Omnipress.
\newblock ISBN 978-1-4503-1285-1.

\bibitem[G\"{o}nen and Alpaydin(2011)]{Gonen2011}
M.~G\"{o}nen and E.~Alpaydin.
\newblock Multiple kernel learning algorithms.
\newblock \emph{J. Mach. Learn. Res.}, 12:\penalty0 2211--2268, July 2011.
\newblock ISSN 1532-4435.

\bibitem[Gonnella et~al.(1984)Gonnella, Hornbrook, and Louis]{Gonnella1984}
J.~S. Gonnella, M.~C. Hornbrook, and D.~Z. Louis.
\newblock Staging of disease: A case-mix measurement.
\newblock \emph{JAMA}, 251\penalty0 (5):\penalty0 637--644, 1984.
\newblock \doi{10.1001/jama.1984.03340290051021}.

\bibitem[Graham(2015)]{Graham2015}
B.~Graham.
\newblock Sparse 3d convolutional neural networks.
\newblock \emph{arXiv preprint arXiv:1505.02890}, 2015.

\bibitem[Gupta et~al.(2013{\natexlab{a}})Gupta, Ayhan, and Maida]{Gupta2013a}
A.~Gupta, M.~S. Ayhan, and A.~S. Maida.
\newblock Natural image bases to represent neuroimaging data.
\newblock In \emph{Proceedings of the 30th International Conference on Machine
  Learning (ICML-13)}, ICML'13, June 2013{\natexlab{a}}.

\bibitem[Gupta et~al.(2013{\natexlab{b}})Gupta, Ayhan, and Maida]{Gupta2013b}
A.~Gupta, M.~S. Ayhan, and A.~S. Maida.
\newblock Evaluation of autoencoders for bases to represent neuroimaging data.
\newblock In NIPS 2013 Workshop on Machine Learning and Interpretation in
  NeuroImaging, December 2013{\natexlab{b}}.

\bibitem[Hinrichs et~al.(2009)Hinrichs, Singh, Xu, and Johnson]{Hinrichs2009}
C.~Hinrichs, V.~Singh, G.~Xu, and S.~Johnson.
\newblock Mkl for robust multi-modality ad classification.
\newblock In \emph{Medical Image Computing and Computer-Assisted
  Intervention--MICCAI 2009}, pages 786--794. Springer, 2009.

\bibitem[Hinrichs et~al.(2012)Hinrichs, Singh, Peng, and Johnson]{Hinrichs2012}
C.~Hinrichs, V.~Singh, J.~Peng, and S.~C. Johnson.
\newblock {Q-MKL}: Matrix-induced regularization in multi-kernel learning with
  applications to neuroimaging.
\newblock In \emph{NIPS}, pages 1430--1438, 2012.

\bibitem[Hinton and Salakhutdinov(2006)]{Hinton2006a}
G.~E. Hinton and R.~R. Salakhutdinov.
\newblock Reducing the dimensionality of data with neural networks.
\newblock \emph{Science}, 313\penalty0 (5786):\penalty0 504--507, 2006.

\bibitem[Hinton et~al.(2006)Hinton, Osindero, and Teh]{Hinton2006b}
G.~E. Hinton, S.~Osindero, and Y.-W. Teh.
\newblock A fast learning algorithm for deep belief nets.
\newblock \emph{Neural Comput.}, 18\penalty0 (7):\penalty0 1527--1554, July
  2006.
\newblock ISSN 0899-7667.

\bibitem[Huang et~al.(2006)Huang, Weng, and Lin]{Huang2006}
T.-K. Huang, R.~C. Weng, and C.-J. Lin.
\newblock Generalized {B}radley-{T}erry models and multi-class probability
  estimates.
\newblock \emph{J. Mach. Learn. Res.}, 7:\penalty0 85--115, Dec. 2006.
\newblock ISSN 1532-4435.

\bibitem[Hyv{\"a}rinen et~al.(2009)Hyv{\"a}rinen, Hurri, and
  Hoyer]{Hyvarinen2009}
A.~Hyv{\"a}rinen, J.~Hurri, and P.~O. Hoyer.
\newblock \emph{Natural Image Statistics: A Probabilistic Approach to Early
  Computational Vision.}
\newblock Springer Publishing Company, Incorporated, 1st edition, 2009.
\newblock ISBN 1848824904, 9781848824904.

\bibitem[Imabayashi et~al.(2004)Imabayashi, Matsuda, Asada, Ohnishi, Sakamoto,
  Nakano, and Inoue]{Imabayashi2004}
E.~Imabayashi, H.~Matsuda, T.~Asada, T.~Ohnishi, S.~Sakamoto, S.~Nakano, and
  T.~Inoue.
\newblock Superiority of 3-dimensional stereotactic surface projection analysis
  over visual inspection in discrimination of patients with very early
  {A}lzheimer's disease from controls using brain perfusion {SPECT}.
\newblock \emph{Journal of Nuclear Medicine}, 45\penalty0 (9):\penalty0
  1450--1457, 2004.

\bibitem[Kl{\"o}ppel et~al.(2008)Kl{\"o}ppel, Stonnington, Barnes, Chen, Chu,
  Good, Mader, Mitchell, Patel, Roberts, et~al.]{Kloppel2008}
S.~Kl{\"o}ppel, C.~M. Stonnington, J.~Barnes, F.~Chen, C.~Chu, C.~D. Good,
  I.~Mader, L.~A. Mitchell, A.~C. Patel, C.~C. Roberts, et~al.
\newblock Accuracy of dementia diagnosis--a direct comparison between
  radiologists and a computerized method.
\newblock \emph{Brain}, 131\penalty0 (11):\penalty0 2969--2974, 2008.

\bibitem[Krizhevsky et~al.(2012)Krizhevsky, Sutskever, and
  Hinton]{Krizhevsky2012}
A.~Krizhevsky, I.~Sutskever, and G.~E. Hinton.
\newblock Image{N}et classification with deep convolutional neural networks.
\newblock In \emph{NIPS}, volume~1, page~4, 2012.

\bibitem[Kuss and Rasmussen(2005)]{Kuss2005}
M.~Kuss and C.~E. Rasmussen.
\newblock Assessing approximate inference for binary {G}aussian process
  classification.
\newblock \emph{The Journal of Machine Learning Research}, 6:\penalty0
  1679--1704, 2005.

\bibitem[Lanckriet et~al.(2004)Lanckriet, Cristianini, Bartlett, Ghaoui, and
  Jordan]{Lanckriet2004}
G.~R.~G. Lanckriet, N.~Cristianini, P.~Bartlett, L.~E. Ghaoui, and M.~I.
  Jordan.
\newblock Learning the kernel matrix with semidefinite programming.
\newblock \emph{J. Mach. Learn. Res.}, 5:\penalty0 27--72, Dec. 2004.
\newblock ISSN 1532-4435.

\bibitem[Le et~al.(2011)Le, Monga, Devin, Corrado, Chen,
  Marc’Aurelio~Ranzato, and Ng]{Le2011}
Q.~V. Le, R.~Monga, M.~Devin, G.~Corrado, K.~Chen, J.~D.
  Marc’Aurelio~Ranzato, and A.~Y. Ng.
\newblock Building high-level features using large scale unsupervised learning.
\newblock \emph{arXiv preprint arXiv:1112.6209}, 2011.

\bibitem[LeCun et~al.(1998)LeCun, Bottou, Bengio, and Haffner]{Lecun1998}
Y.~LeCun, L.~Bottou, Y.~Bengio, and P.~Haffner.
\newblock Gradient-based learning applied to document recognition.
\newblock \emph{Proceedings of the IEEE}, 86\penalty0 (11):\penalty0
  2278--2324, 1998.

\bibitem[Lin et~al.(2007)Lin, Lin, and Weng]{Lin2007}
H.-T. Lin, C.-J. Lin, and R.~C. Weng.
\newblock A note on {P}latt's probabilistic outputs for support vector
  machines.
\newblock \emph{Mach. Learn.}, 68\penalty0 (3):\penalty0 267--276, Oct. 2007.
\newblock ISSN 0885-6125.

\bibitem[Liu et~al.(2015)Liu, Liu, Cai, Che, Pujol, Kikinis, Feng, and
  Fulham]{Liu2015}
S.~Liu, S.~Liu, W.~Cai, H.~Che, S.~Pujol, R.~Kikinis, D.~Feng, and M.~J.
  Fulham.
\newblock Multimodal neuroimaging feature learning for multiclass diagnosis of
  {A}lzheimer's disease.
\newblock \emph{Biomedical Engineering, IEEE Transactions on}, 62\penalty0
  (4):\penalty0 1132--1140, 2015.

\bibitem[MacKay(1996)]{MacKay1996}
D.~J. MacKay.
\newblock Bayesian methods for backpropagation networks.
\newblock In \emph{Models of neural networks III}, pages 211--254. Springer,
  1996.

\bibitem[Matsuda(2007)]{Matsuda2007}
H.~Matsuda.
\newblock Role of neuroimaging in {A}lzheimer's disease, with emphasis on brain
  perfusion spect.
\newblock \emph{Journal of Nuclear Medicine}, 48\penalty0 (8):\penalty0
  1289--1300, 2007.

\bibitem[Melkumyan and Ramos(2011)]{Melkumyan2011}
A.~Melkumyan and F.~Ramos.
\newblock Multi-kernel {G}aussian processes.
\newblock In \emph{{IJCAI} Proceedings-International Joint Conference on
  Artificial Intelligence}, volume~22, page 1408, 2011.

\bibitem[Mercer(1909)]{Mercer1909}
J.~Mercer.
\newblock Functions of positive and negative type, and their connection with
  the theory of integral equations.
\newblock \emph{Philosophical Transactions of the Royal Society of London.
  Series A, Containing Papers of a Mathematical or Physical Character},
  209:\penalty0 pp. 415--446, 1909.
\newblock ISSN 02643952.

\bibitem[Minka(2001)]{Minka2001PhD}
T.~P. Minka.
\newblock \emph{A family of algorithms for approximate {B}ayesian inference}.
\newblock PhD thesis, Massachusetts Institute of Technology, 2001.

\bibitem[Minoshima et~al.(1995)Minoshima, Frey, Koeppe, Foster, and
  Kuhl]{Minoshima1995}
S.~Minoshima, K.~A. Frey, R.~A. Koeppe, N.~L. Foster, and D.~E. Kuhl.
\newblock A diagnostic approach in {A}lzheimer's disease using
  three-dimensional stereotactic surface projections of fluorine-18-{FDG}
  {PET}.
\newblock \emph{Journal of Nuclear Medicine}, 36\penalty0 (7):\penalty0
  1238--1248, 1995.

\bibitem[Neal(1996)]{Neal1996}
R.~M. Neal.
\newblock \emph{Bayesian Learning for Neural Networks}.
\newblock Springer-Verlag New York, Inc., Secaucus, NJ, USA, 1996.
\newblock ISBN 0387947248.

\bibitem[Nickisch and Rasmussen(2008)]{Nickisch2008}
H.~Nickisch and C.~E. Rasmussen.
\newblock Approximations for binary {G}aussian process classification.
\newblock \emph{Journal of Machine Learning Research}, 9:\penalty0 2035--2078,
  2008.

\bibitem[Payan and Montana(2015)]{Payan2015}
A.~Payan and G.~Montana.
\newblock Predicting {A}lzheimer's disease: a neuroimaging study with {3D}
  convolutional neural networks.
\newblock \emph{arXiv preprint arXiv:1502.02506}, 2015.

\bibitem[Petersen et~al.(2001)Petersen, Doody, Kurz, Mohs, Morris, Rabins,
  Ritchie, Rossor, Thal, and Winblad]{Petersen2001}
R.~C. Petersen, R.~Doody, A.~Kurz, R.~C. Mohs, J.~C. Morris, P.~V. Rabins,
  K.~Ritchie, M.~Rossor, L.~Thal, and B.~Winblad.
\newblock Current concepts in mild cognitive impairment.
\newblock \emph{Archives of Neurology}, 58\penalty0 (12):\penalty0 1985, 2001.

\bibitem[Platt(1999)]{Platt1999}
J.~C. Platt.
\newblock Probabilistic outputs for support vector machines and comparisons to
  regularized likelihood methods.
\newblock \emph{Advances in Large Margin Classifiers}, 10\penalty0
  (3):\penalty0 61--74, 1999.

\bibitem[Rakotomamonjy et~al.(2008)Rakotomamonjy, Bach, Canu, and
  Grandvalet]{SimpleMKL}
A.~Rakotomamonjy, F.~R. Bach, S.~Canu, and Y.~Grandvalet.
\newblock {SimpleMKL}.
\newblock \emph{Journal of Machine Learning Research}, 9:\penalty0 2491--2521,
  Nov. 2008.

\bibitem[Rasmussen and Nickisch(2010)]{GPMLtoolbox}
C.~E. Rasmussen and H.~Nickisch.
\newblock Gaussian {P}rocesses for {M}achine {L}earning ({GPML}) toolbox.
\newblock \emph{The Journal of Machine Learning Research}, 9999:\penalty0
  3011--3015, 2010.

\bibitem[Rasmussen and Williams(2005)]{GPMLbook}
C.~E. Rasmussen and C.~K.~I. Williams.
\newblock \emph{Gaussian Processes for Machine Learning (Adaptive Computation
  and Machine Learning)}.
\newblock The MIT Press, 2005.
\newblock ISBN 026218253X.

\bibitem[Rifkin and Klautau(2004)]{Rifkin2004}
R.~Rifkin and A.~Klautau.
\newblock In defense of one-vs-all classification.
\newblock \emph{J. Mach. Learn. Res.}, 5:\penalty0 101--141, Dec. 2004.
\newblock ISSN 1532-4435.

\bibitem[Salakhutdinov et~al.(2007)Salakhutdinov, Mnih, and
  Hinton]{Salakhutdinov2007}
R.~Salakhutdinov, A.~Mnih, and G.~Hinton.
\newblock Restricted boltzmann machines for collaborative filtering.
\newblock In \emph{Proceedings of the 24th international conference on Machine
  learning}, pages 791--798. ACM, 2007.

\bibitem[Sch\"{o}lkopf and Smola(2001)]{Scholkopf2001}
B.~Sch\"{o}lkopf and A.~J. Smola.
\newblock \emph{Learning with Kernels: {S}upport Vector Machines,
  Regularization, Optimization, and Beyond}.
\newblock MIT Press, Cambridge, MA, USA, 2001.
\newblock ISBN 0262194759.

\bibitem[Snelson and Ghahramani(2006{\natexlab{a}})]{Snelson2006a}
E.~Snelson and Z.~Ghahramani.
\newblock Sparse {G}aussian processes using pseudo-inputs.
\newblock In \emph{Advances in Neural Information Processing Systems 18}, pages
  1257--1264. MIT press, 2006{\natexlab{a}}.

\bibitem[Snelson and Ghahramani(2006{\natexlab{b}})]{Snelson2006b}
E.~Snelson and Z.~Ghahramani.
\newblock Variable noise and dimensionality reduction for sparse {G}aussian
  processes.
\newblock In \emph{Proceedings of the Twenty-Second Conference Annual
  Conference on Uncertainty in Artificial Intelligence (UAI-06)}, pages
  461--468, Arlington, Virginia, 2006{\natexlab{b}}. AUAI Press.

\bibitem[Sonnenburg et~al.(2006)Sonnenburg, R\"{a}tsch, Sch\"{a}fer, and
  Sch\"{o}lkopf]{Sonnenburg2006}
S.~Sonnenburg, G.~R\"{a}tsch, C.~Sch\"{a}fer, and B.~Sch\"{o}lkopf.
\newblock Large scale multiple kernel learning.
\newblock \emph{J. Mach. Learn. Res.}, 7:\penalty0 1531--1565, Dec. 2006.
\newblock ISSN 1532-4435.

\bibitem[Szegedy et~al.(2014)Szegedy, Liu, Jia, Sermanet, Reed, Anguelov,
  Erhan, Vanhoucke, and Rabinovich]{Szegedy2014}
C.~Szegedy, W.~Liu, Y.~Jia, P.~Sermanet, S.~Reed, D.~Anguelov, D.~Erhan,
  V.~Vanhoucke, and A.~Rabinovich.
\newblock Going deeper with convolutions.
\newblock \emph{arXiv preprint arXiv:1409.4842}, 2014.

\bibitem[Thatcher(1994)]{Thatcher1994}
R.~W. Thatcher.
\newblock \emph{Functional neuroimaging: technical foundations}.
\newblock Academic Press, 1994.
\newblock ISBN 9780126858457.

\bibitem[Tibshirani(1996)]{Tibshirani1996}
R.~Tibshirani.
\newblock Regression shrinkage and selection via the lasso.
\newblock \emph{Journal of the Royal Statistical Society. Series B
  (Methodological)}, pages 267--288, 1996.

\bibitem[Tipping(2001)]{Tipping2001}
M.~E. Tipping.
\newblock Sparse {B}ayesian learning and the relevance vector machine.
\newblock \emph{J. Mach. Learn. Res.}, 1:\penalty0 211--244, Sept. 2001.
\newblock ISSN 1532-4435.

\bibitem[Tu(2007)]{Tu2007}
Z.~Tu.
\newblock Learning generative models via discriminative approaches.
\newblock In \emph{Computer Vision and Pattern Recognition, 2007. CVPR'07. IEEE
  Conference on}, pages 1--8. IEEE, 2007.

\bibitem[Vapnik(1995)]{Vapnik1995}
V.~N. Vapnik.
\newblock \emph{The nature of statistical learning theory}.
\newblock Springer-Verlag New York, Inc., New York, NY, USA, 1995.
\newblock ISBN 0-387-94559-8.

\bibitem[Whitwell et~al.(2007)Whitwell, Przybelski, Weigand, Knopman, Boeve,
  Petersen, and Jack]{Whitwell2007}
J.~L. Whitwell, S.~A. Przybelski, S.~D. Weigand, D.~S. Knopman, B.~F. Boeve,
  R.~C. Petersen, and C.~R. Jack.
\newblock {3D} maps from multiple {MRI} illustrate changing atrophy patterns as
  subjects progress from mild cognitive impairment to {A}lzheimer's disease.
\newblock \emph{Brain}, 130\penalty0 (7):\penalty0 1777--1786, 2007.

\bibitem[{WHO International Programme on Chemical Safety}(2001)]{WHO2001}
{WHO International Programme on Chemical Safety}.
\newblock Biomarkers in risk assessment: Validity and validation.
\newblock World Health Organization International Programme on Chemical Safety,
  Environmental Health Criteria 155, 2001.

\bibitem[Wiering et~al.(2013)Wiering, Van~der Ree, Embrechts, Stollenga,
  Meijster, Nolte, and Schomaker]{Wiering2013}
M.~Wiering, M.~Van~der Ree, M.~Embrechts, M.~Stollenga, A.~Meijster, A.~Nolte,
  and L.~Schomaker.
\newblock The {N}eural {S}upport {V}ector {M}achine.
\newblock In \emph{The 25th Benelux Artificial Intelligence Conference
  (BNAIC)}. IEEE, 2013.

\bibitem[Williams and Barber(1998)]{Williams1998}
C.~K.~I. Williams and D.~Barber.
\newblock Bayesian classification with {G}aussian processes.
\newblock \emph{IEEE Trans. Pattern Anal. Mach. Intell.}, 20\penalty0
  (12):\penalty0 1342--1351, Dec. 1998.
\newblock ISSN 0162-8828.

\bibitem[Wipf and Nagarajan(2007)]{Wipf2007}
D.~P. Wipf and S.~S. Nagarajan.
\newblock A new view of automatic relevance determination.
\newblock In \emph{Advances in Neural Information Processing Systems}, pages
  1625--1632, 2007.

\bibitem[Wu et~al.(2004)Wu, Lin, and Weng]{Wu2004}
T.-F. Wu, C.-J. Lin, and R.~C. Weng.
\newblock Probability estimates for multi-class classification by pairwise
  coupling.
\newblock \emph{J. Mach. Learn. Res.}, 5:\penalty0 975--1005, Dec. 2004.
\newblock ISSN 1532-4435.

\bibitem[Xu et~al.(2009)Xu, Caramanis, and Mannor]{Xu2009}
H.~Xu, C.~Caramanis, and S.~Mannor.
\newblock Robustness and regularization of {S}upport {V}ector {M}achines.
\newblock \emph{The Journal of Machine Learning Research}, 10:\penalty0
  1485--1510, 2009.

\bibitem[Yang et~al.(2011)Yang, Lui, Gao, Chan, Yau, Sperling, and
  Huang]{Yang2011}
W.~Yang, R.~L.~M. Lui, J.~H. Gao, T.~F. Chan, S.~Yau, R.~A. Sperling, and
  X.~Huang.
\newblock {Independent component analysis-based classification of {A}lzheimer's
  Disease MRI Data}.
\newblock \emph{Journal of Alzheimer's Disease}, 24\penalty0 (4):\penalty0
  775--783, 2011.

\bibitem[Young et~al.(2013{\natexlab{a}})Young, Modat, Cardoso, Ashburner, and
  Ourselin]{Young2013b}
J.~Young, M.~Modat, M.~J. Cardoso, J.~Ashburner, and S.~Ourselin.
\newblock An oblique approach to prediction of conversion to {A}lzheimer's
  disease with multikernel {G}aussian processes.
\newblock In NIPS 2013 Workshop on Machine Learning and Interpretation in
  NeuroImaging, December 2013{\natexlab{a}}.

\bibitem[Young et~al.(2013{\natexlab{b}})Young, Modat, Cardoso, Mendelson,
  Cash, and Ourselin]{Young2013a}
J.~Young, M.~Modat, M.~J. Cardoso, A.~Mendelson, D.~Cash, and S.~Ourselin.
\newblock Accurate multimodal probabilistic prediction of conversion to
  {A}lzheimer's disease in patients with mild cognitive impairment.
\newblock \emph{NeuroImage: Clinical}, 2\penalty0 (0):\penalty0 735--745,
  2013{\natexlab{b}}.
\newblock ISSN 2213-1582.
\newblock \doi{http://dx.doi.org/10.1016/j.nicl.2013.05.004}.

\bibitem[Zhang et~al.(2011)Zhang, Wang, Zhou, Yuan, and Shen]{Zhang2011}
D.~Zhang, Y.~Wang, L.~Zhou, H.~Yuan, and D.~Shen.
\newblock Multimodal classification of {A}lzheimer's disease and mild cognitive
  impairment.
\newblock \emph{Neuroimage}, 55\penalty0 (3):\penalty0 856--867, 2011.

\bibitem[Zhao and Yu(2006)]{Zhao2006}
P.~Zhao and B.~Yu.
\newblock On model selection consistency of lasso.
\newblock \emph{J. Mach. Learn. Res.}, 7:\penalty0 2541--2563, Dec. 2006.
\newblock ISSN 1532-4435.
\newblock URL \url{http://dl.acm.org/citation.cfm?id=1248547.1248637}.

\bibitem[Zou and Hastie(2005)]{Zou2005}
H.~Zou and T.~Hastie.
\newblock Regularization and variable selection via the elastic net.
\newblock \emph{Journal of the Royal Statistical Society: Series B (Statistical
  Methodology)}, 67\penalty0 (2):\penalty0 301--320, 2005.

\end{thebibliography}
\end{spacing}

\end{document}